%% file: Chain.tex
\documentclass[10pt,twocolumn,letterpaper]{article}

\usepackage[pagenumbers]{cvpr} 

\input{preamble}

\input{mathcommands}

\usepackage{utfsym} 
\newcommand{\heart}{$\;\!$\usym{2665}}

\definecolor{cvprblue}{rgb}{0.21,0.49,0.74}
\usepackage[bookmarks=false,pagebackref,breaklinks,colorlinks,citecolor=cvprblue]{hyperref}
\usepackage{colortbl}
\definecolor{tblcolor}{HTML}{BEF8D4}

\def\paperID{2541} 
\def\confName{CVPR}
\def\confYear{2024}

\title{\chainsymb\  CHAIN: Enhancing Generalization in Data-Efficient GANs via lips\textit{CH}itz continuity constr\textit{AI}ned \textit{N}ormalization\vspace{-0.2cm}}

\author{%
Yao Ni$^{\dagger}$ \quad Piotr Koniusz\thanks{\iftoggle{cvprpagenumbers}{The corresponding author}{Corresponding author}.\iftoggle{cvprpagenumbers}{\quad This paper is accepted by CVPR 2024. }{ Code: \href{https://github.com/MaxwellYaoNi/CHAIN}{https://github.com/MaxwellYaoNi/CHAIN}}}$\;^{,\S,\dagger}$\\
$^{\dagger}$The Australian National University \quad $^\S$Data61$\!${\color{red}\heart}CSIRO \\
\vspace{-0.2cm}
{\tt\small $^{\dagger}$firstname.lastname@anu.edu.au}
}

\begin{document}
\maketitle

\begin{abstract}\iftoggle{cvprpagenumbers}{\vspace{-0.1cm}}{}%
Generative Adversarial Networks (GANs) significantly advanced image generation but their performance heavily depends on abundant training data. In scenarios with limited data, GANs often struggle with discriminator overfitting and unstable training. Batch Normalization (BN), despite being known for enhancing generalization and training stability, has rarely been used in the discriminator of Data-Efficient GANs. Our work addresses this gap by identifying a critical flaw in BN: the tendency for gradient explosion during the centering and scaling steps. To tackle this issue, we present CHAIN (lipsCHitz continuity constrAIned Normalization), which replaces the conventional centering step with zero-mean regularization and integrates a Lipschitz continuity constraint in the scaling step. CHAIN further enhances GAN training by adaptively interpolating the normalized and unnormalized features, effectively avoiding discriminator overfitting. Our theoretical analyses firmly establishes CHAIN's effectiveness in reducing gradients in latent features and weights, improving stability and generalization in GAN training. Empirical evidence supports our theory. CHAIN achieves state-of-the-art results in data-limited scenarios on CIFAR-10/100, ImageNet, five low-shot and seven high-resolution few-shot image datasets. \iftoggle{cvprpagenumbers}{Code: \href{https://github.com/MaxwellYaoNi/CHAIN}{https://github.com/MaxwellYaoNi/CHAIN}.}{}
\end{abstract}

\iftoggle{cvprpagenumbers}{\vspace{-0.3cm}}{\vspace{-0.1cm}}%
\section{Introduction}
The availability of abundant data, exemplified by  ImageNet \cite{ImageNet}, has  driven breakthroughs in deep neural networks \cite{lecun2015deep}, particularly in generative models. This data richness has fueled innovations such as Generative Adversarial Networks (GANs) \cite{goodfellow2014generative}, popular in academia and industry. GANs, known for their rapid generation speeds \cite{StyleGAN-T} and high-fidelity image synthesis \cite{sauer2022stylegan}, have become go-to tools for applications such as text-to-image generation \cite{StyleGAN-T, kang2023scaling, tao2023galip},  image-to-image translation \cite{richardson2021encoding, kong2021breaking, parmar2023zero, shiri2018identity}, video synthesis \cite{wang2023styleinv, Yu_2023_CVPR, Shen_2023_CVPR} and 3D generation \cite{10149489, skorokhodov2023d, zhang2023avatargen}.

Despite the advanced capabilities of modern GANs \cite{BigGAN, StyleGAN2, StyleGAN3} in creating high-fidelity images, their success largely depends on access to extensive training data. However, in scenarios with limited data, such as medical \cite{kim2022transfer} or art images \cite{khan2023q, koniusz2018museum}, where data acquisition is expensive and privacy concerns are paramount, GANs face issues such as discriminator overfitting and unstable training \cite{ADA, DA, Lecam}.

To overcome these obstacles,  three main directions stand out. The first leverages massive data augmentation (``MA"), aimed at broadening the available data distribution \cite{APA, ADA, DA, MaskedGAN, FakeCLR, InsGen, CRGAN}. The second strategy borrows knowledge from models trained on large datasets \cite{KDDLGAN, RegDGM, EnsemblingGAN, wang2018transferring}. However, these approaches 
suffer from issues such as the potential leakage of augmentation artifacts \cite{ADA, CRGAN, ni2022manifold, ojha2021few} and the misuse of pre-training knowledge \cite{RoleImageNet, xu2019ganobfuscator, ge2022learning}. The third direction addresses discriminator overfitting and focuses on discriminator regularization to either reduce the capacity of the discriminator \cite{DigGAN, DRAGAN, gulrajani2017improved, CAGAN, ni2024nice} or increase the overlap between real and fake data \cite{tao2020alleviation,APA,Lecam}, making it harder for the discriminator to learn. While such methods are effective, their mechanism preventing overfitting is not clearly elucidated.

Aligning with the third direction of regularizing the discriminator,  we innovate by reconsidering the integration of Batch Normalization (BN) \cite{BN} into the discriminator to improve the generalization. BN has been demonstrated, both theoretically and in practice, to improve neural network generalization. This is achieved via its standardization process, effectively aligning training and test distributions in a common space \cite{seo2020learning, wang2019transferable, li2016revisiting}. Additionally, BN reduces the sharpness of the loss landscape \cite{santurkar2018does, lyu2022understanding, karakida2019normalization, bjorck2018understanding} and stabilizes the training process by mitigating internal covariate shift.

Given these benefits, Batch Normalization appears as a good solution to preventing discriminator overfitting in GANs. However, large-scale experiments \cite{pmlr-v97-kurach19a, xiang2017effects, zhang2018convergence, mroueh2017fisher, SN} have shown that incorporating BN into the discriminator actually impairs performance. Thus, BN is often omitted in the discriminator of modern GANs, \eg, BigGAN \cite{BigGAN}, ProGAN \cite{ProgressiveGAN}, StyleGAN 1-3 \cite{StyleGAN, StyleGAN2, StyleGAN3}, with few models using BN in the discriminator, \ie, DCGAN \cite{radford2015unsupervised}.

Addressing the challenges of BN in GAN discriminator design, we have identified that the centering and scaling steps of BN can lead to gradient explosion, a significant barrier in GAN convergence \cite{zhang2018on, thanh-tung2018improving, DRAGAN, mescheder2018training}. To circumvent this issue while leveraging the benefits of BN, we propose replacing the centering step with zero mean regularization and enforcing the Lipschitz continuity constraint on the scaling step. This modification  resolves gradient issues and also helps the discriminator effectively balance discrimination and generalization \cite{zhang2018on}  through adaptive interpolation of normalized and unnormalized features. 

We call our approach lips\underline{CH}itz continuity constr\underline{AI}ned \underline{N}ormalization, in short, \our, symbolized as \chainsymb. Such a name and symbol represent the role of our model in bridging the gap between seen and unseen data and reducing the divergence between fake and real distributions. Despite \our's simplicity, our theoretical analysis confirms its efficacy in reducing the gradient norm of both latent activations and discriminator weights. Experimental evidence shows that \our stabilizes GAN training and enhances generalization.
\our outperforms existing methods that limit discriminator overfitting, achieving state-of-the-art results on data-limited benchmarks such as CIFAR-10/100, ImageNet, 5 low-shot and 7 high-resolution few-shot image generation tasks. Our contributions are as follows:

\vspace{0.1cm}
\renewcommand{\labelenumi}{\roman{enumi}.}
\begin{enumerate}[leftmargin=0.6cm]
\item We tackle discriminator overfitting by enhancing GAN generalization, deriving a new error bound that emphasizes reducing the gradient of discriminator weights.
\item We identify that applying BN in the discriminator, both theoretically and empirically, tends to cause gradient explosion due to the centering and scaling steps of BN.
\item We provide evidence, both theoretical and practical, that \our stabilizes GAN training by moderating the gradient of latent features, and improves generalization by lowering the gradient of the weights.
\end{enumerate}

\vspace{-0.1cm}
\section{Background}
\textbf{Improving GANs.}
Generative Adversarial Networks \cite{goodfellow2014generative}, effective in image generation \cite{BigGAN, StyleGAN2, FastGAN}, image-to-image translation \cite{kong2021breaking, CycleGAN, fatima_image_to_image,shiri2019identity}, video synthesis \cite{wang2023styleinv, Yu_2023_CVPR, Shen_2023_CVPR}, 3D generation \cite{10149489, skorokhodov2023d, zhang2023avatargen} and text-to-image generation \cite{StyleGAN-T, kang2023scaling, tao2023galip}, suffer from unstable training  \cite{DRAGAN, thanh-tung2018improving}, mode collapse \cite{roth2017stabilizing, mescheder2018training}, and discriminator overfitting \cite{DA, ADA}. Improving GANs includes architecture modifications \cite{BigGAN, StyleGAN, StyleGAN2, StyleGAN3, lee2021vitgan, zhang2022styleswin}, loss function design \cite{arjovsky2017wasserstein, zhao2016energy, fGAN, OmniGAN} and regularization design \cite{gulrajani2017improved, Lecam, DRAGAN, SN, Liu_2019_ICCV}. BigGAN \cite{BigGAN} scales up GANs for large-scale datasets with increased batch sizes. StyleGANs  \cite{StyleGAN, StyleGAN2, StyleGAN3} revolutionize generator architecture by style integration. OmniGAN \cite{OmniGAN} modifies the projection loss \cite{miyato2018cgans} into a multi-label softmax loss. WGAN-GP \cite{gulrajani2017improved}, SNGAN \cite{SN} and SRGAN \cite{Liu_2019_ICCV} regularize discriminator using a gradient penalty or spectral norm constraints for stable training. Our novel normalization effectively enhances GANs under limited data scenarios, applicable across various architectures and loss functions.

\vspace{0.1cm}
\noindent\textbf{Image generation under limited data.} To address discriminator overfitting in limited data scenarios, where data is scarce or privacy-sensitive, previous methods have employed data augmentation techniques such as DA \cite{DA}, ADA \cite{ADA}, MaskedGAN \cite{MaskedGAN}, FakeCLR \cite{FakeCLR} and InsGen \cite{InsGen} to expand the data diversity. Approaches \cite{RegDGM, EnsemblingGAN}, KDDLGAN \cite{KDDLGAN}, and TransferGAN \cite{wang2018transferring}, leverage knowledge from models trained on extensive datasets to enhance performance. However, these approaches may risk leaking augmentation artifacts \cite{ADA, CRGAN, ojha2021few} or misusing pre-trained knowledge \cite{RoleImageNet, xu2019ganobfuscator, ge2022learning}. Alternatives such as LeCam loss \cite{Lecam}, GenCo \cite{GenCo} and the gradient norm reduction of DigGAN \cite{DigGAN} aim to balance real and fake distributions. Our approach uniquely combines generalization benefits from BN with improved stability in GAN training, offering an effective and distinct solution to regularizing discriminator.

\vspace{0.1cm}
\noindent\textbf{GAN Generalization.} Deviating from conventional methods that link the generalization of GANs \cite{zhang2018on, ji2021understanding} with the Rademacher complexity \cite{bartlett2002rademacher} of neural networks \cite{Zhang_2024_WACV}, we introduce a new error bound that highlights the need for reducing discrepancies between seen and unseen data for enhanced generalization. This bound is further refined using the so-called non-vacuous PAC-Bayesian theory \cite{catoni2007pac}, focusing on discriminator weight gradients for a practical GAN generalization improvement.

\vspace{0.1cm}
\noindent\textbf{Normalization.} Batch Normalization (BN) \cite{BN}  and its variants such as Group Normalization (GN) \cite{Wu_2018_ECCV}, Layer Normalization (LN) \cite{ba2016layer}, Instance Normalization (IN) \cite{Ulyanov2016InstanceNT} have been pivotal in normalizing latent features to improve training. BN, in particular, is renowned for its role in improving generalization across various tasks \cite{santurkar2018does, lyu2022understanding, karakida2019normalization, bjorck2018understanding}. However, its application in discriminator design, especially under limited data scenarios where generalization is crucial, remains underexplored. Several BN modifications, such as RMSNorm \cite{RMSNorm}, GraphNorm \cite{GraphNorm}, PowerNorm \cite{PowerNorm}, MBN \cite{MBN} and EvoNorm \cite{EvoNorm} have been proposed to address issues such as the gradient explosion in transformers \cite{Transformer} or information loss in graph learning, often by altering or removing the centering step. Our work stands out in GAN discriminator design by linking centering, scaling, and gradient issues in GAN training. Our innovative solution not only mitigates the gradient explosion but also retains the benefits of BN, offering a robust solution for GAN training.

\section{Method}
We begin by linking GAN generalization with the gradient of discriminator weights, motivating the use of BN for generalization and identifying gradient issues in BN. We then introduce \our, a design that tackles these gradient issues while retaining benefits of BN. Lastly, we present a theoretical justification for \our, underscoring its efficacy in improving generalization and training stability.

\subsection{Generalization Error of GAN}
The goal of GAN is to train a generator capable of deceiving a discriminator by minimizing the integral probability metric (IPM) \cite{muller1997integral}, typically with the assumption of infinite real and fake distributions $(\mu, \nu)$. However, in real-world scenarios, we are usually confined to working with a finite real dataset $\empmu$ of size $n$. This limitation restricts the optimization of GAN to the empirical loss as discussed in \cite{zhang2018on}:
{\setlength{\abovedisplayskip}{0.2cm}%
\setlength{\belowdisplayskip}{0.2cm}%
\begin{equation}
    \!\!\inf_{\nu\in\calG}\!\!\big\{d_\calH(\empmu, \nu)\!:=\!\sup_{h\in\calH}\{\bbE_{\vxr\sim\empmu}[h(\vxr)]-\bbE_{\vxf\sim\nu}[h(\vxf)]\}\!\big\},\label{eq:ipm}
\end{equation}}%
where $\vxr$ and $\vxf$ are real and fake samples. Function sets of discriminator and generator, $\calH$ and $\calG$, are typically parameterized as neural network classes $\calH_\text{nn}\!:=\!\{h(\cdot;\vtheta_d)\!:\!\vtheta_d\!\in\!\mTheta_d\}$ and $\calG_\text{nn}\!:=\!\{g(\cdot;\vtheta_g)\!:\!\vtheta_g\!\in\!\mTheta_g\}$. Given the varied divergence \cite{zhang2018on, shannon2020non} encompassed by the IPM and the variability of discriminator loss function $\phi(\cdot)$ across different tasks and architectures, we integrate it with the discriminator $D$ for simplified analysis \cite{zhang2018on, arjovsky2017wasserstein, arora2017generalization}, yielding $h(\cdot)\!:=\!\phi(D(\cdot))$. This integration streamlines the alternating optimization process between the discriminator and the generator:
{\setlength{\abovedisplayskip}{0.2cm}%
\setlength{\belowdisplayskip}{0.2cm}%
\begin{equation}\label{eq:GAN_loss}
\left \{
\setlength{\jot}{0pt}
\begin{aligned}
    \calL_D &:= \min_{\vtheta_d} \bbE_{\vxf\sim\nu_n}[h(\vxf;\vtheta_d)] - \bbE_{\vxr\sim\empmu}[h(\vxr;\vtheta_d)]\\
    \calL_G &:= \min_{\vtheta_g} - \bbE_{\vz\sim p_{\vz}}[h(g(\vz;\vtheta_g))],
\end{aligned}
\right. 
\end{equation}}%
where $\vz\sim p_z$ represents the noise input to the generator and it is assumed that $\nu_n$ minimizes $d_\calH(\empmu, \nu)$ to a precision $\epsilon\!\geq\!0$, implying that $d_\calH(\empmu, \nu_n)\!\leq\! \inf_{\nu\in\calG}d_\calH(\empmu,\nu)\!+\!\epsilon$.

To evaluate how closely the generator distribution $\nu_n$ approximates the unknown infinite distribution $\mu$, we draw on  work of Ji \etal \cite{ji2021understanding} who extended Theorem 3.1 in \cite{zhang2018on} by considering the limited access to both real and fake images.
\begin{lemma}\label{lemma:gen_error} (Partial results of Theorem 1 in \cite{ji2021understanding}.) Assume the discriminator set $\calH$ is even, \ie, $h\!\in\!\calH$ implies $-h\!\in\!\calH$, and $\lVert h\rVert_\infty\leq\!\Delta$. Let $\empmu$ and $\empnu$ be empirical measures of $\mu$ and $\popnu$ with size $n$. Denote $\optnu\!=\!\inf_{\nu\in\calG}d_\calH(\empmu, \nu)$. The generalization error of GAN, defined as $\ganerr\!:=\!d_\calH(\mu,\popnu)\!-\!\inf_{\nu\in\calG}d_\calH(\mu, \nu)$, is  bounded as:
{\setlength{\abovedisplayskip}{0.2cm}%
\setlength{\belowdisplayskip}{0.2cm}%
\begin{align}
    \ganerr&\leq2\big(\sup_{h\in\calH}\big|\bbE_{\mu}[h]-\bbE_{\empmu}[h]\big|+\sup_{h\in\calH}\big|\bbE_{\optnu}[h]-\bbE_{\empnu}[h]\big|\big)\nonumber\\
    &=2d_\calH(\mu, \empmu)+2d_\calH(\optnu, \empnu).
\end{align}}%
\end{lemma}%
Lemma \ref{lemma:gen_error} (proof in \S \ref{sm:proof:lemma:gen_error}) indicates that GAN generalization can be improved by reducing the divergence between real training and unseen data, as well as observed and unobserved fake distributions. Given that the ideal $\optnu$ aligns with the observed real data $\empmu$, Lemma \ref{lemma:gen_error} also emphasizes narrowing the gap between observed fake and real data to lower $d_\calH(\optnu, \empnu)$. This explains why prior efforts \cite{Lecam, APA, DigGAN, hou2024augmentation, chen2021data} focusing on diminishing the real-fake distribution divergence help limit overfitting. However, excessive reduction should be avoided, as this makes the discriminator struggle to differentiate real and fake data \cite{zhang2018on}.

While reducing $d_\calH(\optnu, \empnu)$ is achievable, lowering $d_\calH(\mu, \empmu)$ remains challenging due to inaccessibility of infinite $\mu$. Fortunately, neural network parameterization of GANs enables adopting PAC Bayesian theory \cite{catoni2007pac} to further analyze $d_\calH(\mu, \empmu)$. Integrating the analysis of Theorem 1 in \cite{foret2021sharpnessaware}, Lemma \ref{lemma:gen_error} is further formulated as follows:
\begin{prop}\label{prop:gen_sharpness} Utilizing notations from Lemma \ref{lemma:gen_error}, we define $\ganerrn$ as the generalization error of GAN parameterized as neural network classes. Let $\gradr$ and $\hessr$ represent the gradient and Hessian matrix of discriminator $h$ evaluated at $\vtheta_d$ over real training data $\empmu$, and $\gradf$ and $\hessf$ over observed fake data $\empnu$. Denoting $\levr$ and $\levf$ as the largest eigenvalues of $\hessr$ and $\hessf$, respectively, and for any $\omega>0$, the generalization error is bounded as:
{\setlength{\abovedisplayskip}{0.2cm}%
\setlength{\belowdisplayskip}{0.2cm}%
\begin{align}
    \ganerrn\leq& 2\omega\big(\lVert\gradr\rVert_2+\lVert\gradf\rVert_2\big)+4R\Big(\frac{\lVert\vtheta_d\rVert_2^2}{\omega^2}, \frac{1}{n}\Big)\nonumber\\
    &+\omega^2\big(|\levr| + |\levf|\big),
\end{align}}%
where $R\big(\frac{\lVert\vtheta_d\rVert_2^2}{\omega^2}, \frac{1}{n}\big)$,  a term related to discriminator weights norm, is inversely related to the data size $n$.
\end{prop}
Prop. \ref{prop:gen_sharpness} (proof in \S \ref{sm:proof:prop:gen_sharpness}) suggests several strategies to lower the generalization error of GANs. These include increasing data size ($n$), implementing regularization to decrease weight norm of the discriminator and the largest eigenvalues in Hessian matrices, and crucially, reducing the gradient norm of discriminator weights. Although this proposition is specific to GANs, the concept of regularizing weight gradient norms aligns with findings in other studies \cite{zhang2023gradient, wang2023sharpness, zhao2022penalizing, liu2022random, sun2020fisher, simon2020modulating}, which emphasize that reducing weight gradients can smooth the loss landscape, thereby enhancing generalization of various deep learning tasks.

\begin{figure*}[t!]
    \centering
    \hspace{-0.2cm}
    \begin{minipage}{0.68\linewidth}
        \includegraphics[width=\linewidth]{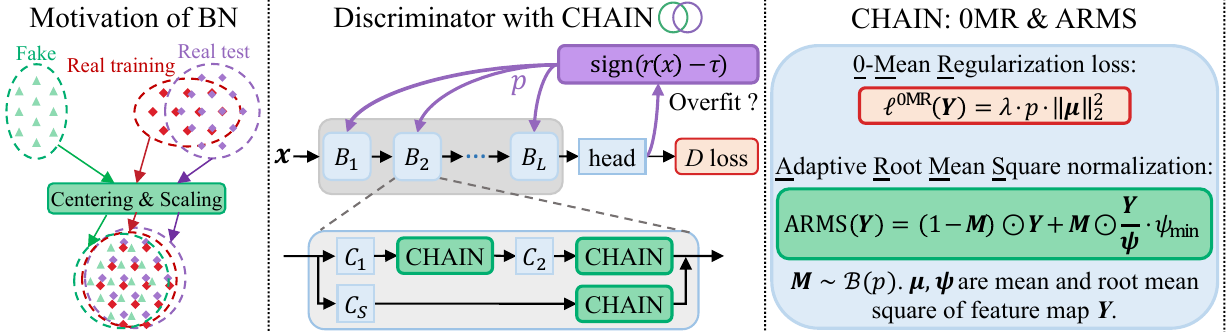}
    \end{minipage}
    \begin{minipage}{0.315\linewidth}
\scriptsize
\newcommand{\bcs}{\hspace{-0.05cm}}
\begin{tabular}{@{\bcs}l@{\bcs}}
\toprule
Pytorch-style pseudo code for CHAIN$_\text{batch}$\\
\midrule
\PyComment{Y:BxdxHxW $\!$size; $\!$lbd:hyperparameter $\lambda$}\\
\PyCode{\pykey{def} \pyfun{CHAIN\_batch}$\!$(\pyvar{Y}, \pyvar{p}, \pyvar{lbd}, \pyvar{eps}\pyop{=}\pynum{1e-5})\pyop{:}} \\
\rule{0pt}{5pt}\quad\PyCode{M\pyop{=}$\!$(torch$\!$.$\!$\pyfun{rand}$\!$($\!$\pyop{*}\pyvar{Y}$\!$.$\!$shape$\!$[$\!$\pyop{:}$\!$\pynum{2}$\!$]$\!$,$\!$\pynum{1}$\!$,$\!$\pynum{1}$\!$)\pyop{<}\pyvar{p}) $\!$\pyop{*} $\!$\pynum{1.0}}\\
\rule{0pt}{5pt}\quad\PyCode{psi\_s\pyop{=}\pyvar{Y}$\!$.$\!$\pyfun{square}$\!$()$\!\!$.$\!$\pyfun{mean}$\!$($\!$[$\!$\pynum{0}$\!$,$\!$\pynum{2}$\!$,$\!$\pynum{3}$\!$]$\!$,$\!$keepdim\pyop{=}\pykey{True}$\!$)}\\
\rule{0pt}{5pt}\quad\PyCode{psi\pyop{=}(psi\_s $\!$+$\!$ \pyvar{eps})$\!\!$.\pyfun{sqrt}$\!$()}\\
\rule{0pt}{5pt}\quad\PyCode{psi\_min\pyop{=}psi$\!$.\pyfun{min}$\!$()$\!\!$.$\!$\pyfun{detach}$\!$()}\\
\rule{0pt}{5pt}\quad\PyCode{Y\_arms\pyop{=}$\!$(\pynum{1}\pyop{-}M)\pyop{*}\pyvar{Y}$\!\!$ \pyop{+} $\!\!$M\pyop{*}$\!$(\pyvar{Y}\pyop{/}psi\pyop{*} $\!\!$psi\_min)}\\
\rule{0pt}{5pt}\quad\PyCode{reg\pyop{=}Y$\!$.\pyfun{mean}$\!$($\!$[$\!$\pynum{0}$\!$,$\!$\pynum{2}$\!$,$\!$\pynum{3}$\!$]$\!$)$\!\!$.$\!$\pyfun{square}$\!$()$\!$.$\!$\pyfun{sum}$\!$()\pyop{*}(\pyvar{p}\pyop{*}\pyvar{lbd})}\\ 
\rule{0pt}{5pt}\quad\PyCode{\pykey{return} Y\_arms, reg}\\
\bottomrule
\end{tabular}
\end{minipage}
\vspace{-0.1cm}
\caption{Motivation of using BN, discriminator with \our, modules in \our and the Pytorch-style pseudo-code for CHAIN$_\text{batch}$.}
\vspace{-0.2cm}
\label{fig:pipline}
\end{figure*}

\subsection{Motivation and the Batch Normalization Issues}
Leveraging Lemma \ref{lemma:gen_error} and Prop. \ref{prop:gen_sharpness} insights that reducing real-fake divergence and gradient norms boosts generalization, we propose applying BN in the discriminator to normalize real and fake data \textit{in separate batches}. As depicted in Figure \ref{fig:pipline}, normalizing real and fake data in separate batches via the centering and scaling steps aligns their statistical moments to lower the real-fake divergence per Lemma \ref{lemma:gen_error}. Moreover, BN's ability to reduce sharpness, as indicated by the maximum Hessian eigenvalue \cite{santurkar2018does, lyu2022understanding, karakida2019normalization}, supports the motivation of using BN for better generalization. Yet, incorporating BN risks gradient explosion.

For a specific layer in a network, consider $\mA\in\bbR^{B\times d}$ as the feature input, where $B$ is the batch size and $d$ is the feature size. For brevity, we exclude bias and focus on layer weights $\mW\in\bbR^{d\times d}$. In line with studies \cite{PowerNorm, GraphNorm, luo2018towards, santurkar2018does}, we also omit the affine transformation step for theoretical clarity, as it does not impact the theoretical validity, and does not change our method. The processing of features through the weights and the Batch Normalization contains:
{\setlength{\abovedisplayskip}{0.2cm}%
\setlength{\belowdisplayskip}{0.2cm}%
\setlength{\jot}{0pt}
\begin{align}
    \text{Linear transformation:\quad} & \mY=\mA\mW \\
    \text{Centering:\quad} & \mYc = \mY-\vmuy\label{eq:centering}\\
    \text{Scaling:\quad} & \mYs = \mYc / \vsigmay.\label{eq:scaling}
\end{align}}

Using these notations, we identify the gradient issues in the centering and scaling steps, as detailed below.
\begin{theorem}\label{theorem:bn_centering}(The issue of the centering step.)
 Consider $\vy_1, \vy_2$ as i.i.d. samples from a symmetric distribution centered at $\vmu$, where the presence of $\vy$ implies $2\vmuy-\vy$ is also included (important in proof). After the centering step, $\vyc_1, \vyc_2$ are i.i.d. samples from the centered distribution. The expected cosine similarity between these samples is given by:
{\setlength{\abovedisplayskip}{0.2cm}%
\setlength{\belowdisplayskip}{0.2cm}%
\begin{equation}
     \bbE_{\vy_1, \vy_2}\big[\cos(\vy_1, \vy_2)]\geq\bbE_{\vyc_1, \vyc_2}\big[\cos(\vyc_1, \vyc_2)\big]=0.
 \end{equation}}%
\end{theorem}
Theorem \ref{theorem:bn_centering} (proof in \S \ref{sm:proof:theorem:bn_centering}) states that after centering by batch normalization, the expected cosine similarity between features drops to zero. This implies that features which are similar in early network layers diverge significantly in the later layers, suggesting that minor perturbations in early layers have the risk to lead to abrupt changes in later layers. Consequently, such an effect implies large gradients.

\begin{theorem}\label{theorem:bn_scaling}(The issue of the scaling step.) The scaling step, defined in Eq. \ref{eq:scaling}, can be expressed as matrix multiplication $\mYs = \mYc\diag(1/\vsigmay)$. The Lipschitz constant \wrt the 2-norm of the scaling step is:
{\setlength{\abovedisplayskip}{0.2cm}%
\setlength{\belowdisplayskip}{0.2cm}%
\begin{equation}
    \Big\lVert\diag\bigg(\frac{1}{\vsigmay}\bigg)\Big\rVert_\lc=\frac{1}{\sigmamin},
\end{equation}}%
where $\sigmamin=\min_{c}\sigma_c$ represents the minimum value in $\vsigmay$.
\end{theorem}
\vspace{-0.2cm}%
Theorem \ref{theorem:bn_scaling} (proof in \S \ref{sm:proof:theorem:bn_scaling}) establishes that the Lipschitz constant for the scaling step in batch normalization is inversely proportional to $\sigmamin$. This means if $\sigmamin$ is less than 1, the Lipschitz constant exceeds 1. Given the emphasis placed by previous studies \cite{SN, arjovsky2017wasserstein, gulrajani2017improved, lin2021why, Chu2020Smoothness} on the importance of lowering the Lipschitz constant in the discriminator, it follows that without a Lipschitz continuity constraint on the scaling step, discriminators employing batch normalization are prone to gradient explosion. See  \cite{gouk2021regularisation} for further insights into the Lipschitz constant of batch normalization concerning the affine transformation step.

\subsection{\our \chainsymb}
To harness the generalization benefits of BN while sidestepping its gradient issue in GAN discriminator, we introduce \our. Our modification involves replacing the centering step (as in Eq. \ref{eq:centering}) with zero-mean regularization, substituting the scaling step (as in Eq. \ref{eq:scaling}) with Lipschitz continuity constrained root mean square normalization, and removing the affine transformation step for enhanced performance. 

We start by calculating the mean $\vmuy$ and the root mean square $\vpsiy$ across batch and spatial dimensions for features $\mY\in\bbR^{B\times d\times H \times W}$ in a discriminator layer as follows:
 {\setlength{\abovedisplayskip}{0.2cm}%
\setlength{\belowdisplayskip}{0.2cm}%
\begin{align}
    \muyc&=\frac{1}{B\times H\times W}\sum_{b}^B\sum_h^{H}\sum_w^{W}Y_{b,c,h,w},\\
    \psiyc&=\sqrt{\bigg(\frac{1}{B\!\times\!H\!\times\!W}\sum_{b}^B\sum_h^{H}\sum_w^{W}Y^2_{b,c,h,w}\bigg)+\epsilon},
\end{align}}%
where $\epsilon$ is a small constant to avoid division by 0. The term $Y_{b,c,h,w}$ denotes the $(b,c,h,w)$-th entry in $\mY$ while $\muyc$ and $\psiyc$ represent the $c$-th element in $\vmuy$ and $\vpsiy$, respectively. 

To achieve a soft zero-mean effect akin to the centering step in Eq. \ref{eq:centering} while also avoid its gradient issue, we adopt \underline{0}-\underline{M}ean \underline{R}egularization (0MR) as follows:
 {\setlength{\abovedisplayskip}{0.2cm}%
\setlength{\belowdisplayskip}{0.2cm}%
\begin{equation}
    \ell^{\text{0MR}}(\mY)=\lambda\cdot p\cdot\lVert\vmuy\rVert_2^2, \label{eq:0mr}
\end{equation}}%
where $\lambda$ is a hyperparameter and $p\in[0,1]$ adaptively controls the regularization strength. The term $\ell^{\text{0MR}}$ for layers applying \our is added to the discriminator loss. 0MR gradually adjusts feature means toward 0 during training and regularizes preceding layers to collaboratively achieve the 0-mean effect, ensuring smooth transitions between layers and training iterations, thereby avoiding gradient issues.

The root mean square normalization, constrained by Lipschitz condition, is defined as follows:
 {\setlength{\abovedisplayskip}{0.2cm}%
\setlength{\belowdisplayskip}{0.2cm}%
\begin{equation}
\widehat{\mY} = \widecheck{\mY} \cdot \psimin, \quad \text{with} \quad \widecheck{\mY}=\frac{\mY}{\vpsiy}.\label{eq:rms}
\end{equation}%
where $\psimin\!\!=\!\!\min_c\psi_c$ is the minimum in $\vpsiy$, severing to constrain the Lipschitz constant of the normalization to 1.

Normalized features are then adaptively interpolated with unnormalized features to balance discrimination and generalization, as emphasized in \cite{zhang2018on}, leading to the \underline{A}daptive \underline{R}oot \underline{M}ean \underline{S}quare normalization (ARMS):
 {\setlength{\abovedisplayskip}{0.2cm}%
\setlength{\belowdisplayskip}{0.2cm}%
\begin{align}
    \text{ARMS}(\mY)\!=\!(1\!-\!\mM)\odot\mY\!+\!\mM\odot\frac{\mY}{\vpsiy}\cdot\psimin,\label{eq:brmsn}
\end{align}}%
where $\odot$ is the element-wise multiplication after expanding the left-side matrix to $B\!\times\!d\!\times\!H\!\times\!W$ dimension. The matrix $\mM\in\bbR^{B\times d}$, with values from a Bernoulli distribution $\calB(p)$ with $p\in[0,1]$, controls the interpolation ratio. 

To mitigate discriminator overfitting, we allow the factor $p$, controlling both the regularization strength in Eq. \ref{eq:0mr} and the interpolation ratio in Eq. \ref{eq:brmsn}, to be adaptive based on the discriminator output. Specifically, we calculate the expectation of discriminator output $r(\vxr) = \bbE[\text{sign}(D(\vxr))]$ \wrt real samples $\vxr$ and assess $\varepsilon=\text{sign}(r(\vxr)\!-\!\tau)\!\in\!\{-1,0,1\}$ against a predefined threshold $\tau$. Exceeding $\tau$ suggests potential overfitting, as indicated by previous studies \cite{ADA, APA}. We then adjust $p$ using $p_{t+1} = p_t + \Delta_p \cdot \varepsilon$ with a small $\Delta_p$.

To limit the dependency on the minibatch size in high-resolution GAN training across multiple GPUs, we adopt running cumulative forward/backward statistics, inspired by \cite{ioffe2017batch, MBN, PowerNorm}. We contrast CHAIN$_\text{batch}$, using batch statistics, with \our that applies running cumulative statistics. CHAIN$_\text{batch}$ is elegantly coded as shown in Figure \ref{fig:pipline}, whereas implementation for \our is detailed in \S \ref{sm:sec:impl_chain}.

As outlined in Figure \ref{fig:pipline}, \our is integrated after convolutional layers $c\!\in\!\{C_1, C_2, C_S\}$ within the discriminator blocks $B_l$ for $l\!\in\!\{1,...,L\}$. By applying \our separately on real and fake data, Eq. \ref{eq:0mr} naturally reduces divergence across seen/unseen and observed real/fake data, consistent with Lemma \ref{lemma:gen_error}. Additionally, Eq. \ref{eq:brmsn} effectively lowers  weight gradients of discriminator, aligning with Prop. \ref{prop:gen_sharpness}.

\subsection{Theoretical analysis for \our \chainsymb} 
Although \our is straightforward and easy to implement, its importance in GAN training is substantial. We provide analyses of how \our modulates gradients, underlining its critical role in enhancing GAN performance. 
\begin{theorem}\label{theorem:chain_grad}(\our reduces the gradient norm of weights/latent features.) Denote the loss of discriminator with \our as $\calL$, and the resulting batch features as $\mYo$. Let $\vysch\!\in\!\bbR^B$ be $c$-th column of $\widecheck{\mY}$, $\Deltavych, \Deltavyoch\!\in\!\bbR^{B}$ be the $c$-th column of gradient $\frac{\partial \calL}{\partial \mY}, \frac{\partial \calL}{\partial \mYo}$. Denote $\Deltavwc$ as the $c$-th column of weight gradient $\frac{\partial \calL}{\partial \mW}$ and $\lev$ as the largest eigenvalue of pre-layer features $\mA$. Then we have:
 {\setlength{\abovedisplayskip}{0.2cm}%
\setlength{\belowdisplayskip}{0.2cm}%
\begin{align}
    \lVert\Deltavych\rVert_2^2\leq&\lVert\Deltavyoch\rVert_2^2\Big(\frac{(1-p)\psi_c+p\psimin}{\psi_c}\Big)^2\nonumber\\
    &-\frac{2(1-p)p\psimin}{B\psi_c}(\Deltavyoch^T\vysch)^2,\\
    \lVert\Deltavwc\rVert_2^2\leq&\lev^2\lVert\Deltavych\rVert_2^2.
\end{align}}%
\end{theorem}
Theorem \ref{theorem:chain_grad} (proof in \S \ref{sm:proof:theorem:chain_grad}) reveals that \our significantly modulates gradient norms in GAN training. It states that the squared gradient norm of normalized output is rescaled by $\big(\frac{(1-p)\psi_c+p\psimin}{\psi_c}\big)^2\!\leq\!1$, minus a non-negative term where $(\Deltavyoch^T\vysch)^2\!\geq\!0$. Considering that $\lVert\Deltavych\rVert_2^2\!\geq\!0$, \our effectively reduces the gradient norm of latent features. Moreover, given that the eigenvectors of $\diag(1/\vsigmay)$ and pre-layer features $\mA$ are less likely to align, using \our with a Lipschitz constant of exactly 1 before $\mA$ further reduces $\lev$. This dual action not only stabilizes GAN training by reducing latent feature gradients but also improves generalization by lowering the weight gradients.

We additionally present theory and experiments in \S \ref{sm:sec:stochastic_M} to justify the decorrelation effect of the stochastic $\mM$ design.

\begin{table*}[t!]
\begin{center}
\vspace{-0.1cm}
\caption{Comparing CIFAR-10/100 results with varying data percentages, using \our \vs without it. MA: Massive Augmentation.}
\vspace{-0.3cm}
\label{tab:results_CIFAR}
\footnotesize
\setlength{\tabcolsep}{0.05cm}
\renewcommand{\arraystretch}{0.95}
\begin{tabular}{!{\vrule width \boldlinewidth}l|c!{\vrule width \boldlinewidth}ccc|ccc|ccc!{\vrule width \boldlinewidth}ccc|ccc|ccc!{\vrule width \boldlinewidth}}
\topline
 \multirow{3}{*}{Method} & \multirow{3}{*}{MA} & \multicolumn{9}{c!{\vrule width \boldlinewidth}}{CIFAR-10} & \multicolumn{9}{c!{\vrule width \boldlinewidth}}{CIFAR-100}\\
 \cline{3-20}
& & \multicolumn{3}{c|}{10\% data} &  \multicolumn{3}{c|}{20\% data} & \multicolumn{3}{c!{\vrule width \boldlinewidth}}{100\% data} & \multicolumn{3}{c|}{10\% data} &  \multicolumn{3}{c|}{20\% data}& \multicolumn{3}{c!{\vrule width \boldlinewidth}}{100\% data}\\ 
\cline{3-20}
& & IS$\uparrow$ & tFID$\downarrow$ & vFID$\downarrow$ & IS$\uparrow$
& tFID$\downarrow$ & vFID$\downarrow$ & IS$\uparrow$
& tFID$\downarrow$ & vFID$\downarrow$ & IS$\uparrow$
& tFID$\downarrow$ & vFID$\downarrow$ & IS$\uparrow$
& tFID$\downarrow$ & vFID$\downarrow$ & IS$\uparrow$
& tFID$\downarrow$ & vFID$\downarrow$ \\
\middleline
BigGAN($d\!=\!256$) & $\times$ & 8.24 & 31.45 & 35.59 & 8.74 & 16.20 & 20.27 & 9.21 & 5.48 & 9.42      & 7.58 & 50.79 & 55.04 & 9.94 & 25.83 & 30.79  & 11.02 & 7.86 & 12.70  \\
\hline
+DA                 & \checkmark &  8.65 & 18.35 & 22.04  &  8.95 & 9.38 & 13.26 & 9.39 & 4.47 & 8.58  &  8.86 & 27.22 & 31.80 & 9.73 & 16.32 & 20.88  & 10.91 & 7.30 & 11.99 \\
+DigGAN+DA          & \checkmark &  $-$ & $-$ & 17.87 & $-$ & $-$ & 13.01 & $-$ & $-$ & 8.49 & $-$ & $-$  & 24.59 & $-$ & $-$ & 19.79 & $-$ & $-$ & 11.63 \\ 
+LeCam              & $\times$ & 8.44 & 28.36 & 33.65 & 8.95 & 11.34 & 15.25  & 9.45 & 4.27 & 8.29     & 8.14 & 41.51 & 46.43 & 10.05 & 20.81 & 25.77 &  \textbf{11.41} & 6.82 & 11.54 \\ 
\rowcolor{tblcolor}+\our       & $\times$ & \textbf{8.63} & \textbf{12.02} & \textbf{16.00} 
                               & \textbf{8.98} & \textbf{8.15} & \textbf{12.12} & \textbf{9.49} & \textbf{4.18} & \textbf{8.21}
                               & \textbf{10.04}&\textbf{13.13}&\textbf{18.00}
            				   & \textbf{10.15}&\textbf{11.58}&\textbf{16.38}
            				   & 11.16&\textbf{6.04}&\textbf{10.84}
            \\
\middleline
LeCam+DA            & \checkmark & 8.81 & 12.64 & 16.42 & 9.01 &  8.53 & 12.47 & 9.45 & 4.32 & 8.40         & 9.17 & 22.75 & 27.14 & 10.12 & 15.96 & 20.42  & 11.25 & 6.45 & 11.26 \\
\hline
+KDDLGAN           & \checkmark & $-$ & $-$ & 13.86 &  $-$ & $-$ &  11.15 & $-$ & $-$ &  8.19              & $-$ & $-$ & 22.40 & $-$ & $-$ & 18.70 & $-$ & $-$ & 10.12 \\
\rowcolor{tblcolor}+\our               & \checkmark & \textbf{8.96} & \textbf{8.54} & \textbf{12.51}       & \textbf{9.27} & \textbf{5.92} & \textbf{9.90} & \textbf{9.52} & \textbf{3.51} & \textbf{7.47} & \textbf{10.11}&\textbf{12.69}&\textbf{17.49} & \textbf{10.62}& \textbf{9.02}&\textbf{13.75} &  \textbf{11.37}&\textbf{5.26}&\textbf{9.85} \\
\middleline
OmniGAN($d\!=\!\!1024$) & $\times$ & 6.69 & 53.02 & 57.68 & 8.64 & 36.75 & 41.17  & 10.01 & 6.92 & 10.75        & 6.91 & 60.46 & 64.76 & 10.14 & 40.59 & 44.92 & 12.73 & 8.36 & 13.18\\
\hline
+DA  & \checkmark  & 8.99 & 19.45 & 23.48 & 9.49 & 13.45 & 17.27 & 10.13 & 4.15 & 8.06           & 10.01 & 30.68 & 34.94 & 11.35 & 17.65 & 22.37  & 12.94 & 7.41 & 12.08 \\
+ADA & \checkmark  & 7.86 & 40.05 & 44.01 & 9.41 & 27.04 & 30.58 & 10.24 & 4.95 & 9.06          & 8.95 & 44.65 & 49.08 & 12.07 & 13.54 & 18.20 & 13.07 & 6.12 & 10.79 \\
+\our & $\times$ & 9.85   & 6.81 & 10.64    & 9.92      & 4.78 & 8.68 & 10.26 & 2.63 & 6.64 & 12.05     & 13.12     & 17.87 &   12.65 & 9.61 & 14.57 & 13.88 & 4.09 & 9.00\\
\rowcolor{tblcolor}+ADA+\our & \checkmark & \textbf{10.10}  & \textbf{6.22} & \textbf{10.09}    & \textbf{10.26}     & \textbf{3.98} & \textbf{7.93} & \textbf{10.31} & \textbf{2.22} & \textbf{6.28} & \textbf{12.70} & \textbf{9.49}      & \textbf{14.23} &   \textbf{12.98} & \textbf{7.02} & \textbf{11.87} & \textbf{13.98} & \textbf{4.02} & \textbf{8.93} 
\\
\bottomline
\end{tabular}
\end{center}
\vspace{-0.4cm}
\end{table*}

\begin{table*}[t]
\begin{center}
\caption{Comparing ImageNet results with varying training data percentages, using our method \vs without it.}\label{tab:results_I64}
\vspace{-0.3cm}
\footnotesize
\setlength{\tabcolsep}{0.115cm}
\renewcommand{\arraystretch}{0.95}
\begin{tabular}{!{\vrule width \boldlinewidth}l|c!{\vrule width \boldlinewidth}ccc|cc!{\vrule width \boldlinewidth}ccc|cc!{\vrule width \boldlinewidth}ccc|cc!{\vrule width \boldlinewidth}}
\topline
\multirow{3}{*}{Method} & \multirow{3}{*}{MA} & \multicolumn{5}{c!{\vrule width \boldlinewidth}}{2.5\% data} & \multicolumn{5}{c!{\vrule width \boldlinewidth}}{5\% data} & \multicolumn{5}{c!{\vrule width \boldlinewidth}}{10\% data}\\
\cline{3-17}
& & \multicolumn{3}{c|}{50$k$ fake imgs} & \multicolumn{2}{c!{\vrule width \boldlinewidth}}{10$k$ fake imgs} & \multicolumn{3}{c|}{50$k$ fake imgs} & \multicolumn{2}{c!{\vrule width \boldlinewidth}}{10$k$ fake imgs} & \multicolumn{3}{c|}{50$k$ fake imgs} & \multicolumn{2}{c!{\vrule width \boldlinewidth}}{10$k$ fake imgs}\\
\cline{3-17}
& & IS$\uparrow$ & tFID$\downarrow$ & vFID$\downarrow$ & IS$\uparrow$ & tFID$\downarrow$ & IS$\uparrow$ & tFID$\downarrow$ & vFID$\downarrow$ & IS$\uparrow$ & tFID$\downarrow$ & IS$\uparrow$ & tFID$\downarrow$ & vFID$\downarrow$ & IS$\uparrow$ & tFID$\downarrow$\\
\middleline
BigGAN          & $\times$  & 8.61 & 101.62 & 100.09 & 8.43 & 103.40 & 6.27 & 90.32 & 88.01 & 6.28 & 93.26  &12.44 & 50.75& 49.84 & 12.17&52.90\\
\hline
+DA             & \checkmark    & 11.07 & 86.07 & 84.48   & 10.82 &  87.30  & 9.15 & 68.61 & 66.85 & 9.01&70.86  &16.30&35.16&34.01 & 15.78&37.76\\
+ADA            & \checkmark    & 7.93 & 67.84 & 66.55 & 7.86 & 70.01 & 11.56&47.56&46.25 & 11.28&50.15 & 14.82&31.75&30.68 & 14.68&34.35\\
+MaskedGAN      & \checkmark    & $-$ & $-$ & $-$     & 12.68 & 38.62 &  $-$ & $-$ & $-$  & 12.85 & 35.70 &  $-$ & $-$ & $-$ & 13.34 & 26.51 \\
+ADA+KDDLGAN    & \checkmark    & $-$ & $-$ & $-$     & 14.65 & 28.79 &  $-$ & $-$ & $-$  & 14.06 & 22.35 &  $-$ & $-$ & $-$ & 14.14 & 20.32 \\

+\our           & $\times$    & 14.68 & 30.66 & 29.32  & 14.25 & 32.93  & 17.34 & 21.13 & 19.95  &16.64 & 23.62 & 20.45 & 14.70 & 13.84 & 19.16 & 17.34\\
\rowcolor{tblcolor}+ADA+\our       & \checkmark  &\textbf{16.57} & \textbf{23.01} & \textbf{21.90} & \textbf{15.70} & \textbf{25.98} & \textbf{19.15} & \textbf{16.14} & \textbf{15.17}  &\textbf{18.17} & \textbf{18.77} & \textbf{22.04} & \textbf{12.91} & \textbf{12.17} & \textbf{21.16} & \textbf{15.83}\\
\bottomline
\end{tabular}
\end{center}
\vspace{-0.4cm}
\end{table*}

\begin{table*}[t!]
\begin{center}
\caption{FID$\downarrow$ on seven few-shot datasets, comparing w/ \vs w/o \our, based on mean and standard deviation from 5 trails.}\label{tab:fastgan}
\vspace{-0.3cm}
\footnotesize
\newcommand{\cs}{\hspace{0.3cm}}
\newcommand{\ccs}{\hspace{0.cm}}
\begin{tabular}{!{\vrule width \boldlinewidth}l|c!{\vrule width \boldlinewidth}ccccccc!{\vrule width \boldlinewidth}}
\topline
\multirow{2}{*}{Method} & \multirow{2}{*}{sec/$k$img} & Shells & Skulls & AnimeFace & BreCaHAD & MessidorSet1 & Pokemon & ArtPainting \\
\cline{3-9}
& & 64 imgs & 97 imgs & 120 imgs & 162 imgs & 400 imgs & 833 imgs & 1000 imgs\\
\middleline
FastGAN \cite{FastGAN} & 34.40 & 138.50$_{\pm\text{3.65}}$ & 97.87$_{\pm\text{1.05}}$ & 54.05$_{\pm\text{0.55}}$ & 63.83$_{\pm\text{1.36}}$ & 38.33$_{\pm\text{4.30}}$ & 45.70$_{\pm\text{1.65}}$ & 43.21$_{\pm\text{0.14}}$ \\
FreGAN \cite{yang2022fregan} & 44.75 & 123.75$_{\pm\text{4.92}}$ & 84.58$_{\pm\text{0.50}}$ & 49.09$_{\pm\text{0.58}}$ & \textbf{57.87}$_{\pm\text{0.55}}$ & 34.61$_{\pm\text{2.48}}$ & 39.09$_{\pm\text{1.35}}$ & 43.14$_{\pm\text{0.69}}$ \\
\middleline
\fastgandbig & \textbf{32.79} & 171.35$_{\pm\text{6.91}}$ & 165.64$_{\pm\text{11.47}}$ & 76.02$_{\pm\text{5.37}}$ & 68.63$_{\pm\text{1.18}}$ & 37.38$_{\pm\text{1.73}}$ & 53.48$_{\pm\text{3.55}}$ & 43.04$_{\pm\text{0.24}}$ \\
\rowcolor{tblcolor}\fastgandbig+\our & 35.94 & \textbf{78.62}$_{\pm\text{1.21}}$ & \textbf{82.47}$_{\pm\text{2.82}}$ & \textbf{46.27}$_{\pm\text{0.36}}$ & 58.98$_{\pm\text{1.59}}$ & \textbf{28.76}$_{\pm\text{1.52}}$ & \textbf{31.94}$_{\pm\text{2.82}}$ & \textbf{38.83}$_{\pm\text{0.49}}$ \\
\bottomline
\end{tabular}
\end{center}
\vspace{-0.5cm}
\end{table*}

\section{Experiments}
We conduct experiments on CIFAR-10/100 \cite{CIFAR} using BigGAN \cite{BigGAN} and OmniGAN \cite{OmniGAN}, as well as on ImageNet \cite{ImageNet} using BigGAN for conditional image generation. We evaluate our method on 5 low-shot datasets \cite{DA}, which include 100-shot Obama/Panda/Grumpy Cat and AnimalFace Dog/Cat \cite{si2011learning}, using StyleGAN2 \cite{StyleGAN2}. Additionally, we assess our method on 7 high-resolution few-shot datasets, including Shells, Skulls, AnimeFace \cite{AnimeFace}, Pokemon, ArtPainting, and two medical datasets BreCaHAD \cite{BreCaHAD}, MessidorSet1 \cite{MessidorSet}, building upon FastGAN \cite{FastGAN}. For comparative purposes, methods involving massive augmentation include DA \cite{DA} and ADA \cite{ADA}, termed ``MA" in \cite{GenCo}, are also included in our evaluation.

\vspace{0.03cm}
\noindent\textbf{Datasets.} CIFAR-10 has $50K/10K$ training/testing images in 10 categories at $32\times32$ resolution, while CIFAR-100 has 100 classes. ImageNet compreises $1.2M/50K$ training/validation images across 1K categories. Following \cite{MaskedGAN, KDDLGAN}, we center-crop and downscale its images to $64\times64$ resolution. The five low-shot datasets include 100-shot Obama/Panda/Grumpy Cat images, along with AnimalFace (160 cats and 389 dogs) images at $256\times256$ resolution. The seven few-shot datasets, Shells, Skulls, AnimeFace, Pokemon, Artpainting, BreCaHAD, MessidorSet1, vary from 64 to 1000 images, each at a high $1024\times1024$ resolution. Following \cite{DA}, we augment all datasets with $x$-flips.

\vspace{0.04cm}
\noindent\textbf{Evaluation metrics.} We generate 50K images for CIFAR-10/100 and ImageNet to calculate Inception Score (IS) \cite{salimans2016improved} and Fr\'{e}chet Inception Distance (FID) \cite{FID}. For these datasets, tFID is calculated by comparing $50K$ generated images against all training images. Additionally, we compute vFID for CIFAR-10/100 and ImageNet between $10K/50K$ fake and real testing/validation images. For the five low-shot and seven few-shot datasets, FID is measured between $5K$ fake images and the full dataset. Following \cite{DA, FakeCLR, DigGAN}, we run five trails for methods employing \our, reporting average results and omitting standard deviations for clarity, as they fall below 1\%. Implementation details and generated images are available  in \S \ref{sm:sec:impl_net} and \S \ref{sm:sec:generated_images}.

\subsection{Comparison with sate-of-the-art methods}
\textbf{Results on CIFAR-10/100 w/ BigGAN/OmniGAN.} Table \ref{tab:results_CIFAR} demonstrates that our method achieves state-of-the-art results on CIFAR-10/100, surpassing even KDDLGAN \cite{KDDLGAN}, which leverages knowledge from CLIP \cite{CLIP}.

\noindent\textbf{Results on ImageNet with BigGAN.}
Maintaining consistency with established benchmarks in \cite{MaskedGAN, KDDLGAN} (using 10K generated images for IS and tFID), Table \ref{tab:results_I64} demonstrates the superiority of \our, outperforming all leading models and underscoring its exceptional performance.

\noindent\textbf{Results on the seven few-shot datasets with FastGAN.}
FastGAN \cite{FastGAN}, known for its memory and time efficiency, yields desirable results on $1024\times1024$ resolution within one-day training on a single GPU. To integrate our method, we swapped large FastGAN discriminator with BigGAN and removed the small discriminator due to  multidimensional output of FastGAN being unsuitable for adjusting our $p$. This new variant, named \fastgandbig, is described in Figure \ref{fig:sm:FastGAN-Dbig} of \S \ref{sm:sec:impl_net}. Table \ref{tab:fastgan} demonstrates the superior performance of \our on seven $1024\times1024$ low-shot datasets.

\noindent\textbf{Results on the five low-shot datasets w/ StyleGAN2.}
Table \ref{tab:low-shot} presents a comparison of \our with other baselines, clearly demonstrating that \our achieves the best results.

\subsection{Experimental analysis}
\textbf{Gradient analysis for centering step.} Figure \ref{fig:similarity} illustrates the mean cosine similarity among pre-activation features in the discriminator and the gradient norm of the feature extractor output \wrt input for OmniGAN, OmniGAN+0C (using Eq. \ref{eq:centering} centering), and OmniGAN+A0C (adaptive interpolation of centered and uncentered features). The near-zero mean cosine similarity in OmniGAN+0C and OmniGAN+A0C corroborates Theorem \ref{theorem:bn_centering}, indicating that centering leads to feature difference in later layers and amplifying the gradient effect, as seen in Figure \ref{subfig:gradient_0c_C10}. This observation supports the decision to modify the centering step.

\noindent\textbf{Gradient analysis for scaling step.}
Figure \ref{subfig:grad_C10} shows gradient norms of the discriminator output \wrt the input and effective rank (eRank) \cite{roy2007effective} for various models. The CHAIN$_{-LC}$ variant (\our w/o Lipschitz constraint) exhibits gradient explosion, confirming Theorem \ref{theorem:bn_scaling}. While CHAIN$_{+0C}$ avoids gradient explosion, its centering step causes abrupt feedback changes to the generator, leading to the dimensional collapse \cite{thanh-tung2018improving, mescheder2018training, DRAGAN, ACGAN}, evidenced by rank deficiencies in Figure \ref{subfig:rank_C10}. In contrast, \our maintains smaller gradient than  OmniGAN, aligning with the analysis in Theorem \ref{theorem:chain_grad} \wrt reducing gradient in latent features.

\noindent\textbf{Generalization analysis.} Figures \ref{subfig:D_out_C10:grad} and \ref{subfig:D_out_C100:grad} show that \our achieves smaller gradient norm of discriminator output \wrt weight, supporting the assertion of Theorem \ref{theorem:chain_grad} on reducing weight gradient. This leads to a lower generalization error, as per Prop. \ref{prop:gen_sharpness} and Lemma \ref{lemma:gen_error}, evidenced in Figures \ref{subfig:D_out_C10_Dx} and \ref{subfig:D_out_C100_Dx}. Here, compared to the baseline, \our maintains a smaller discrepancy in discriminator output between real and test images, as well as discrepancy between real and fake images, indicating the effectiveness of \our in improving GAN generalization.

\begin{table}[t!]
\begin{center}
\caption{FID$\downarrow$ of unconditional image generation with StyleGAN2 on five low-shot datasets. $^\dag$ marks a generator pre-trained on full FFHQ \cite{StyleGAN} dataset, $^\ddagger$ signifies a pre-trained CLIP \cite{CLIP} model. ``MA" means Massive Augmentation, ``PT" refers to Pretrained.}\label{tab:low-shot}
\vspace{-0.2cm}
\footnotesize
\setlength{\tabcolsep}{0.08cm}
\begin{tabular}{!{\vrule width \boldlinewidth}l|c|c!{\vrule width \boldlinewidth}ccc!{\vrule width \boldlinewidth}cc!{\vrule width \boldlinewidth}}
\topline
\multirow{2}{*}{Method} & \multirow{2}{*}{MA} & \multirow{2}{*}{PT}  & \multicolumn{3}{c!{\vrule width \boldlinewidth}}{100-shot} & \multicolumn{2}{c!{\vrule width \boldlinewidth}}{Animal Face} \\
\cline{4-8}
& & &  Obama & GrumpyCat & Panda & Cat & Dog\\
\middleline
StyleGAN2       & $\times$ & $\times$ & 80.20 & 48.90& 34.27& 71.71 & 131.90\\
\rowcolor{tblcolor}+\our           & $\times$ & $\times$ & \textbf{28.72}
                         & \textbf{27.21}
                         & \textbf{9.51}  
                         & \textbf{38.93} 
                         & \textbf{53.27} 
                         \\
\middleline
AdvAug \cite{chen2021data} & \checkmark & $\times$ & 52.86 & 31.02 & 14.75 & 47.40 & 68.28\\
ADA             & \checkmark & $\times$ & 45.69 & 26.62 & 12.90 & 40.77 & 56.83 \\
DA              & \checkmark & $\times$ & 46.87 & 27.08 & 12.06 & 42.44 & 58.85 \\
ADA+DigGAN      & \checkmark & $\times$ & 41.34 & 26.75 & $-$  & 37.61 & 59.00 \\
LeCam           & \checkmark & $\times$ & 33.16 & 24.93 & 10.16 & 34.18 & 54.88 \\
GenCo           & \checkmark & $\times$ & 32.21 & 17.79 & 9.49  & 30.89 & 49.63 \\
InsGen          & \checkmark & $\times$ & 32.42 & 22.01 & 9.85  & 33.01 & 44.93 \\
MaskedGAN       & \checkmark & $\times$ & 33.78 & 20.06 & 8.93  & $-$ & $-$ \\
FakeCLR         & \checkmark & $\times$ & 26.95 & 19.56 & 8.42 & 26.34 & 42.02 \\
TransferGAN $^\dag$     & \checkmark & \checkmark & 39.85 & 29.77 & 17.12 & 49.10 & 65.57\\
KDDLGAN $^\ddagger$    & \checkmark & \checkmark & 29.38  & 19.65 & 8.41 & 31.89 & 50.22 \\
AugSelf \cite{hou2024augmentation} & \checkmark & $\times$ & 26.00 & 19.81 & 8.36 & 30.53 & 48.19 \\
\middleline
ADA+\our & \checkmark & $\times$ & \textbf{20.94} & 17.61 & 7.50 & 19.74 & 39.10 \\
\rowcolor{tblcolor} DA+\our & \checkmark & $\times$ & 22.87 & \textbf{17.57} & \textbf{6.93} & \textbf{19.58} & \textbf{30.88} \\
\bottomline
\end{tabular}
\end{center}
\vspace{-0.3cm}
\end{table}

\begin{figure}[t!]
\begin{center}
\begin{subfigure}{0.493\linewidth}
\includegraphics[width=\linewidth]{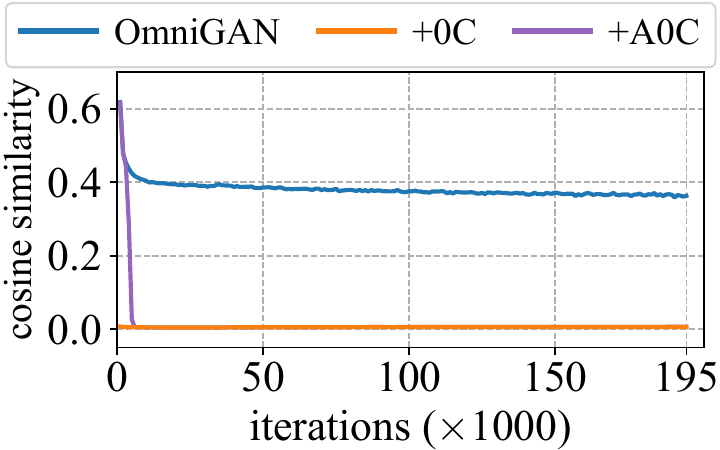}
\caption{Mean cosine similarity.}\label{subfig:similarity_C10}
\end{subfigure}
\begin{subfigure}{0.493\linewidth}
\includegraphics[width=\linewidth]{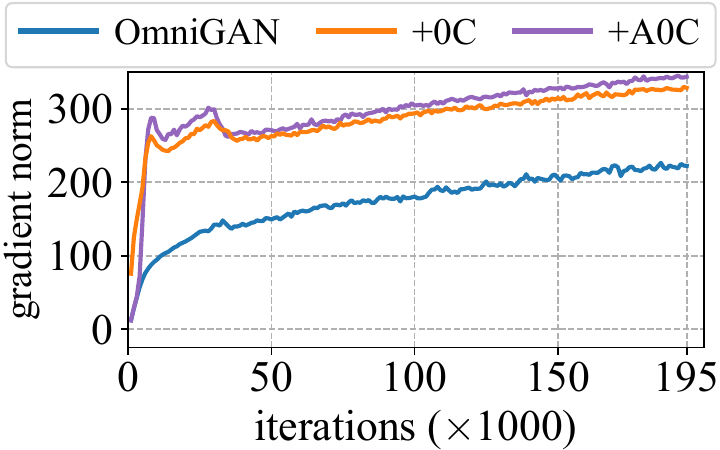}
\caption{Gradient norm.}\label{subfig:gradient_0c_C10}
\end{subfigure}
\vspace{-0.6cm}
\caption{(a) Mean cosine similarity of discriminator pre-activation features, and (b) gradient norm of the feature extractor \wrt the input are evaluated for OmniGAN, OmniGAN+0C (using the centering step in Eq. \ref{eq:centering}), and OmniGAN+A0C (adaptive interpolation between centered and uncentered features). Evaluation conducted on 10\% CIFAR-10 data with OmniGAN ($d=256$).}
\label{fig:similarity}
\end{center}
\vspace{-0.7cm}
\end{figure}

\begin{figure}[t!]
    \centering
    \includegraphics[width=1\linewidth]{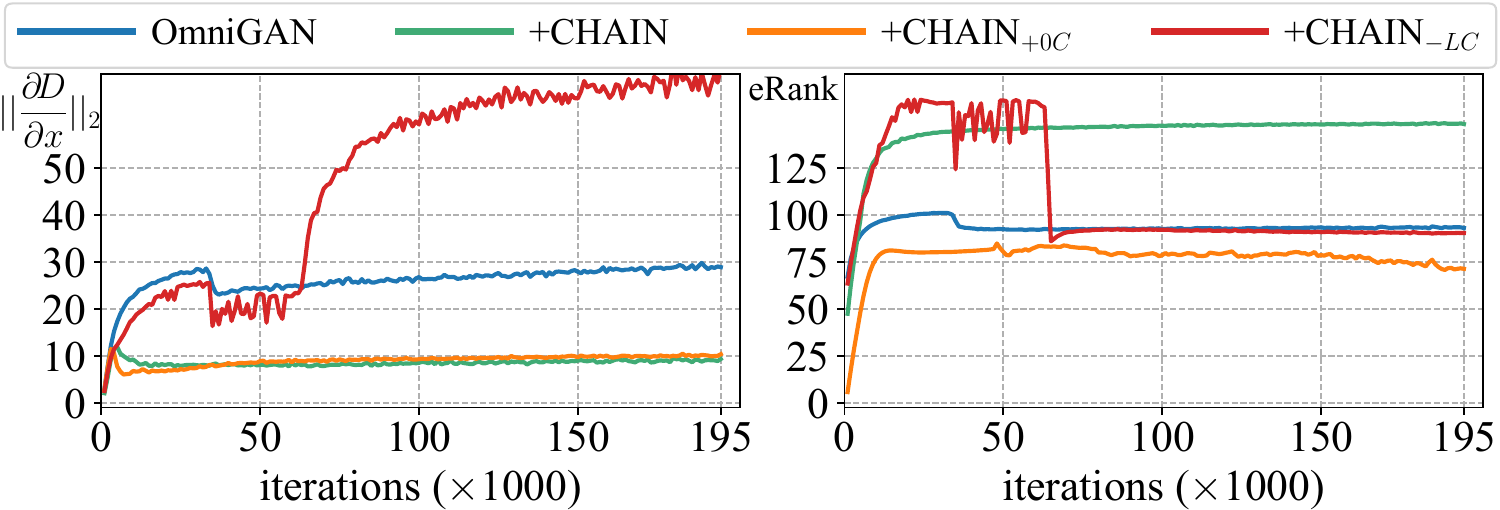}
    \begin{subfigure}{0.494\linewidth}
        \caption{Gradient norm.}\label{subfig:grad_C10}
    \end{subfigure}
    \begin{subfigure}{0.494\linewidth}
        \caption{Effective rank.}\label{subfig:rank_C10}
    \end{subfigure}
    \vspace{-0.6cm}
    \caption{(a) Gradient norm of discriminator output \wrt input during training, and (b) effective rank \cite{roy2007effective} of the pre-activation features in discriminator, are evaluated on 10\% CIFAR-10 data with OmniGAN ($d\!\!=\!\!256$). CHAIN$_{+0C}$: \our w/ the centering step. CHAIN$_{-LC}$: \our w/o the Lipschitzness constraint.}
    \label{fig:gradient_rank_C10}
    \vspace{-0.2cm}
\end{figure}

\begin{figure}[t!]
    \centering
    \includegraphics[width=1\linewidth]{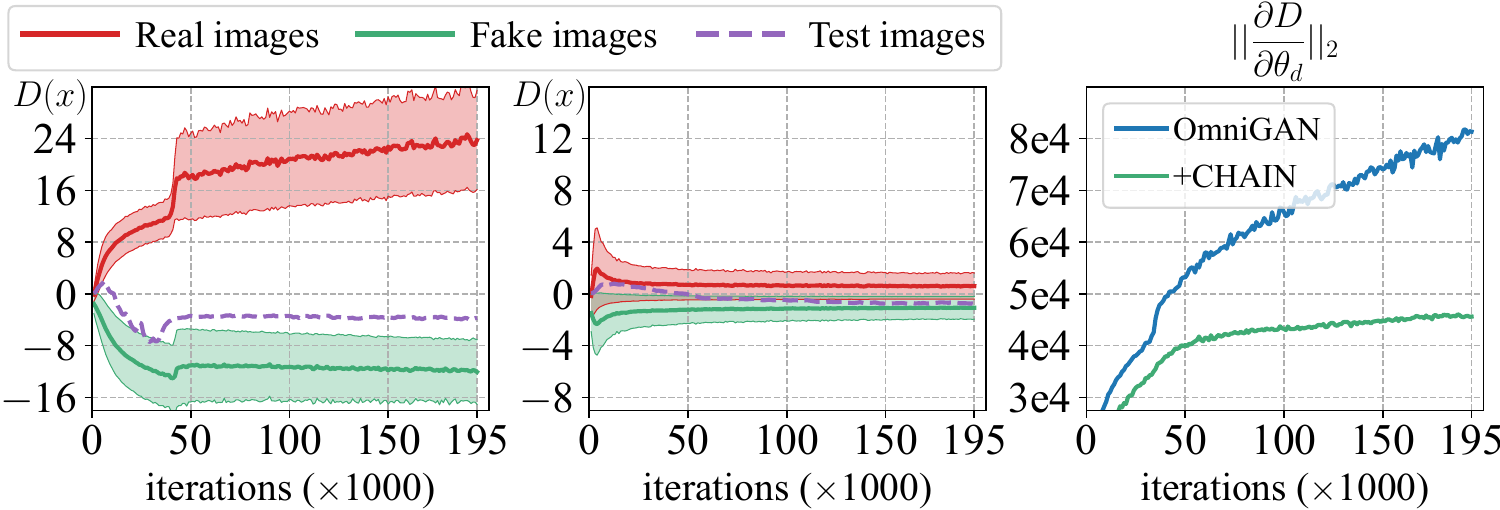}
    \begin{subfigure}{0.325\linewidth}
        \caption{OmniGAN}
    \end{subfigure}
    \begin{subfigure}{0.325\linewidth}
        \caption{OmniGAN+\our} \label{subfig:D_out_C10_Dx}
    \end{subfigure}
    \begin{subfigure}{0.325\linewidth}
        \caption{$\lVert\frac{\partial D}{\partial \vtheta_d}\rVert_2$}\label{subfig:D_out_C10:grad}
    \end{subfigure}
    \vspace{-0.6cm}
    \caption{The discriminator output \wrt real, fake and test images using (a) OmniGAN, (b) OmniGAN+\our, and (c) the gradient norm of the discriminator output \wrt discriminator weights on 10\% CIFAR-10 using OmniGAN ($d=256$). Note the $y$-axis in (b) is scaled for clearer visualization.}
    \label{fig:D_out_C10}
    \vspace{-0.3cm}
\end{figure}

\begin{figure}[t!]
    \centering
    \includegraphics[width=1\linewidth]{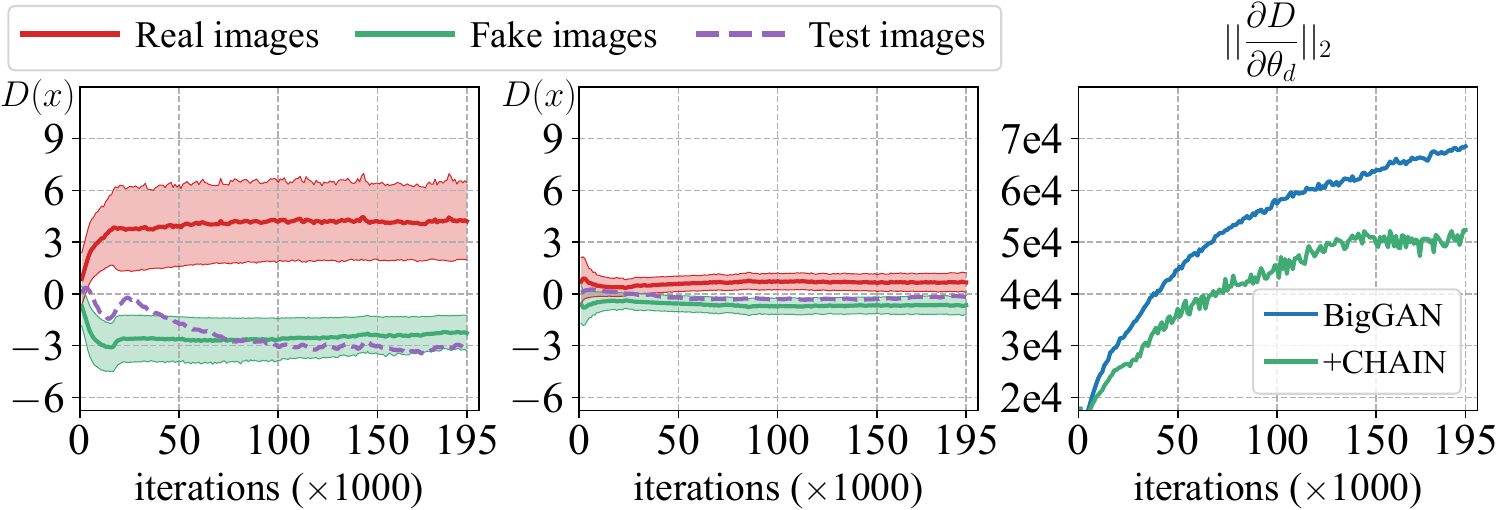}
    \begin{subfigure}{0.325\linewidth}
        \caption{BigGAN}
    \end{subfigure}
    \begin{subfigure}{0.325\linewidth}
        \caption{BigGAN+\our}\label{subfig:D_out_C100_Dx}
    \end{subfigure}
    \begin{subfigure}{0.325\linewidth}
        \caption{$\lVert\frac{\partial D}{\partial \vtheta_d}\rVert_2$}\label{subfig:D_out_C100:grad}
    \end{subfigure}
    \vspace{-0.6cm}
    \caption{The discriminator output \wrt real, fake and test images of (a) BigGAN, (b) BigGAN+\our, along with (c) the gradient norm of the discriminator output \wrt discriminator weights on 10\% CIFAR-100 with BigGAN ($d=256$).}
    \label{fig:D_out_C100}
    \vspace{-0.4cm}
\end{figure}

\subsection{Ablation studies}
\textbf{Ablation for \our design.} Table \ref{tab:ablation} provides quantitative evidence supporting the design of our method. The inferior results of CHAIN$_{-\text{0MR}}$ and CHAIN$_{-\text{ARMS}}$ highlight the significance of the 0MR and ARMS modules. Poorer performance of CHAIN$_{+\text{0C}}$ underscores the need to omit the centering step. The notably worse outcomes of CHAIN$_{-\text{LC}}$ emphasize the importance of the Lipschitzness constraint. CHAIN$_\text{batch}$ underperforming suggests the advantage of using running cumulative statistics. The suboptimal performance of CHAIN$_\text{Dtm.}$ validate the stochastic $\mM$ design (Eq. \ref{eq:brmsn}), while marginally poorer results of CHAIN$_{+\text{0MR}_g}$ indicate limited benefits of applying 0MR in generator training.

\vspace{0.1cm}
\noindent\textbf{Ablation of each factor.} Figure \ref{fig:different_hyper_parameters} explores the impact of applying \our at different points and varying the hyperparameters $\lambda$, $\tau$. In Figure \ref{subfig:places}, optimal performance is achieved by placing \our after all convolutional layers. Figure \ref{subfig:blocks} demonstrates that employing our approach across all blocks yields the best results. Figure \ref{subfig:lambdas} shows that varying $\lambda$ between 2 to 50 does not significantly affect performance, indicating the robustness of \our to $\lambda$. Lastly, Figure \ref{subfig:taus} suggests that setting $\tau$ to be 0.5 is preferable.
\begin{table}[t!]
\vspace{-0.3cm}
\begin{center}
\caption{Ablation studies. 0C: using centering. A0C: adaptively interpolating centered and uncentered features. CHAIN$_{-\text{0MR}}$: \our w/o  0-mean regularization (0MR, Eq. \ref{eq:0mr}). CHAIN$_{-\text{ARMS}}$: \our w/o the adaptive root mean square normalization (ARMS, Eq. \ref{eq:brmsn}). CHAIN$_{+\text{0C}}$: \our w/ centering. CHAIN$_{-\text{LC}}$: \our w/o the Lipschitzness constraint. CHAIN$_\text{batch}$: replacing the cumulative with batch statistics. CHAIN$_\text{Dtm.}$: replacing the stochastic $\mM$ in Eq. \ref{eq:brmsn} with deterministic $p$. CHAIN$_{+\text{0MR}_g}$: Applying $\ell^\text{0MR}$ in generator training. CHAIN$_{+\text{Aff.}}$: applying learnable affine transformation. ADrop: adaptive dropout. }\label{tab:ablation}
\vspace{-0.3cm}
\footnotesize
\setlength{\tabcolsep}{0.175cm}
\begin{tabular}{!{\vrule width \boldlinewidth}l!{\vrule width \boldlinewidth}ccc!{\vrule width \boldlinewidth}ccc!{\vrule width \boldlinewidth}}
\topline
\multirow{3}{*}{Method} & \multicolumn{3}{c!{\vrule width \boldlinewidth}}{10\% CIFAR-10} & \multicolumn{3}{c!{\vrule width \boldlinewidth}}{10\% CIFAR-100} \\
& \multicolumn{3}{c!{\vrule width \boldlinewidth}}{OmniGAN ($d\!=\!256$)} & \multicolumn{3}{c!{\vrule width \boldlinewidth}}{BigGAN($d\!=\!256$)} \\
\cline{2-7}
& IS$\uparrow$ & tFID$\downarrow$ & vFID$\downarrow$ & IS$\uparrow$ & tFID$\downarrow$ & vFID$\downarrow$ \\
\middleline
Baseline            & 8.49 & 22.24 & 26.33 &        7.58 & 50.79 & 55.04 \\
\hline
\ \ w/ 0C           & 8.93 & 31.82 & 35.57 &        7.89 & 37.47 & 42.27 \\
\ \ w/ A0C          & 8.83 & 26.45 & 30.30 &        8.47 & 36.86 & 41.80 \\
\middleline
\rowcolor{tblcolor}\our    & 9.52 & \textbf{8.27} & \textbf{12.06} &        10.04 & \textbf{13.13} & \textbf{18.00} \\
\hline
CHAIN$_{-\text{0MR}}$      & 9.37 & 9.20 & 13.05 &         9.71  & 24.26 & 29.20 \\
CHAIN$_{-\text{ARMS}}$     & 9.33 & 12.87 & 16.87 &        9.09  & 24.14 & 29.59 \\
CHAIN$_{+\text{0C}}$       & 9.43 & 8.99 & 12.71 &         8.84  & 22.85 & 27.91 \\
CHAIN$_{-\text{LC}}$       & 8.68 & 22.14 & 26.37 &        8.05  & 30.43 & 35.15 \\
CHAIN$_\text{batch}$     & 9.42 & 8.51 & 12.32 &         9.85  & 14.49 & 19.18 \\
CHAIN$_\text{Dtm.}$      & \textbf{9.59} & 9.44 & 13.21 &         9.76  & 15.07 & 19.85 \\
CHAIN$_{+\text{0MR}_g}$  & 9.37 & 8.42 & 12.25 &         \textbf{10.99} & 17.09 & 22.06 \\
CHAIN$_{+\text{Aff.}}$  & 9.45 & 8.49 & 12.24 &         10.02 & 14.19 & 19.07 \\
ADrop                   & 8.72 & 14.76 & 18.48 &        9.04 & 29.05 & 34.01 \\
\bottomline
\end{tabular}
\end{center}
\vspace{-0.3cm}
\end{table}

\begin{figure}[t!]
\begin{center}
\begin{subfigure}{0.49\linewidth}
\includegraphics[width=\linewidth]{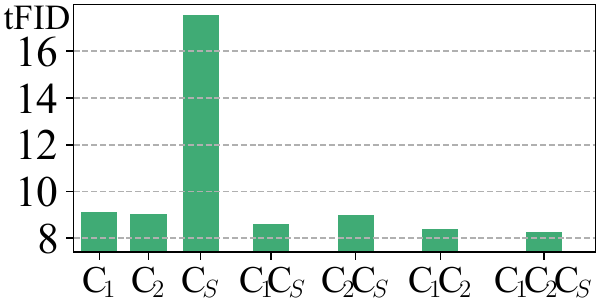}
\caption{$c\in\{C_1, C_2, C_S\}$}
\label{subfig:places}
\end{subfigure}
\begin{subfigure}{0.49\linewidth}
\includegraphics[width=\linewidth]{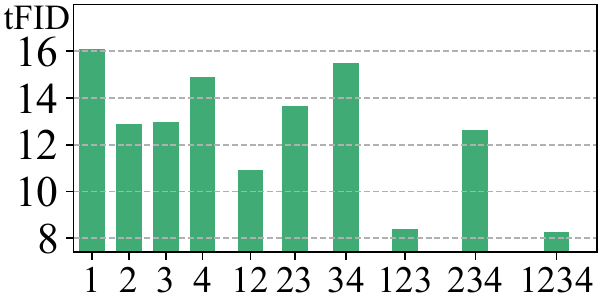}
\caption{$l\in\{1,\cdots,L\}$}
\label{subfig:blocks}
\end{subfigure}
\vspace{0.2cm}
\begin{subfigure}{0.49\linewidth}
\includegraphics[width=\linewidth]{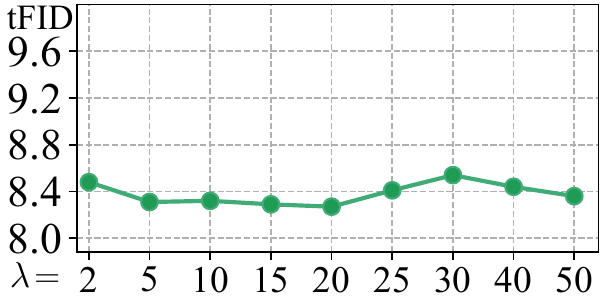}
\caption{$\lambda$}
\label{subfig:lambdas}
\end{subfigure}
\begin{subfigure}{0.49\linewidth}
\includegraphics[width=\linewidth]{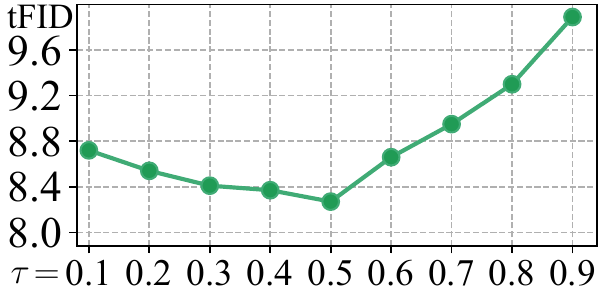}
\caption{$\tau$}\label{subfig:taus}
\end{subfigure}
\vspace{-0.5cm}
\caption{tFID $\downarrow$ under different factors. Ablation studies on 10\% CIFAR-10 with OmniGAN ($d\!=\!256)$ \wrt different conv. configurations, different blocks, $\lambda$ and $\tau$ for \our.}
\label{fig:different_hyper_parameters}
\end{center}
\vspace{-0.8cm}
\end{figure}

\vspace{0.1cm}
\noindent\textbf{Comparison with other variants.} We compare \our against other normalization techniques such as BN, IN, LN, GN, and BN w/ Lipschitzness constraint (BN$_{+\text{LC}}$), methods preventing discriminator overfitting such as DA, ADA, LeCam, and gradient penalizations for improving generalization. Table \ref{tab:variants} details these comparisons. For GN, we optimized group number ($n_g$) for CIFAR-10 ($n_g\!=\!32$) and CIFAR-100 ($n_g\!=\!16$). Implementations for AGP$_{\text{weight}}$ and  AGP$_{\text{input}}$ are explained in \S \ref{sm:subsec:impl_variants}. The results in Table \ref{tab:variants} show \our outperforms other methods, with AGP$_{\text{weight}}$ also yielding competitive results, supporting Prop. \ref{prop:gen_sharpness} about weight gradient reduction enhancing generalization. Furthermore, Figure \ref{fig:different_sizes} indicates that \our benefits from increased network width, unlike other models that deteriorate with wider networks, confirming the superiority of \our. 

\vspace{0.1cm}
\noindent\textbf{More analyses.} \S \ref{sm:sec:add_experiment} compares leading methods, analyzes gradients on CIFAR-100 w/ BigGAN, evaluates eRank against AGP$_{\text{weight}}$, and examines feature norm. \our gains significant improvements with mild extra load (\S \ref{sm:sec:overhead}).

\begin{table}[t!]
\vspace{-0.3cm}
\centering
\footnotesize
\setlength{\tabcolsep}{0.208cm}
\caption{Ablation studies. BN$_{+\text{LC}}$: BN w/ Lipschitz constraint. AGP$_{\text{weight}}$: adaptive gradient penalty \wrt weights. AGP$_{\text{input}}$: adaptive gradient penalty \wrt inputs. LeCam fails to converge on OmniGAN due to its multi-dimensional output design.}\label{tab:variants}
\vspace{-0.3cm}
\begin{tabular}{!{\vrule width \boldlinewidth}l!{\vrule width \boldlinewidth}ccc!{\vrule width \boldlinewidth}ccc!{\vrule width \boldlinewidth}}
\topline
\multirow{3}{*}{Method} & \multicolumn{3}{c!{\vrule width \boldlinewidth}}{10\% CIFAR-10} & \multicolumn{3}{c!{\vrule width \boldlinewidth}}{10\% CIFAR-100}\\
& \multicolumn{3}{c!{\vrule width \boldlinewidth}}{OmniGAN($d\!\!=\!\!256$)} & \multicolumn{3}{c!{\vrule width \boldlinewidth}}{BigGAN($d\!\!=\!\!256$)}\\
\cline{2-7}
& IS$\uparrow$ & tFID$\downarrow$ & vFID$\downarrow$ & IS$\uparrow$ & tFID$\downarrow$ & vFID$\downarrow$\\
\middleline
Baseline            & 8.49 & 22.24   & 26.33    & 7.58  & 50.79     & 55.04\\
\hline
BN             & 7.56  & 37.37  & 41.52    & 7.07  & 55.83     & 60.46\\
BN$_{+\text{LC}}$         &  9.40  & 14.32  & 17.75    & 9.15  & 25.87     & 30.83\\
IN                  & 6.71  & 53.80  & 57.76    & 5.13  & 83.06     & 87.40\\
LN                  & 6.23  & 101.97 & 105.58   & 9.04  & 26.25     & 31.22\\
GN                  & 7.38  & 49.39  & 53.46    & 8.80  & 31.40     & 36.53\\
\hline
DA                  & 8.84  & 12.90  & 16.67    & 8.86  & 27.22     & 31.80\\
ADA                 & \textbf{9.67}  & 13.86  & 17.70   & 8.96   & 20.09 & 24.90  \\
LeCam               & $-$   & $-$    & $-$      & 8.30  & 31.52     & 36.26\\
\hline
AGP$_{\text{input}}$ & 8.75 & 14.78 & 18.65 & 8.48 & 24.95 & 29.58 \\
AGP$_{\text{weight}}$ & 9.42 & 11.86 & 15.78 & 9.24 & 18.52 & 23.28\\
\hline
\rowcolor{tblcolor}\our                & 9.52  & \textbf{8.27}   & \textbf{12.06}   & \textbf{10.04} & \textbf{13.13} & \textbf{18.00} \\
\bottomline
\end{tabular}
\end{table}

\begin{figure}[t!]
\includegraphics[width=\linewidth]{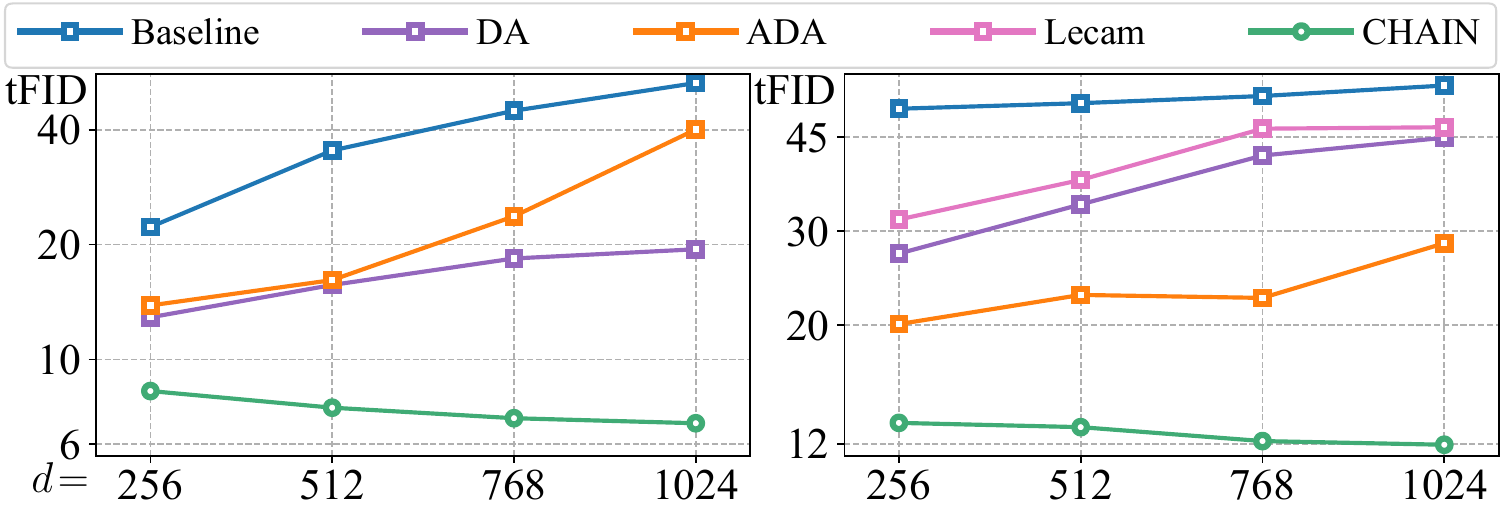}
\begin{subfigure}{0.495\linewidth}
        \caption{10\% CIFAR-10 w/ OmniGAN}
    \end{subfigure}
    \begin{subfigure}{0.495\linewidth}
        \caption{10\% CIFAR-100 BigGAN}
    \end{subfigure}
    \vspace{-0.6cm}
\caption{tFID for different methods with varying feature sizes $d$.}\label{fig:different_sizes}
\vspace{-0.3cm}
\end{figure}

\section{Conclusions}
Our method, LipsCHitz contuity constrAIned Normalization (\our), harnesses the generalization benefits of BN to counter discriminator overfitting in GAN training. We refine standard BN by implementing the zero-mean regularization and the Lipschitzness constraint, effectively reducing gradient norms in latent features and discriminator weights. This approach not only stabilizes GAN training but also boosts generalization. Proven in theory and practice, \our excels across diverse backbones and datasets, consistently surpassing existing methods and effectively addressing discriminator overfitting in GANs. 

\vspace{0.1cm}
\noindent\textbf{Acknowledgements.} We thank Moyang Liu, Fei Wu, Melody Ip, and Kanghong Shi for their discussions and encouragement that significantly shaped this work. PK is funded by CSIRO’s Science Digital.
\newpage
{
    \small
    \bibliographystyle{ieeenat_fullname}
    \bibliography{references}
}
\newpage
\input{supp}
\end{document}

%% file: preamble.tex
\usepackage[dvipsnames]{xcolor}


%% file: mathcommands.tex
\usepackage{multirow}
\usepackage{booktabs}
\usepackage{xspace}
\usepackage{subcaption}
\usepackage{graphicx}
\usepackage{amssymb}
\usepackage{textcomp}
\usepackage{tikz}
\usepackage{makecell}
\usepackage{array}
\usepackage{xcolor}
\usepackage{mathabx}
\usepackage[ruled,vlined]{algorithm2e}

\definecolor{commentcolor}{RGB}{110,154,155}   
\newcommand{\PyComment}[1]{\ttfamily\textcolor{commentcolor}{\# #1}}  
\newcommand{\PyCode}[1]{\ttfamily\textcolor{black}{#1}} 

\makeatletter
\DeclareRobustCommand\onedot{\futurelet\@let@token\mathbfv@onedotaux}
\def\mathbfv@onedotaux{\ifx\@let@token.\else.\null\fi\xspace}
\def\eg{\emph{e.g}\onedot} 
\def\ie{\emph{i.e}\onedot} 
 
 \def\vs{\emph{vs}\onedot}
\def\wrt{w.r.t\onedot}

\def\etal{\emph{et al}\onedot}
\definecolor{fakecolor}{rgb}{0.25, 0.67, 0.46}
\definecolor{realcolor}{rgb}{0.84, 0.15, 0.15}
\definecolor{unseencolor}{rgb}{0.54, 0.41, 0.72}
\newtheorem{theorem}{Theorem}[section]
\newtheorem{prop}{Proposition}[section]
\newtheorem{lemma}{Lemma}[section]
\newtheorem{smlemma}{Lemma}[section]

\newcommand{\empmu}{\hat{\mu}_n}
\newcommand{\empnu}{\hat{\nu}_n}
\newcommand{\popnu}{\nu_n}

\newcommand{\emprho}{\hat{\rho}}

\newcommand{\optnu}{\nu_n^*}
\newcommand{\mH}{\boldsymbol{H}}
\newcommand{\vnabla}{\boldsymbol{\nabla}}
\newcommand{\ganerr}{\epsilon_\text{gan}}
\newcommand{\ganerrn}{\epsilon_\text{gan}^\text{nn}}
\newcommand{\calG}{\mathcal{G}}
\newcommand{\calH}{\mathcal{H}}
\newcommand{\calB}{\mathcal{B}}
\newcommand{\calHn}{\mathcal{H}_{\text{nn}}}

\newcommand{\calF}{\mathcal{F}}
\newcommand{\calM}{\mathcal{M}}
\newcommand{\calN}{\mathcal{N}}
\newcommand{\calL}{\mathcal{L}}

\newcommand{\calD}{\mathcal{D}}
\newcommand{\calX}{\mathcal{X}}

\newcommand{\bbE}{\mathbb{E}}
\newcommand{\bbR}{\mathbb{R}}
\newcommand{\vxr}{\boldsymbol{x}}
\newcommand{\vx}{\boldsymbol{x}}
\newcommand{\vz}{\boldsymbol{z}}
\newcommand{\va}{\boldsymbol{a}}

\newcommand{\vy}{\boldsymbol{y}}

\newcommand{\ysbc}{\widecheck{y}_c^{(b)}}
\newcommand{\ysic}{\widecheck{y}_c^{(i)}}

\newcommand{\ybc}{y_c^{(b)}}

\newcommand{\vv}{\boldsymbol{v}}

\newcommand{\vxf}{\boldsymbol{\Tilde{x}}}
\newcommand{\vtheta}{\boldsymbol{\theta}}
\newcommand{\vzero}{\boldsymbol{0}}
\newcommand{\vvarepsilon}{\boldsymbol{\varepsilon}}
\newcommand{\mTheta}{\boldsymbol{\Theta}}
\newcommand{\vmu}{\boldsymbol{\mu}}
\newcommand{\vsigma}{\boldsymbol{\sigma}}
\newcommand{\mI}{\boldsymbol{I}}
\newcommand{\mW}{\boldsymbol{W}}
\newcommand{\mA}{\boldsymbol{A}}
\newcommand{\mY}{\boldsymbol{Y}}
\newcommand{\mYc}{\overset{\raisebox{-0.5ex}{\scriptsize $c$}}{\mY}}
\newcommand{\mYs}{\overset{\raisebox{-0.5ex}{\scriptsize $s$}}{\mY}}
\newcommand{\mYo}{\dot{\mY}}

\newcommand{\vmuy}{\vmu}
\newcommand{\muyc}{\mu_c}
\newcommand{\psiyc}{\psi_c}
\newcommand{\vsigmay}{\vsigma}
\newcommand{\vpsiy}{\vpsi}
\newcommand{\psimin}{\psi_{\text{min}}}

\newcommand{\mLambda}{\boldsymbol{\Lambda}}
\newcommand{\mM}{\boldsymbol{M}}

\newcommand{\diag}{\text{diag}}
\newcommand{\lc}{\text{lc}}
\newcommand{\vyc}{\overset{\raisebox{-0.5ex}{\scriptsize $c$}}{\vy}}
\newcommand{\vpsi}{\boldsymbol{\psi}}
\newcommand{\sigmamin}{\sigma_{\text{min}}}

\newcommand{\proof}{\noindent\textbf{\textit{Proof:\quad}}}
\newcommand{\our}{CHAIN\xspace}
\newcommand{\gradr}{\vnabla_{\vtheta_d}}
\newcommand{\gradf}{\widetilde{\vnabla}_{\vtheta_d}}
\newcommand{\levr}{\lambda_\text{max}^{\mH}}
\newcommand{\levf}{\lambda_{\text{max}}^{\widetilde{\mH}}}
\newcommand{\hessr}{\mH_{\vtheta_d}}
\newcommand{\hessf}{\widetilde{\mH}_{\vtheta_d}}
\newcommand{\Deltaybc}{\Delta y_c^{(b)}}
\newcommand{\Deltayobc}{\Delta \dot{y}_c^{(b)}}
\newcommand{\Deltayoic}{\Delta \dot{y}_c^{(i)}}

\newcommand{\Deltavych}{\Delta\vy_c}
\newcommand{\Deltavychj}{\Delta\vy_c}
\newcommand{\Deltavyoch}{\Delta\dot{\vy}_c}

\newcommand{\vysch}{\widecheck{\vy}_c}

\newcommand{\lev}{\lambda_\text{max}}
\newcommand{\vachc}{\va_c}
\newcommand{\vw}{\boldsymbol{w}}

\newcommand{\Deltavwc}{\Delta\vw_c}

\newlength{\boldlinewidth}
\newcommand{\topline}{\Xhline{1.3pt}}
\newcommand{\bottomline}{\Xhline{1.3pt}}
\newcommand{\middleline}{\Xhline{1pt}}
\newcommand{\chainsymb}{${\color{fakecolor}{\bigcirc}}\!\!\!{\color{unseencolor}{\bigcirc}}$}
\newcommand{\fastgandbig}{FastGAN$^{_{_-}}\!D_\text{big}$}

\definecolor{commentcolor}{HTML}{808080}
\definecolor{keywordcolor}{HTML}{D73A49}
\definecolor{functioncolor}{HTML}{6C3CAC}
\definecolor{operatorcolor}{HTML}{b31E28}
\definecolor{variablecolor}{HTML}{E16227}
\definecolor{numbercolor}{HTML}{005CC5}
\newcommand{\pyfun}[1]{\textcolor{functioncolor}{#1}}
\newcommand{\pyvar}[1]{\textcolor{variablecolor}{#1}}
\newcommand{\pyop}[1]{\textcolor{operatorcolor}{#1}}
\newcommand{\pynum}[1]{\textcolor{numbercolor}{#1}}
\newcommand{\pykey}[1]{\textcolor{keywordcolor}{#1}}

\newcommand{\circled}[1]{
  \tikz[baseline=(char.base)]{
    \node[shape=circle,draw,inner sep=0.1pt] (char) {#1};
  }
}

%% file: supp.tex
\newcommand{\setappendixtitle}[1]{\def\@appendixtitle{#1}}
\newcommand{\setappendixauthor}[1]{\def\@appendixauthor{#1}}

\def\maketitleappendix{
    \nolinenumbers
   \null
   \iftoggle{cvprrebuttal}{\vspace*{-.3in}}{\vskip .375in}
   \begin{center}
      \iftoggle{cvprrebuttal}{{\large \bf \@appendixtitle \par}}{{\Large {\bf \@appendixtitle} (\textit{Supplementary Material}) \par}}
      \iftoggle{cvprrebuttal}{\vspace*{-22pt}}{\vspace*{24pt}}{
        \large
        \lineskip .5em
        \begin{tabular}[t]{c}
          \iftoggle{cvprfinal}{
            \@appendixauthor
          }{
            \iftoggle{cvprrebuttal}{}{
              Anonymous \confName~submission\\
              \vspace*{1pt}\\
              Paper ID \paperID
            }
          }
        \end{tabular}
        \par
      }
      \vskip .5em
      \vspace*{12pt}
   \end{center}
}

\appendix
\onecolumn

\setappendixtitle{\chainsymb CHAIN: Enhancing Generalization in Data-Efficient GANs via lips\textit{CH}itz continuity constr\textit{AI}ned \textit{N}ormalization}

\setappendixauthor{%
Yao Ni$^{\dagger}$ \quad Piotr Koniusz$^{*,\S,\dagger}$\\
$^{\dagger}$The Australian National University \quad $^\S$Data61$\!${\color{red}\heart}CSIRO \\
\vspace{-0.2cm}
{\tt\small $^{\dagger}$firstname.lastname@anu.edu.au}
}
\maketitleappendix

The supplementary material contains notations (\S \ref{sm:sec:notation}), theoretical proofs (\S \ref{sm:sec:proof}), an explanation for stochastic $\mM$ design (\S \ref{sm:sec:stochastic_M}), implementation guidelines (\S \ref{sm:sec:impl}), extra experimental results (\S \ref{sm:sec:add_experiment}), training overhead (\S \ref{sm:sec:overhead}), and examples of generated images (\S \ref{sm:sec:generated_images}).

\section{Notations}
\label{sm:sec:notation}
Below, we explain the notations used in this work.

\noindent\textbf{Scalars}: Represented by lowercase letters (\eg, $m$, $n$, $p$).

\noindent\textbf{Vectors}: Bold lowercase letters (\eg, $\vx$, $\vz$, $\vmu$).

\noindent\textbf{Matrices}: Bold uppercase letters (\eg, $\mW$, $\mM$, $\mH$).

\noindent\textbf{Functions}: Letters followed by brackets (\eg, $\phi(\cdot)$, $h(\cdot)$, $\diag(\cdot)$).

\noindent\textbf{Function sets}: Calligraphic uppercase letters are used (\eg, $\calH$, $\calG$, $\calF$). But note $\calB$ specifically denotes the Bernoulli distribution.

\noindent\textbf{Probability measures}: Denoted by letters $\mu$, $\nu$, $\pi$, $\emprho$ and $p_z$.

\noindent\textbf{Expectation}: $\bbE[\cdot]$ represents the average or expected value of a random variable.

\section{Proofs}\label{sm:sec:proof}
We start with a lemma on the Pac-Bayesian bound, followed by in-depth proofs for the theories outlined in the main paper.
\begin{smlemma}\label{sm:lemma:pac}(A variant of the PAC-Bayesian bound adapted from Theorem 4.1 in \cite{supp_alquier2016properties} and from \cite{catoni2007pac}.) Let $\calD$ be a distribution over $\calX$. Denote the prior and posterior probability measure on a hypothesis set $\calF$ as $\pi(\cdot),\emprho(\cdot)\in\calM_+^1$, where $\calM_+^1$ (positive and normalized to $1$) is the set of all probability measures on $\calF$. Denote $\phi:\calF\times\calX\rightarrow\bbR$ and $\calL_{\calD}^\phi:=\bbE_\calD[\phi]$ as the loss. For $\alpha>0$ and $\delta\in(0,1]$, with probability at least $1-\delta$ over the choice of $\vx\sim\calD^n$ (a subset from $\calD$ with size of $n$), we have:
{\setlength{\abovedisplayskip}{0.1cm}%
\setlength{\belowdisplayskip}{0.1cm}%
\begin{align}
    \bbE_{f\sim\emprho}\calL_{\calD}^\phi(f)&\leq\bbE_{f\sim\emprho}\widehat{\calL}_{\calD^n}^\phi(f)+\frac{1}{\alpha}\Big[\text{KL}(\emprho\lVert\pi)+\ln \frac{1}{\delta}+\Omega(\alpha, n)\Big]\nonumber\\
    \text{where}\quad\quad &\Omega(\alpha, n)=\ln\bbE_{f\sim\pi}\bbE_{\calD^n}\exp\big\{\alpha\big(\calL_{\calD}^\phi(f)-\widehat{\calL}_{\calD^n}^\phi(f)\big)\big\}.
\end{align}}%
\end{smlemma}
\proof The Donsker-Varadhan change of measure states that for any measurable function $\varphi\!:\!\calF\!\rightarrow\!\bbR$ and $\forall\emprho$ on $\calF$, we have:
{\setlength{\abovedisplayskip}{0.1cm}%
\setlength{\belowdisplayskip}{0.1cm}%
\begin{equation}
    \bbE_{f\sim\emprho}\varphi(f)\leq\text{KL}(\emprho\lVert\pi)+\ln\bbE_{f\sim\pi}e^{\varphi(f)}.\nonumber
\end{equation}}%
Denoting $\varphi(f):=\alpha\big(\calL_{\calD}^\phi(f)-\widehat{\calL}_{\calD^n}^\phi(f)\big)$, the above inequality yields:
{\setlength{\abovedisplayskip}{0.1cm}%
\setlength{\belowdisplayskip}{0.1cm}%
\begin{align}
    \alpha\big(\bbE_{f\sim\emprho}\calL_{\calD}^\phi(f)-\bbE_{f\sim\emprho}\widehat{\calL}_{\calD^n}^\phi(f)\big)&=\bbE_{f\sim\emprho}\alpha\big(\calL_{\calD}^\phi(f)-\widehat{\calL}_{\calD^n}^\phi(f)\big)\nonumber\\
    &\leq\text{KL}(\emprho\lVert\pi)+\ln\bbE_{f\sim\pi}e^{\alpha(\calL_{\calD}^\phi(f)-\widehat{\calL}_{\calD^n}^\phi(f))}\nonumber.
\end{align}}%
Applying Markov's inequality to the random variable $\xi_\pi(X):=\bbE_{f\sim\pi}e^{\alpha(\calL_{\calD}^\phi(f)-\widehat{\calL}_{\calD^n}^\phi(f))}$, we obtain: 
{\setlength{\abovedisplayskip}{0.1cm}%
\setlength{\belowdisplayskip}{0.1cm}%
\begin{equation}
    \text{Pr}\big(\xi_\pi\leq\frac{1}{\delta}\bbE[\xi_\pi]\big)\geq1-\delta.\nonumber
\end{equation}}%
Thus, with probability at least $1-\delta$ over the choice of $\vx\sim\calD^n$, we obtain:
{\setlength{\abovedisplayskip}{0.1cm}%
\setlength{\belowdisplayskip}{0.1cm}%
\begin{equation}
    \bbE_{f\sim\emprho}\calL_{\calD}^\phi(f)\leq\bbE_{f\sim\emprho}\widehat{\calL}_{\calD^n}^\phi(f)+\frac{1}{\alpha}\Big[\text{KL}(\emprho\lVert\pi)+\ln \frac{1}{\delta}+\ln\bbE_{f\sim\pi}\bbE_{\calD^n}e^{\alpha(\calL_{\calD}^\phi(f)-\widehat{\calL}_{\calD^n}^\phi(f))}\Big].\nonumber
\end{equation}}%

\subsection{Proof of Lemma \ref{lemma:gen_error}}
\label{sm:proof:lemma:gen_error}
\textbf{Lemma \ref{lemma:gen_error}} \textit{
(Partial results of Theorem 1 in \cite{ji2021understanding}.) Assume the discriminator set $\calH$ is even, \ie, $h\in\calH$ implies $-h\in\calH$ and $\lVert h\lVert_\infty\leq\Delta$. Let $\empmu$ and $\empnu$ be empirical measures of $\mu$ and $\popnu$ with size $n$. Denote $\optnu=\inf_{\nu\in\calG}d_\calH(\empmu, \nu)$. The generalization error of GAN, defined as $\ganerr:=d_\calH(\mu,\popnu)-\inf_{\nu\in\calG}d_\calH(\mu, \nu)$, is bounded as:
{\setlength{\abovedisplayskip}{0.1cm}%
\setlength{\belowdisplayskip}{0.1cm}%
\begin{align}
    \ganerr&\leq2\big(\sup_{h\in\calH}\big|\bbE_{\mu}[h]-\bbE_{\empmu}[h]\big|+\sup_{h\in\calH}\big|\bbE_{\optnu}[h]-\bbE_{\empnu}[h]\big|\big)=2d_\calH(\mu, \empmu)+2d_\calH(\optnu, \empnu).\nonumber
\end{align}}}%
\proof
{\setlength{\abovedisplayskip}{0.1cm}%
\setlength{\belowdisplayskip}{0.1cm}%
\begin{align}
    \ganerr:=&d_\calH(\mu,\popnu)-\inf_{\nu\in\calG}d_\calH(\mu, \nu)=d_\calH(\mu, \popnu)-d_\calH(\empmu, \popnu)+d_\calH(\empmu, \popnu)-\inf_{\nu\in\calG}d_\calH(\mu,\nu)\nonumber\\
    =&\underbrace{d_\calH(\mu, \popnu)-d_\calH(\empmu, \popnu)}_{\circled{1}}+\underbrace{\inf_{v\in\calG}d_\calH(\empmu, \nu)-\inf_{v\in\calG}d_\calH(\mu, \nu)}_{\circled{2}}+\underbrace{d_\calH(\empmu, \popnu)-\inf_{v\in\calG}d_\calH(\empmu, \nu)}_{\circled{3}}.\nonumber
\end{align}}%
The three components in the above equation are upper-bounded as follows:

Upper bound \circled{1}:
{\setlength{\abovedisplayskip}{0.cm}%
\setlength{\belowdisplayskip}{0.2cm}%
\begin{align}
    &d_\calH(\mu, \popnu)-d_\calH(\empmu, \popnu)=\sup_{h\in\calH}|\bbE_\mu[h]-\bbE_{\popnu}[h]|-\sup_{h\in\calH}|\bbE_{\empmu}[h]-\bbE_{\popnu}[h]|\nonumber\\
    \leq&\sup_{h\in\calH}\big|\bbE_\mu[h]-\bbE_{\popnu}[h]-\bbE_{\empmu}[h]+\bbE_{\popnu}[h]\big|=\sup_{h\in\calH}\big|\bbE_\mu[h]-\bbE_{\empmu}[h]\big|.\nonumber
\end{align}}%

Upper bound \circled{2}: Denote $\nu^*=\inf_{\nu\in\calG}d_\calH(\mu, \nu)$. Then similar to derivation for \circled{1}, we obtain:
{\setlength{\abovedisplayskip}{0.1cm}%
\setlength{\belowdisplayskip}{0.1cm}%
\begin{align}
    &\inf_{v\in\calG}d_\calH(\empmu, \nu)-\inf_{v\in\calG}d_\calH(\mu, \nu)=\inf_{v\in\calG}d_\calH(\empmu, \nu)-d_\calH(\mu, \nu^*)\nonumber\\
    \leq&d_\calH(\empmu, \nu)-d_\calH(\mu, \nu^*)\leq\sup_{h\in\calH}\big|\bbE_\mu[h]-\bbE_{\empmu}[h]\big|.\nonumber
\end{align}}%

Upper bound \circled{3}: Here, we consider a practical scenario where the discriminator only has access to finite fake data during optimization. Recall that we denote $\optnu:=\inf_{\nu\in\calG}d_\calH(\empmu, \nu)$, thus $d_\calH(\empmu, \popnu)\geq d_\calH(\empmu, \optnu)$, leading to the inequality that: 
{\setlength{\abovedisplayskip}{0.1cm}%
\setlength{\belowdisplayskip}{0.1cm}%
\begin{align}
    &d_\calH(\empmu, \popnu)-\inf_{v\in\calG}d_\calH(\empmu, \nu)= d_\calH(\empmu, \popnu)-d_\calH(\empmu, \optnu)\nonumber\\
    =&\big(d_\calH(\empmu, \popnu)-d_\calH(\empmu, \empnu)\big)+\big(d_\calH(\empmu, \empnu)-d_\calH(\empmu, \optnu)\big)\nonumber\\
    \leq&\sup_{h\in\calH}\big|\bbE_{\popnu}[h]-\bbE_{\empnu}[h]\big|+\sup_{h\in\calH}\big|\bbE_{\optnu}[h]-\bbE_{\empnu}[h]\big|\leq2\sup_{h\in\calH}\big|\bbE_{\optnu}[h]-\bbE_{\empnu}[h]\big|.\nonumber
\end{align}}%
Integrating the three bounds we achieve the final result.

\subsection{Proof of Proposition \ref{prop:gen_sharpness}}
\label{sm:proof:prop:gen_sharpness}
\textbf{Proposition \ref{prop:gen_sharpness}} \textit{Utilizing notations from Lemma \ref{lemma:gen_error}, we define $\ganerrn$ as the generalization error of GAN parameterized as neural network classes. Let $\gradr$ and $\hessr$ represent the gradient and Hessian matrix of discriminator $h$ evaluated at $\vtheta_d$ over real training data $\empmu$, and $\gradf$ and $\hessf$ over observed fake data $\empnu$. Denoting $\levr$ and $\levf$ as the largest eigenvalues of $\hessr$ and $\hessf$, respectively, and for any $\omega>0$, the generalization error is bounded as:
{\setlength{\abovedisplayskip}{0.1cm}%
\setlength{\belowdisplayskip}{0.cm}%
\begin{align}
    \ganerrn\leq& 2\omega\big(\lVert\gradr\rVert_2+\lVert\gradf\rVert_2\big)+4R\bigg(\frac{\lVert\vtheta_d\rVert_2^2}{\omega^2}, \frac{1}{n}\bigg)+\omega^2\big(|\levr| + |\levf|\big),\nonumber
\end{align}}%
where $R\Big(\frac{\lVert\vtheta_d\rVert_2^2}{\omega^2}, \frac{1}{n}\Big)$, a term related to discriminator weights norm,  is inversely related to the data size $n$.}

\proof We start by deriving a PAC-Bayesian bound for GAN generalization error on real data. This is followed by an approach similar to Theorem 1 in \cite{foret2021sharpnessaware}, establishing a connection between this error and the discriminator's gradient direction. Finally, a Taylor expansion of the discriminator in the gradient direction is applied, paralleled by a similar formulation for fake data, culminating in our final results.

\vspace{0.1cm}
\noindent
\textbf{PAC-Bayesian bound for GAN.} Denoting $\calL_\mu:=\bbE_\mu[h]$ and the parameter of the discriminator as $\vtheta_d\in\mTheta_d$, and applying Lemma \ref{sm:lemma:pac}, we obtain:
{\setlength{\abovedisplayskip}{0.cm}%
\setlength{\belowdisplayskip}{0.1cm}%
\begin{align}
    \bbE_{\vtheta_d\sim\emprho}\calL_{\mu}(\vtheta_d)&\leq\bbE_{\vtheta_d\sim\emprho}\widehat{\calL}_{\empmu}(\vtheta_d)+\frac{1}{\alpha}\Big[\text{KL}(\emprho\lVert\pi)+\ln \frac{1}{\delta}+\Omega(\alpha, n)\Big]\nonumber\\
    \text{where}\quad\quad &\Omega(\alpha, n)=\ln\bbE_{\vtheta_d\sim\pi}\bbE_{\empmu}\exp\big\{\alpha\big(\calL_{\mu}(\vtheta_d)-\widehat{\calL}_{\empmu}(\vtheta_d)\big)\big\}.\label{sm:eq:gan_pac_init}
\end{align}}%
We then derive the upper bound for $\Omega(\alpha, n)$ on the discriminator. Let $\ell_i$ represent a realization of the random variable $\calL_\mu-h(\vx_i;\vtheta_d)$. Given that $h\in[-\Delta, \Delta]$ stated in Lemma \ref{lemma:gen_error}, changing variable $\vx_i$ to another independent copy $\vx_i'$, alters $\ell_i$ by at most $\frac{2\Delta}{n}$. Utilizing Hoeffding's lemma, we obtain:
{\setlength{\abovedisplayskip}{0.cm}%
\setlength{\belowdisplayskip}{0.1cm}%
\begin{align}
    \bbE_{\empmu}e^{\alpha(\calL_{\mu}(\vtheta_d)-\widehat{\calL}_{\empmu}(\vtheta_d))}&=\bbE_{\empmu}\exp\Big\{\frac{\alpha}{n}\sum_{i=1}^n \ell_i\Big\}=\prod_{i=1}^n\bbE\exp\Big\{\frac{\alpha}{n}\ell_i\Big\}\nonumber\\
    &\leq\prod_{i=1}^n\exp\Big\{\frac{\alpha^2(2\Delta)^2}{8n^2}\Big\}=\exp\Big\{\frac{\alpha^2\Delta^2}{2n}\Big\}.\label{sm:eq:gan_bound_hoeff}
\end{align}}%
By inserting Eq. \ref{sm:eq:gan_bound_hoeff} into Eq. \ref{sm:eq:gan_pac_init}, and setting $\alpha=n$, we arrive at:
{\setlength{\abovedisplayskip}{0.cm}%
\setlength{\belowdisplayskip}{0.1cm}%
\begin{equation}
    \bbE_{\vtheta_d\sim\emprho}\calL_{\mu}(\vtheta_d)\leq\bbE_{\vtheta_d\sim\emprho}\widehat{\calL}_{\empmu}(\vtheta_d)+\frac{1}{n}\Big[\text{KL}(\emprho\lVert\pi)+\ln\frac{1}{\delta}\Big]+\frac{\Delta^2}{2}.\label{sm:eq:gan_bound_pac}
\end{equation}}

\vspace{0.1cm}
\noindent
\textbf{Generalization error and the gradient direction of the weight.} Continuing, we adopt an analysis parallel to the proof of Theorem 1 in \cite{foret2021sharpnessaware}. According to Eq. 12 in their work, if $\pi$ is a measure on $\calN(\vzero, \sigma_\pi^2\mI)$ and $\emprho$ is a measure on $\calN(\vtheta_d, \sigma^2\mI)$ with the dimension of $\vtheta$ being $k$, it follows that:
{\setlength{\abovedisplayskip}{0.cm}%
\setlength{\belowdisplayskip}{0.cm}%
\begin{equation}
    \text{KL}(\emprho\lVert\pi)=\frac{1}{2}\Big[1+k\ln\big(1+\frac{\lVert\vtheta_d\lVert_2^2}{k\sigma^2}\big)\Big].\nonumber
\end{equation}}%
Subsequently, Eq. \ref{sm:eq:gan_bound_pac} transforms into:
{\setlength{\abovedisplayskip}{0.cm}%
\setlength{\belowdisplayskip}{0.1cm}%
\begin{equation}
    \bbE_{\vvarepsilon\sim\calN(\vzero, \sigma^2\mI)}\calL_{\mu}(\vtheta_d\!+\!\vvarepsilon)\leq\bbE_{\vvarepsilon\sim\calN(\vzero, \sigma^2\mI)}\widehat{L}_{\empmu}(\vtheta_d\!+\!\vvarepsilon) +\frac{1}{n}\Big[\frac{1}{2}+\frac{k}{2}\ln\big(1+\frac{\lVert\vtheta_d\rVert_2^2}{k\sigma^2}\big)+\ln\frac{1}{\delta}\Big]+\frac{\Delta^2}{2}.\label{sm:eq:gan_bound_parameter}
\end{equation}}%
By Lemma 1 of \cite{supp_laurent2000adaptive}, for any positive $t$, we have:
{\setlength{\abovedisplayskip}{0.1cm}%
\setlength{\belowdisplayskip}{0.1cm}%
\begin{equation}
    \text{Pr}(\lVert\vvarepsilon\rVert_2^2-k\sigma^2\geq2\sigma^2\sqrt{kt}+2t\sigma^2)\leq e^{-t}.\nonumber
\end{equation}}%
Thus, with probability $1-1/n$ (where $t=\ln n$), it follows that:
{\setlength{\abovedisplayskip}{0.1cm}%
\setlength{\belowdisplayskip}{0.1cm}%
\begin{equation}
    \lVert\vvarepsilon\rVert_2^2\leq\sigma^2\big(2\ln n+k+2\sqrt{k\ln n }\big)\leq\sigma^2k\bigg(1+\sqrt{\frac{2\ln n}{k}}\bigg)^2\leq\omega^2.\nonumber
\end{equation}}%
Assuming, as in \cite{foret2021sharpnessaware}, that perturbations in discriminator weights have negligible impact on performance over an infinite dataset, and integrating 
$\sigma$ back into Eq. \ref{sm:eq:gan_bound_parameter}, we deduce:
{\setlength{\abovedisplayskip}{0.1cm}%
\setlength{\belowdisplayskip}{0.cm}%
\begin{align}
    \calL_{\mu}(\vtheta_d)&\leq\bbE_{\vvarepsilon\sim\calN(\vzero, \sigma^2\mI)}\widehat{\calL}_{\empmu}(\vtheta_d\!+\!\vvarepsilon) +\frac{1}{n}\Big[\frac{1}{2}+\frac{k}{2}\ln\big(1+\frac{\lVert\vtheta_d\rVert_2^2}{k\sigma^2}\big)+\ln\frac{1}{\delta}\Big]+\frac{\Delta^2}{2}\nonumber\\
    &\leq\max_{\lVert\vvarepsilon\rVert_2^2\leq\omega^2}\widehat{\calL}_{\empmu}(\vtheta_d\!+\!\vvarepsilon)+\frac{1}{n}\Big[\frac{1}{2}+\frac{k}{2}\ln\bigg(1+\frac{\lVert\vtheta_d\rVert_2^2}{\omega^2}\Big(1+\sqrt{(2\ln n)/k}\Big)^2\bigg)+\ln\frac{1}{\delta}\Big]+\frac{\Delta^2}{2}.\nonumber
\end{align}}%

\vspace{0.1cm}
\noindent
\textbf{Taylor expansion in the weight gradient direction.} Observe that the maximum of $\widehat{\calL}_{\empmu}$ occurs when $\vvarepsilon$ is chosen as $\vvarepsilon\!=\!\frac{\omega\vnabla_{\empmu,\vtheta_d}}{\lVert\vnabla_{\empmu,\vtheta_d}\lVert_2}$, which is aligned with the gradient of $\widehat{\calL}_{\empmu}$ at $\vtheta_d$ over $\empmu$. We perform a second-order Taylor expansion of $\widehat{\calL}_{\empmu}$ around $\vtheta_d$. Incorporating the remainder and the higher-order terms from the Taylor expansion into $R\Big(\frac{\lVert\vtheta_d\rVert_2^2}{\omega^2}, \frac{1}{n}\Big)$, we derive:
{\setlength{\abovedisplayskip}{0.cm}%
\setlength{\belowdisplayskip}{0.1cm}%
\begin{align}
    L_{\mu}(\vtheta_d)&\leq\widehat{L}_{\empmu}\Big(\vtheta_d+\frac{\omega\vnabla_{\empmu,\vtheta_d}}{\lVert\vnabla_{\empmu,\vtheta_d}\lVert_2}\Big)+R\Big(\frac{\lVert\vtheta_d\rVert_2^2}{\omega^2}, \frac{1}{n}\Big)\nonumber\\
    &\approx \widehat{L}_{\empmu}(\vtheta_d) +\omega\lVert\vnabla_{\empmu,\vtheta_d}\rVert_2 +\frac{\omega^2}{2\lVert\vnabla_{\empmu,\vtheta_d}\rVert_2^2}\vnabla_{\empmu,\vtheta_d}^T\mH_{\empmu,\vtheta_d}\vnabla_{\empmu,\vtheta_d} + R\Big(\frac{\lVert\vtheta_d\rVert_2^2}{\omega^2},\frac{1}{n}\Big).\nonumber
\end{align}}%
Simplifying notations, we use $\gradr$ and  $\hessr$ for the gradient and Hessian matrix evaluated at $\vtheta_d$ over real seen data $\empmu$, and similar $\gradf$ and  $\hessf$ for observed fake data $\empnu$. Considering the largest eigenvalue of $\hessr$ as $\levr$, implying $\vv^T\hessr\vv\leq\levr\lVert\vv\rVert_2^2$, we bound the real data part of the generalization error of a GAN (Lemma \ref{lemma:gen_error}) parameterized as network as follows:
{\setlength{\abovedisplayskip}{0.1cm}%
\setlength{\belowdisplayskip}{0.1cm}%
\begin{align}
    \sup_{h\in\calHn}\big|\bbE_\mu[h]-\bbE_{\empmu}[h]\big|&\leq\omega\lVert\gradr\rVert_2+\frac{\omega^2}{2\lVert\gradr\rVert_2^2}\Big|\gradr^T\hessr\gradr\Big|+R\big(\frac{\lVert\vtheta_d\rVert_2^2}{\omega^2}, \frac{1}{n}\big)\nonumber\\
    &\leq\omega\lVert\gradr\rVert_2+\frac{\omega^2}{2}|\levr|+R\big(\frac{\lVert\vtheta_d\rVert_2^2}{\omega^2}, \frac{1}{n}\big).\nonumber
\end{align}}%
Similarly, the fake data part in the generalization error of GAN is:
{\setlength{\abovedisplayskip}{0.1cm}%
\setlength{\belowdisplayskip}{0.1cm}%
\begin{align}
    \sup_{h\in\calHn}\big|\bbE_{\optnu}[h]-\bbE_{\empnu}[h]\big|\leq\omega\lVert\gradf\rVert_2+\frac{\omega^2}{2}|\levf|+R\big(\frac{\lVert\vtheta_d\rVert_2^2}{\omega^2}, \frac{1}{n}\big).\nonumber
\end{align}}%
{\setlength{\abovedisplayskip}{0.1cm}%
\setlength{\belowdisplayskip}{0.1cm}%
By integrating the aforementioned two inequalities into the generalization error as detailed in Lemma \ref{lemma:gen_error}, we arrive at:
\begin{equation}
\ganerrn\leq2\omega\big(\lVert\gradr\rVert_2+\lVert\gradf\rVert_2\big)+\omega^2(|\levr|+|\levf|)+4R\big(\frac{\lVert\vtheta_d\rVert_2^2}{\omega^2}, \frac{1}{n}\big).\nonumber
\end{equation}}

\subsection{Proof of Theorem \ref{theorem:bn_centering}}
\label{sm:proof:theorem:bn_centering}
\textbf{Theorem \ref{theorem:bn_centering}} \textit{(The issue of the centering step.) Consider $\vy_1, \vy_2$ as i.i.d. samples from a symmetric distribution centered at $\vmu$, where the presence of $\vy$ implies $2\vmuy-\vy$ is also included. After the centering step, $\vyc_1, \vyc_2$ are i.i.d. samples from the centered distribution. The expected cosine similarity between these samples is given by:
{\setlength{\abovedisplayskip}{0.1cm}%
\setlength{\belowdisplayskip}{0.1cm}%
\begin{equation}
     \bbE_{\vy_1, \vy_2}\big[\cos(\vy_1, \vy_2)]\geq\bbE_{\vyc_1, \vyc_2}\big[\cos(\vyc_1, \vyc_2)\big]=0.\nonumber
 \end{equation}}}
 
\proof Given that the distribution is symmetric and even, and $\vmu_Y\!\neq\!\vzero$, the mean of the $L_2$ normalized distribution $\bbE\big[\frac{\vy}{\lVert\vy\rVert_2}\big]\neq\vzero$. Denoting the mean of the $L_2$ normalized sample as $\vmu_Z\neq\vzero$, we can derive the expectation of the cosine similarity as follows:
{\setlength{\abovedisplayskip}{0.cm}%
\setlength{\belowdisplayskip}{0.1cm}%
\begin{align}
    \bbE_{\vy_1, \vy_2}\big[\cos(\vy_1, \vy_2)]=\bbE_{\vy_1, \vy_2}\Big[\langle\frac{\vy_1}{\lVert\vy_1\lVert_2}, \frac{\vy_2}{\lVert\vy_2\lVert_2}\rangle\Big]=\bbE_{\vz_1, \vz_2}[\langle\vz_1, \vz_2\rangle]=\langle\vmu_{Z},\vmu_{Z}\rangle=\lVert\vmu_Z\rVert_2^2\geq0.\nonumber
\end{align}}%
\newcommand{\muYc}{\overset{\raisebox{-0.5ex}{\scriptsize $c$}}{\vmu}_{Y}}%

In the centered distribution with $\muYc=\vzero$ and the symmetric probability, the presence of $\vyc_2$ implies the inclusion of $-\vyc_2$. This leads us to the following derivation:
{\setlength{\abovedisplayskip}{0.cm}%
\setlength{\belowdisplayskip}{0.1cm}%
\begin{align}
    \bbE_{\vyc_1, \vyc_2}\big[\cos(\vyc_1, \vyc_2)\big]=\bbE_{\vyc_1, \vyc_2}\Big[\langle\frac{\vyc_1}{\lVert\vyc_1\rVert_2},\frac{\vyc_2}{\lVert\vyc_2\rVert_2}\rangle\Big]=\frac{1}{2}\bbE_{\vyc_1, \vyc_2}\Big[\langle\frac{\vyc_1}{\lVert\vyc_1\rVert_2},\frac{\vyc_2}{\lVert\vyc_2\rVert_2}\rangle+\langle\frac{\vyc_1}{\lVert\vyc_1\rVert_2},\frac{-\vyc_2}{\lVert\vyc_2\rVert_2}\rangle\Big]=0\nonumber
\end{align}}

Comparing the above two Equations we obtain the final inequality.

\subsection{Proof of Theorem \ref{theorem:bn_scaling}}
\label{sm:proof:theorem:bn_scaling}
\textbf{Theorem \ref{theorem:bn_scaling}} \textit{(The issue of the scaling step.) The scaling step, defined in Eq. \ref{eq:scaling}, can be expressed as matrix multiplication $\mYs = \mYc\diag(1/\vsigmay)$. The Lipschitz constant \wrt the 2-norm of the scaling step is:
{\setlength{\abovedisplayskip}{0.1cm}%
\setlength{\belowdisplayskip}{0.1cm}%
\begin{equation}
    \Big\lVert\diag\bigg(\frac{1}{\vsigmay}\bigg)\Big\rVert_\lc=\frac{1}{\sigmamin},\nonumber
\end{equation}}%
where $\sigmamin=\min_{c}\sigma_c$ represents the minimum value in $\vsigmay$.}

\proof Consider $\mLambda=\diag(\lambda_1, \cdots, \lambda_d)$, a diagonal matrix. We establish that:
{\setlength{\abovedisplayskip}{0.1cm}%
\setlength{\belowdisplayskip}{0.1cm}%
\begin{align}
\lVert\mLambda\rVert_{\text{lc}}=\max_{\lVert\vx\rVert_2=1}\lVert\Lambda\vx\rVert_2=\max_{\lVert\vx\rVert_2=1}\Big(\sum_{i=1}^d\lambda_ix_i^2\Big)^{1/2}\leq\max_{\lVert\vx\rVert_2=1}\max_i|\lambda_i|\Big(\sum_{i=1}^dx_i^2\Big)^{1/2}=\max_i|\lambda_i|\cdot\max_{\lVert\vx\rVert_2=1}\lVert\vx\rVert_2=\max_i|\lambda_i|.\nonumber
\end{align}}%

From this, it follows that:
{\setlength{\abovedisplayskip}{0.cm}%
\setlength{\belowdisplayskip}{0.1cm}%
\begin{align}
\Big\lVert\diag(\frac{1}{\vsigmay})\Big\rVert_{\lc}=\max_c\Big|\frac{1}{\sigma_c}\Big|=\frac{1}{\min_c\sigma_c}=\frac{1}{\sigmamin}.\nonumber
\end{align}}%

\subsection{Proof of Theorem \ref{theorem:chain_grad}}
\label{sm:proof:theorem:chain_grad}
\textbf{Theorem \ref{theorem:chain_grad}} \textit{(\our reduces the gradient norm of weights/latent features.) Denote the loss of discriminator with \our as $\calL$, and the resulting  batch features as $\mYo$. Let $\vysch\in\bbR^B$ be $c$-th column of $\widecheck{\mY}$, $\Deltavych, \Deltavyoch\in\bbR^{B}$ be the $c$-th column of gradient $\frac{\partial \calL}{\partial \mY}, \frac{\partial \calL}{\partial \mYo}$. Denote $\Deltavwc$ as the $c$-th column of weight gradient $\frac{\partial \calL}{\partial \mW}$ and $\lev$ as the largest eigenvalue of pre-layer features $\mA$. Then we have:
{\setlength{\abovedisplayskip}{0.cm}%
\setlength{\belowdisplayskip}{0.cm}%
\begin{equation}
    \lVert\Deltavych\rVert_2^2\leq\lVert\Deltavyoch\rVert_2^2\Big(\frac{(1-p)\psi_c+p\psimin}{\psi_c}\Big)^2-\frac{2(1-p)p\psimin}{B\psi_c}(\Deltavyoch^T\vysch)^2.\nonumber
\end{equation}
\begin{equation}
     \lVert\Deltavwc\rVert_2^2\leq\lev^2\lVert\Deltavych\rVert_2^2.\nonumber
\end{equation}}}

\proof Aligning with Theorem 4.1 from \cite{santurkar2018does} we derive the gradients of the latent feature and the weight. For convenience, we define $\mYo$ as the resulted interpolated batch features from Eq. \ref{eq:brmsn}. By applying the expectation over the $\mM$, replacing it with $p$, and using the chain rule of the backward propagation, we determine the expected gradient for each $\Deltaybc$ within $\Deltavych\in\bbR^{B}$ as follows:
{\setlength{\abovedisplayskip}{0.cm}%
\setlength{\belowdisplayskip}{0.cm}%
\setlength{\jot}{-2pt}
\begin{align}
    \Deltaybc&=\Deltayobc(1-p) + p\frac{\psimin}{\psi_c}\big(\Deltayobc - \ysbc\cdot\frac{1}{B}\sum_i^B(\Deltayoic\cdot\ysic)\big)\nonumber\\
    &=\Deltayobc\Big(\frac{(1-p)\psi_c+p\psimin}{\psi_c}\Big)-p\frac{\psimin}{\psi_c}\ysbc\frac{1}{B}\sum_{i=1}^B\Deltayoic\cdot\ysic.\nonumber
\end{align}}
The squared gradient norm for $\Deltavych$ is calculated as follows:
{\setlength{\abovedisplayskip}{0.cm}%
\setlength{\belowdisplayskip}{0.1cm}%
\begin{align}
    \lVert\Deltavych\rVert_2^2&=\lVert\Deltavyoch\rVert_2^2\Big(\frac{(1-p)\psi_c+p\psimin}{\psi_c}\Big)^2-(\frac{2(1-p)p\psimin}{B\psi_c}+\frac{p^2\psimin^2}{B\psi_c^2})(\Deltavyoch^T\vysch)^2\nonumber\\
    &\leq \lVert\Deltavyoch\rVert_2^2\Big(\frac{(1-p)\psi_c+p\psimin}{\psi_c}\Big)^2-\frac{2(1-p)p\psimin}{B\psi_c}(\Deltavyoch^T\vysch)^2.\nonumber
\end{align}}%

Using the chain rule, we derive the gradient \wrt the weight as follows:
{\setlength{\abovedisplayskip}{0.cm}%
\setlength{\belowdisplayskip}{0.cm}%
\begin{align}
    \frac{\partial \calL}{\partial W_{ic}}=\sum_{b=1}^B\frac{\partial L}{\partial\ybc}\frac{\partial\ybc}{\partial W_{ic}}=\Deltavychj^T\vachc.\nonumber
\end{align}}%
This leads to:
{\setlength{\abovedisplayskip}{0.cm}%
\setlength{\belowdisplayskip}{0.1cm}%
\begin{align}
\Deltavwc=\mA^T\Deltavych.\nonumber
\end{align}}%
Considering $\lev$ as the largest eigenvalues of $\mA$, which suggests $\vv^T\mA\vv\leq\lev\lVert\vv\rVert_2^2$, we obtain the following result:
{\setlength{\abovedisplayskip}{0.2cm}%
\setlength{\belowdisplayskip}{0.2cm}%
\begin{align}
\lVert\Deltavwc\rVert_2^2=\Deltavych^T\mA\mA^T\Deltavych\leq\lev^2\lVert\Deltavych\rVert_2^2.\nonumber
\end{align}}%

\section{The decorrelation effect of the stochastic design $\mM$}
\label{sm:sec:stochastic_M}
To analyze why the stochastic design $\mM$ outperforms the deterministic $p$, we examine the correlation coefficient between two random variables $Y_i, Y_j$ from two different channels.
\newcommand{\rvnY}{\widehat{Y}}
\newcommand{\rvsY}{\dot{Y}}
\newcommand{\rvdY}{Y'}
\newcommand{\stdd}{\sigma'}
\newcommand{\stds}{\dot{\sigma}}
\newcommand{\coed}{\varrho'}
\newcommand{\coes}{\dot{\varrho}}

\begin{theorem}\label{theorem:sm:dtm}
    Let $Y_i$, $Y_j$ be random variables from the $i$-th and $j$-th channels, respectively, where $i\neq j$. Define $\rvnY_i=\frac{Y_i}{\psi_i}\psimin$ as the normalized random variable from channel $i$ after root mean square normalization. Considering an adaptive $p$ under our control, we distinguish between the deterministic version of \our, \ie CHAIN$_\text{Dtm.}$ and our stochastic \our as:
    {\setlength{\abovedisplayskip}{0.2cm}%
    \setlength{\belowdisplayskip}{0.2cm}%
    \begin{align}
        \text{Deterministic (CHAIN$_\text{Dtm.}$):}\quad & \rvdY_i=(1-p)Y_i+p\rvnY_i,\\
        \text{Stochastic (\our):}\quad & \rvsY_i=(1-m)Y_i+m\rvnY_i, \ \text{where}\  m\sim\calB(p).
    \end{align}}%
    Assuming $\bbE[Y_i]=\bbE[Y_j]=0$, achievable through our zero mean regularization in Eq. \ref{eq:0mr}, and letting $\sigma_i, \stdd_i, \stds_i$ represent the standard deviations of $Y_i, \rvdY_i, \rvsY_i$, respectively, we define and relate the correlation coefficients of the two versions as follows:
    {\setlength{\abovedisplayskip}{0.cm}%
    \setlength{\belowdisplayskip}{0.cm}%
    \begin{align}
        \coed_{ij}=\frac{Cov(\rvdY_i, \rvdY_j)}{\stdd_i\stdd_j}\quad \geq \quad \coes_{ij}=\frac{Cov(\rvsY_i, \rvsY_j)}{\stds_i\stds_j}.
    \end{align}}
\end{theorem}

Theorem \ref{theorem:sm:dtm} reveals that the stochastic \our has a lower correlation coefficient among features from different channels than the deterministic CHAIN$_\text{Dtm.}$, indicating that the stochastic design $\mM$ exhibits a decorrelation effect.

\vspace{0.1cm}
\proof Given $\bbE[Y_i]=0$, it follows that $\bbE[\rvdY_i]=\bbE[\rvsY_i]=0$. Using the covariance definition $Cov(Z_1,Z_2)=\bbE[(Z_1-\mu_{Z_1})(Z_2-\mu_{Z_2})]$ for any two random variables $Z_1, Z_2$, we get:
{\setlength{\abovedisplayskip}{0.1cm}%
    \setlength{\belowdisplayskip}{0.1cm}%
\begin{equation}
    Cov(\rvdY_i, \rvdY_j)=\bbE[\rvdY_i\rvdY_j], \quad Cov(\rvsY_i, \rvsY_j)=\bbE[\rvsY_i\rvsY_j].\nonumber
\end{equation}}%
Since $m$ is stochastic noise independent of $Y_i, \rvnY_i$, and $m\sim\calB(p)$ implying in $\bbE[m]=p$, we conclude:
{\setlength{\abovedisplayskip}{0.1cm}%
\setlength{\belowdisplayskip}{0.1cm}%
\begin{align}
    \bbE[\rvdY_i\rvdY_j]=\bbE[\rvsY_i\rvsY_j]\ \rightarrow \  Cov(\rvdY_i, \rvdY_j)=Cov(\rvsY_i, \rvsY_j).\label{eq:sm:cov}
\end{align}}%
Next, we explore the relationship between the variances $\stdd^2_i$ and $\stds^2_i$:
{\setlength{\abovedisplayskip}{0.1cm}%
\setlength{\belowdisplayskip}{0.1cm}%
\begin{align}
    \stdd^2_i=\bbE[\rvdY^2_i]-\bbE[\rvdY_i]^2=\bbE\Big[\Big(\big(1-p+\frac{p\psimin}{\psi_i}\big)Y_i\Big)^2\Big]-0&=\big(1-p+\frac{p\psimin}{\psi_i}\big)^2\bbE[Y_i^2],\label{eq:sm:variance_dtm}\\
    \stds^2_i=\bbE[\rvsY^2_i]-\bbE[\rvsY_i]^2=(1-p)\bbE[Y_i^2]+p\bbE[\rvnY_i^2]-0&=(1-p+p\frac{\psimin^2}{\psi_i^2})\bbE[Y_i^2].\label{eq:sm:variance_sto}
\end{align}}%
Comparing Eq. \ref{eq:sm:variance_dtm} and \ref{eq:sm:variance_sto}, and considering $p\in[0,1]$, we establish the following relationship:
{\setlength{\abovedisplayskip}{0.1cm}%
\setlength{\belowdisplayskip}{0.1cm}%
\begin{align}
    &\Big(1-p+p\frac{\psimin^2}{\psi_i^2}\Big) - \Big(1-p+\frac{p\psimin}{\psi_i}\Big)^2=p(1-p)+p(1-p)\frac{\psimin^2}{\psi_i^2}-2p(1-p)\frac{\psimin}{\psi_i}\nonumber\\
    =&p(1-p)\Big(1-\frac{\psimin}{\psi_i}\Big)^2\geq0.\nonumber
\end{align}}%
Therefore, $\stdd_i\leq\stds_i$, and similarly $\stdd_j\leq\stds_j$. Coupled with Eq. \ref{eq:sm:cov}, we derive the following conclusion:
{\setlength{\abovedisplayskip}{0.1cm}%
\setlength{\belowdisplayskip}{0.1cm}%
\begin{equation}
\left \{
\setlength{\jot}{2pt}
\begin{aligned}
    Cov(\rvdY_i, \rvdY_j)&=Cov(\rvsY_i, \rvsY_j)\\
    \stdd_i\stdd_j&\leq\stds_i\stds_j
\end{aligned}
\right. \rightarrow \coed_{ij}=\frac{Cov(\rvdY_i, \rvdY_j)}{\stdd_i\stdd_j}\  \geq \  \coes_{ij}=\frac{Cov(\rvsY_i, \rvsY_j)}{\stds_i\stds_j}.\nonumber
\end{equation}}%

\begin{figure}[h!]
    \centering
    \includegraphics[width=0.7\linewidth]{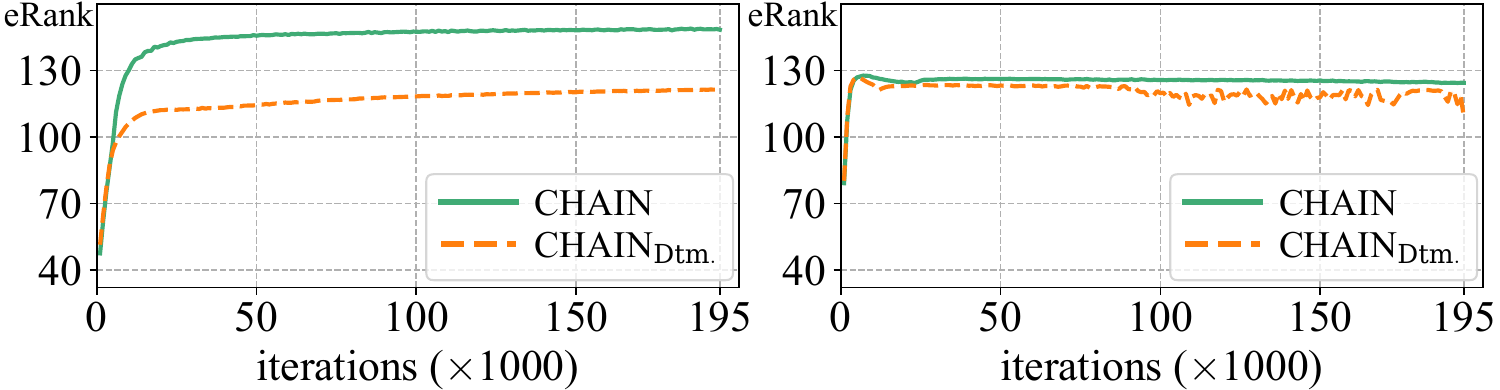}
    \begin{subfigure}{0.35\linewidth}
        \caption{10\% CIFAR-10 with OmniGAN ($d\!=\!256$).}\label{subfig:sm:rank_dtm_C10}
    \end{subfigure}
    \begin{subfigure}{0.35\linewidth}
        \caption{10\% CIFAR-100 with BigGAN ($d\!=\!256$).}\label{subfig:sm:rank_dtm_C100}
    \end{subfigure}
    \vspace{-0.3cm}
    \caption{Effective rank \cite{roy2007effective} of all pre-activation features in the discriminator for \our and CHAIN$_\text{Dtm.}$ on (a) 10\% CIFAR-10 using OmniGAN ($d\!=\!256$) and (b) 10\% CIFAR-100 with BigGAN ($d\!=\!256$).}
    \label{fig:sm:rank_dtm}
    \vspace{-0.2cm}
\end{figure}

\noindent\textbf{Experimental validation.} Decorrelation diversifies feature patterns, promoting a higher feature rank. This is demonstrated in Figure \ref{fig:sm:rank_dtm}, where \our, employing the stochastic $\mM$ over the deterministic value $p$ used by CHAIN$_\text{Dtm.}$, achieves a higher effective rank (eRank) \cite{roy2007effective}. This supports Theorem \ref{theorem:sm:dtm}, underscoring the beneficial effect of stochastic design in $\mM$ for decorrelation, and validates the design choice of \our.

\section{Implementation Details}\label{sm:sec:impl}
In this section, we overcome the mini-batch size limitation of CHAIN$_\text{batch}$, which relies solely on current batch data statistics, by developing it to \our, which ultilizes cumulative running forward/backward statistics across training. We also provide detailed implementation for Network and hyper-parameter choices, and methods applied in our ablation studies.

\newcommand{\ysb}{\widecheck{y}^{(b)}}
\newcommand{\yob}{\widehat{y}^{(b)}}
\newcommand{\yb}{y^{(b)}}
\newcommand{\ysi}{\widecheck{y}^{(i)}}

\subsection{Implementation of \our (running cumulative forward/backward statistics across training)}
\label{sm:sec:impl_chain}
Inspired by \cite{MBN, PowerNorm, ioffe2017batch}, we enhance \our to use running cumulative forward/backward statistics. We simplify our analysis by focusing on the Root Mean Square Normalization (RMSNorm), considering features of a single channel and omitting the channel index. Additionally, we exclude the constant $\epsilon$, used to avoid division by zero, as it is unnecessary for this analysis. This refinement enables the representation of the forward process for the root mean square normalization as follows:
{\setlength{\abovedisplayskip}{0.1cm}%
\setlength{\belowdisplayskip}{0.1cm}%
\begin{equation}
    \psi^2=\frac{1}{B}\sum_{b=1}^B(\yb)^2,\label{eq:sm:forward_psi_sqr}
\end{equation}}%
{\setlength{\abovedisplayskip}{0.1cm}%
\setlength{\belowdisplayskip}{0.1cm}%
\begin{equation}
     \psi=\sqrt{\psi^2},\label{eq:sm:forward_psi}
\end{equation}}%
{\setlength{\abovedisplayskip}{0.1cm}%
\setlength{\belowdisplayskip}{0.1cm}%
\begin{equation}
    \ysb=\frac{\yb}{\psi},\label{eq:sm:forward_ycheck}
\end{equation}
\begin{equation}
     \yob=\ysb \cdot \psimin. \label{eq:sm:forward_yhat}
\end{equation}}%
Leveraging the chain rule, the gradient calculation can be expressed as follows:
{\setlength{\abovedisplayskip}{0.1cm}%
\setlength{\belowdisplayskip}{0.1cm}%
\begin{align}
    \frac{\partial \calL}{\partial \ysb} &= \frac{\partial \calL}{\partial \yob} \cdot \psimin, \label{eq:sm:backward_ycheck}\\
    \frac{\partial \calL}{\partial \yb}&=\frac{1}{\psi}\Big[\frac{\partial \calL}{\partial\ysb}-\ysb\cdot \Psi\Big],\label{eq:sm:backward_ygrad}\\
    \text{where}\quad \Psi&=\frac{1}{B}\sum_{i=1}^B\frac{\partial \calL}{\partial \ysi}\cdot\ysi.\label{eq:sm:backward_Psi}
\end{align}}%
Examining the forward and backward processes reveals that Eq. \ref{eq:sm:forward_psi_sqr} and \ref{eq:sm:backward_Psi} are dependent on the batch size. To eliminate this dependency, we propose updating the cumulative statistics for these terms as follows:
{\setlength{\abovedisplayskip}{0.1cm}%
\setlength{\belowdisplayskip}{0.1cm}%
\begin{align}
    \overline{\psi^2}_{t+1}&=\overline{\psi^2}_t\cdot\alpha_\text{d}+\psi^2\cdot(1-\alpha_\text{d}),\label{eq:sm:update_psi_sqr}\\
    \overline{\Psi}_{t+1}&=\overline{\Psi}_t\cdot\alpha_\text{d}+\Psi\cdot(1-\alpha_\text{d}),\label{eq:sm:update_Psi}
\end{align}}%
where $\alpha_\text{d}$, a decay hyperpamameter, is typically set as 0.9. We replace $\psi^2, \Psi$ with their cumulative versions $\overline{\psi^2}, \overline{\Psi}$. This forms an effective algorithm for the normalization part of \our, using cumulative forward/backward statistics, as shown in Alg. \ref{algo:RMSNorm} 

\vspace{-0.2cm}
\begin{algorithm}[h!]
\caption{PyTorch-style pseudo code for Root Mean Square Normalization (RMSNorm) in \our.}
\label{algo:RMSNorm}
\small
\SetAlgoLined
    \PyComment{Y:BxdxHxW feature, running\_psi\_sqr:$\overline{\psi^2}$, decay:$\alpha_\text{d}$, eps:a small constant }\\
    \PyCode{\pykey{def} \pyfun{RMSNorm\_forward}(\pyvar{Y}, \pyvar{running\_psi\_sqr}, \pyvar{decay}\pyop{=}\pynum{0$\!$.$\!$9}, \pyvar{eps}\pyop{=}\pynum{1e-5)}\pyop{:}}\\
    \Indp 
        \PyCode{psi\_sqr\pyop{=}\pyvar{Y}$\!$.\pyfun{square}()$\!$.\pyfun{mean}(axis\pyop{=}[\pynum{0},\pynum{2},\pynum{3}], keepdim\pyop{=}\pykey{True})} \PyComment{Eq.\ref{eq:sm:forward_psi_sqr}}\\
        \PyCode{\pyvar{running\_psi\_sqr}.data.\pyfun{mul\_}(\pyvar{decay}).\pyfun{add\_}(psi\_sqr, alpha\pyop{=}\pynum{1}\pyop{-}\pyvar{decay})} \PyComment{Eq.\ref{eq:sm:update_psi_sqr}}\\
        \PyCode{running\_psi\pyop{=}(\pyvar{running\_psi\_sqr} \pyop{+} \pyvar{eps}).\pyfun{sqrt}()} \PyComment{Eq.\ref{eq:sm:forward_psi}}\\
        \PyCode{psi\_min \pyop{=} running\_psi.\pyfun{min}().\pyfun{detach}()}\\
        \PyCode{Ycheck \pyop{=} \pyvar{Y} \pyop{/} running\_psi}   \PyComment{Eq.\ref{eq:sm:forward_ycheck}}\\
        \PyCode{\pykey{return} Ycheck \pyop{*} psi\_min} \PyComment{Eq.\ref{eq:sm:forward_yhat}}\\
    \Indm
    \PyCode{}\\
    \PyComment{grad\_Yhat:BxdxHxW $\frac{\partial\calL}{\partial \widehat{Y}}$, running\_psi:$\overline{\psi}$, running\_Psi\_grad:$\overline{\Psi}$, psi\_min:$\psimin$ decay:$\alpha_\text{d}$}\\
\PyCode{\pykey{def} \pyfun{RMSNorm\_backward}$\!$(\pyvar{grad\_Yhat}, $\!$\pyvar{Ycheck}, $\!$\pyvar{running\_psi, $\!$running\_Psi\_grad}, $\!$\pyvar{psi\_min}, $\!$\pyvar{decay}\pyop{=}\pynum{0$\!$.$\!$9)$\!$}\pyop{:}}\\
    \Indp
    \PyCode{grad\_Ycheck \pyop{=} \pyvar{grad\_Yhat} \pyop{*} \pyvar{psi\_min}}  \PyComment{Eq.\ref{eq:sm:backward_ycheck}} \\
    \PyCode{Psi\_grad \pyop{=} (\pyvar{Ycheck} * grad\_Ycheck).\pyfun{mean}(axis\pyop{=}[\pynum{0},\pynum{2},\pynum{3}], keepdim\pyop{=}\pykey{True})} \PyComment{Eq.\ref{eq:sm:backward_Psi}}\\
    \PyCode{\pyvar{running\_Psi\_grad}.data.\pyfun{mul\_}(\pyvar{decay}).\pyfun{add\_}(Psi\_grad, alpha\pyop{=}1-\pyvar{decay})} \PyComment{Eq.\ref{eq:sm:update_Psi}}\\
    \PyCode{\pykey{return} (grad\_Ycheck \pyop{-} \pyvar{Ycheck} \pyop{*} \pyvar{running\_Psi\_grad}) \pyop{/} \pyvar{running\_psi}} \PyComment{Eq.\ref{eq:sm:backward_ygrad}}\\
\end{algorithm}

\subsection{Network and hyper-parameters}\label{sm:subsec:hyperparameter}
\label{sm:sec:impl_net}
\noindent\textbf{CIFAR-10/100.} We utilize OmniGAN ($d\!=\!256$ and $1024$) and BigGAN ($d\!=\!256$) with a batch size of 32. Following \cite{DA},  OmniGAN and BigGAN are trained for $1K$ epochs on full data and $5K$ epochs on 10\%/20\% data setting. \our is integrated into the discriminator, after convolutional layers $c\in\{C_1, C_2, C_S\}$ at all blocks $l\in\{1, 2, 3, 4\}$, with hyperparameters set as $\Delta_p=0.001, \tau=0.5, \lambda=20$.

\vspace{0.1cm}
\noindent\textbf{ImageNet}. We build \our upon BigGAN with 512 batch size. We adopt learning rate of 1e-4 for generator and 2e-4 for discriminator. \our is applied after convolutional layers  $c\in\{C_1, C_2, C_S\}$ at all blocks $l\in\{1, 2, 3,4, 5\}$, with hyperparameters $\Delta_p\!=\!0.001, \tau\!=\!0.5, \lambda\!=\!20$.

\vspace{0.1cm}
\noindent\textbf{5 Low-shot images ($\boldsymbol{256\!\times\!256}$).} We build \our upon StyleGAN2 with a batch size of 64, training until the discriminator has seen 25M real images. \our is applied after convolutions $c\!\in\!\{C_1, C_2\}$ at blocks $l\in\{3,4,5,6\}$. We set $\Delta_p\!=\!0.0001, \tau\!=\!0.9, \lambda\!=\!0.05$.

\begin{figure}[h!]
    \centering
    \includegraphics[width=\linewidth]{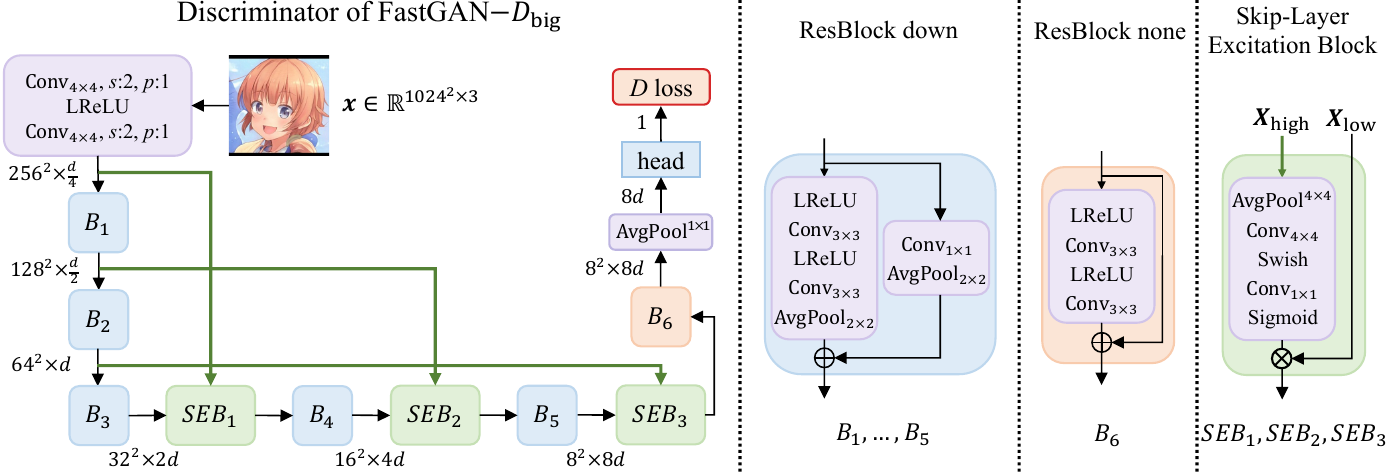}
    \caption{The discriminator of \fastgandbig. $d$: The base feature dimension. Conv$_{4\times4}$: A convolutional layer with a $4\times4$ kernel size. LReLU: Leaky ReLU activation with a slope 0.2. AvgPool$_{2\times2}$: Average pooling downscales by a factor of 2.  AvgPool$^{4\times4}$: Adaptive average pooling with a $4\times4$ output spatial size. $\boldsymbol{X}_\text{high}$: The higher resolution feature map. $\boldsymbol{X}_\text{low}$: The lower resolution feature map. For more details on skip-layer excitation block, please refer to \cite{FastGAN} and \cite{supp_hu2018squeeze}.}
    \label{fig:sm:FastGAN-Dbig}
\end{figure}

\vspace{0.1cm}
\noindent\textbf{7 Few-shot images ($\boldsymbol{1024\!\times\!1024}$)}
We replace the large discriminator in FastGAN with the one from BigGAN while removing the smaller discriminator. This modification yields \fastgandbig, with the  discriminator network architecture illustrated in Figure \ref{fig:sm:FastGAN-Dbig}. We employ a batch size of 8 and run for $100K$ iterations. We equip the discriminator with \our after convolutional layers $c\in\{C_1, C_2, C_S\}$ at blocks $l\in\{1, 2, 3, 4, 5\}$. We set $\Delta_p=0.001, \tau=0.5, \lambda=20$.

\subsection{Implementation for AGP$_\text{input}$ and AGP$_\text{weight}$}\label{sm:subsec:impl_variants}
In Table \ref{tab:variants}, we provide a comparison of \our with two gradient penalization methods: AGP$_\text{input}$ and AGP$_\text{weight}$. For AGP$_\text{input}$, we implement $\lVert\frac{\partial D}{\partial \vx}\rVert_2^2$ and $\lVert\frac{\partial f}{\partial \vx}\rVert_2^2$ where $f$ represents the feature extractor of discriminator $D$. Regarding AGP$_\text{weight}$, we also implement $\lVert\frac{\partial D}{\partial \vtheta_d}\rVert_2^2$ and $\lVert\frac{\partial f}{\partial \vtheta_d}\rVert_2^2$. We search the penalization strength $\lambda_\text{GP}$ within the range [1e-10, 20] for each dataset and variant. For 10\% CIFAR-10 w/ OmniGAN ($d\!=\!256$), the optimal settings are: AGP$_\text{input}$ with $\lVert\frac{\partial f}{\partial \vx}\rVert_2^2$ and $\lambda_\text{GP}\!\!=\!\!5$, and AGP$_\text{weight}$ with $\lVert\frac{\partial f}{\partial \vtheta_d}\rVert_2^2$ and $\lambda_\text{GP}$ set to 1e-6. For 10\% CIFAR-100 w/ BigGAN ($d\!=\!256$), the best configurations are: AGP$_\text{input}$ with $\lVert\frac{\partial f}{\partial \vx}\rVert_2^2$ and $\lambda_\text{GP}\!\!=\!\!5$, and AGP$_\text{weight}$ with $\lVert\frac{\partial D}{\partial \vtheta_d}\rVert_2^2$ and $\lambda_\text{GP}$ set to 2e-6.

\section{Additional Experiments}\label{sm:sec:add_experiment}

\subsection{Comparison with leading methods}
\label{sm:sec:add_experiments_comparison}

Table \ref{sm:tab:add_results_CIFAR} compares \our with Lottery-GAN \cite{chen2021data}, LCSA \cite{ni2022manifold}, AugSelf-GAN \cite{hou2024augmentation}, and NICE \cite{ni2024nice}, showing the superiority of \our. Unlike AugSelf-GAN, LotteryGAN, and NICE, which need extra forward or backward passes for augmentation, and LCSA, which demands more computation and weights for dictionary learning, \our is more efficient, needing negligible computation for normalization.

\begin{table*}[h!]
\begin{center}
\vspace{-0.1cm}
\caption{Comparing CIFAR-10/100 results with varying data percentages, using \our \vs other leading methods, on BigGAN ($d\!=\!256$).}
\vspace{-0.3cm}
\label{sm:tab:add_results_CIFAR}
\footnotesize
\setlength{\tabcolsep}{0.095cm}
\begin{tabular}{!{\vrule width \boldlinewidth}l!{\vrule width \boldlinewidth}ccc|ccc|ccc!{\vrule width \boldlinewidth}ccc|ccc|ccc!{\vrule width \boldlinewidth}}
\topline
 \multirow{3}{*}{Method} & \multicolumn{9}{c!{\vrule width \boldlinewidth}}{CIFAR-10} & \multicolumn{9}{c!{\vrule width \boldlinewidth}}{CIFAR-100}\\
 \cline{2-19}
& \multicolumn{3}{c|}{10\% data} &  \multicolumn{3}{c|}{20\% data} & \multicolumn{3}{c!{\vrule width \boldlinewidth}}{100\% data}
& \multicolumn{3}{c|}{10\% data} &  \multicolumn{3}{c|}{20\% data} & \multicolumn{3}{c!{\vrule width \boldlinewidth}}{100\% data}\\ 
\cline{2-19}
& IS$\uparrow$ & tFID$\downarrow$ & vFID$\downarrow$ 
& IS$\uparrow$ & tFID$\downarrow$ & vFID$\downarrow$ 
& IS$\uparrow$ & tFID$\downarrow$ & vFID$\downarrow$
& IS$\uparrow$ & tFID$\downarrow$ & vFID$\downarrow$ 
& IS$\uparrow$ & tFID$\downarrow$ & vFID$\downarrow$ 
& IS$\uparrow$ & tFID$\downarrow$ & vFID$\downarrow$\\
\middleline
LeCam+DA        & 8.81 & 12.64 & 16.42 & 9.01 & 8.53 & 12.47 & 9.45 & 4.32 & 8.40  &     9.17 & 22.75 & 27.14  & 10.12 & 15.96 & 20.42  & 11.25 & 6.45 & 11.26 \\
\hline
+Lottery-GAN        & 8.77 & 11.47 & 15.48 & 8.99 & 7.91 & 11.83 & 9.39 & 4.21 & 8.25  &     9.05 & 20.63 & 25.31  & 9.55  & 15.18 & 20.01  & 11.28 & 6.32 & 11.10 \\
+LCSA           & 8.96 & 10.05 & 13.88 & 9.04 & 6.95 & 10.95 & 9.47 & 3.75 & 7.83  &     \textbf{10.28} & 18.24 & 23.12  & \textbf{10.67} & 10.16 & 15.00  & 11.17 & 5.85 & 10.64 \\
+NICE           & 8.99 & 9.86  & 13.81 & 9.12 & 6.92 & 10.89  & 9.52 & 3.72 & 7.81  &     9.35 & 14.95 & 19.60  & 10.54 & 10.02 & 14.93  & 11.28 & 5.72 & 10.40 \\
+AugSelf-GAN        & \textbf{9.04} & 8.98 & 12.94 & 9.13 & 6.42 & 10.54 & 9.48 & 3.68  & 7.73 &  9.89 & 14.02 & 18.84 & 10.43 & 11.32 & 16.02 & 11.25 & 5.43 & 10.14 \\
\rowcolor{tblcolor}+\our           & 8.96 & \textbf{8.54} & \textbf{12.51} 
                & \textbf{9.27} & \textbf{5.92} & \textbf{9.90} 
                & \textbf{9.52} & \textbf{3.51} & \textbf{7.47}
                & 10.11 & \textbf{12.69} & \textbf{17.49}
                & 10.62 & \textbf{9.02} & \textbf{13.75}
                & \textbf{11.37}    & \textbf{5.26} & \textbf{9.85}
\\
\bottomline
\end{tabular}
\end{center}
\vspace{-0.4cm}
\end{table*}

\subsection{Gradient analysis on 10\% CIFAR-100 using BigGAN ($\boldsymbol{d\!=\!256}$)}
\begin{figure}[h!]
\begin{center}
\begin{subfigure}{0.35\linewidth}
\includegraphics[width=\linewidth]{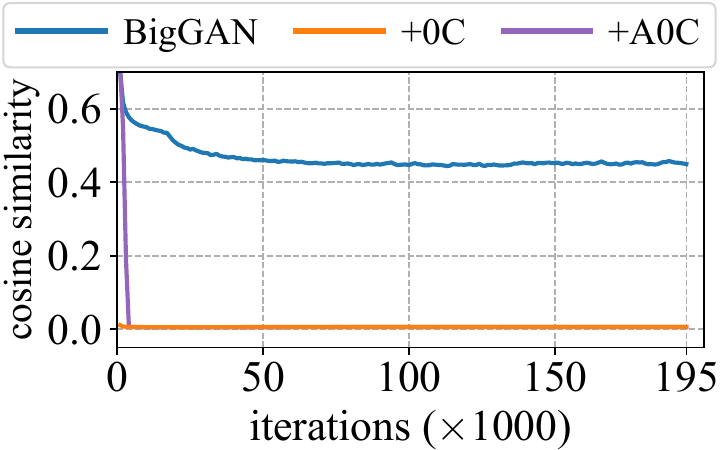}
\caption{Mean cosine similarity.}
\end{subfigure}
\hspace{0.01\linewidth}
\begin{subfigure}{0.35\linewidth}
\includegraphics[width=\linewidth]{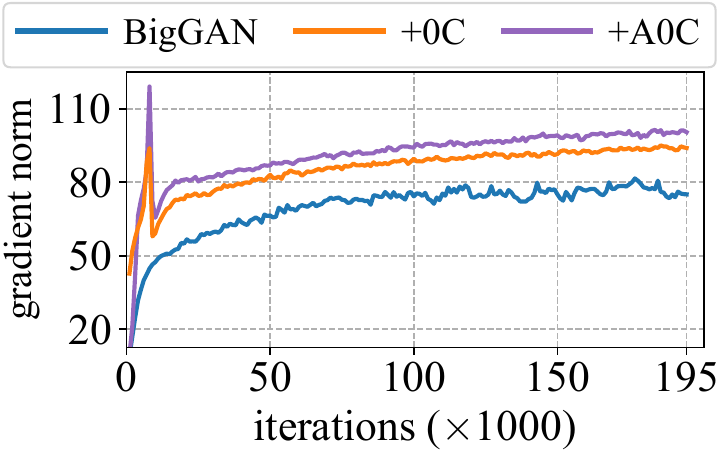}
\caption{Gradient norm.}\label{subfig:sm:gradient_0c_C100}
\end{subfigure}
\vspace{-0.3cm}
\caption{(a) Mean cosine similarity of discriminator pre-activation features, and (b) gradient norm of the feature extractor \wrt the input are evaluated for BigGAN, BigGAN+0C (using the centering step in Eq. \ref{eq:centering}), and BigGAN+A0C (adaptive interpolation between centered and uncentered features). Evaluation conducted on 10\% CIFAR-100 data with BigGAN ($d\!=\!256$).}
\label{fig:sm:similarity_C100}
\end{center}
\vspace{-0.7cm}
\end{figure}

\begin{figure}[h!]
    \centering
    \includegraphics[width=0.7\linewidth]{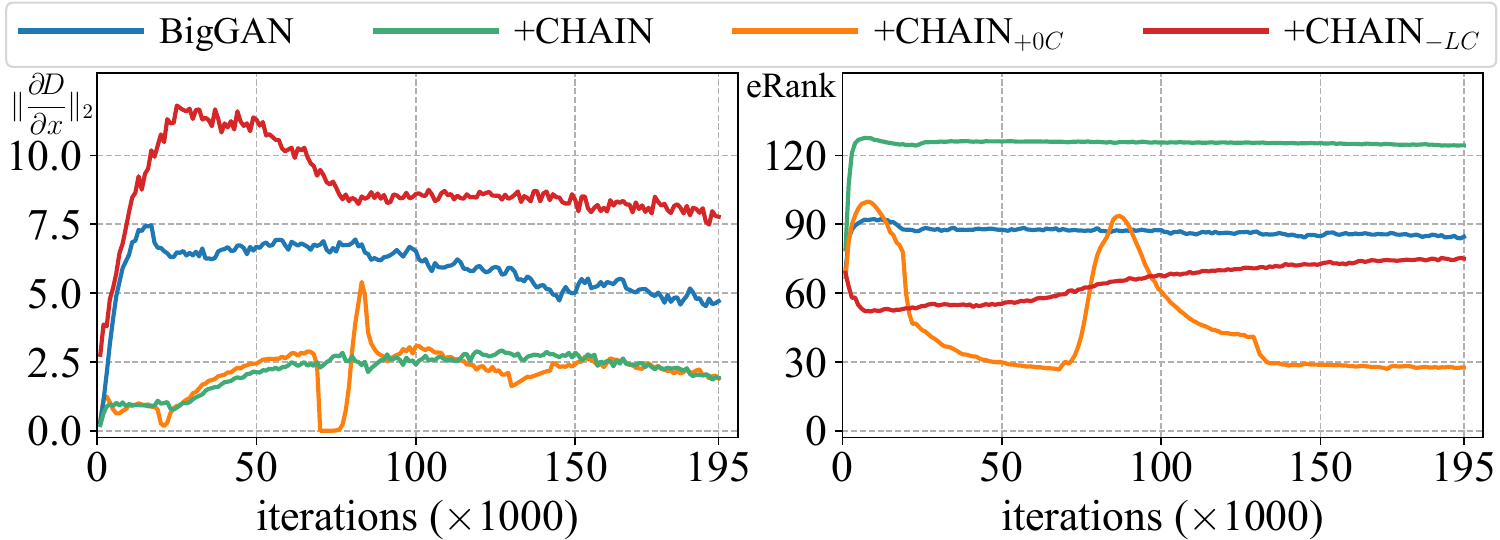}
    \begin{subfigure}{0.35\linewidth}
        \caption{Gradient norm.}\label{subfig:sm:grad_C100}
    \end{subfigure}
    \begin{subfigure}{0.35\linewidth}
        \caption{Effective rank.}\label{subfig:sm:rank_C100}
    \end{subfigure}
    \vspace{-0.3cm}
    \caption{(a) Gradient norm of discriminator output \wrt input during training, and (b) effective rank \cite{roy2007effective} of the pre-activation features in discriminator, are evaluated on 10\% CIFAR-100 data with BigGAN ($d\!=\!256$). CHAIN$_{+0C}$ indicates \our with the centering step included, while CHAIN$_{-LC}$ represents \our without the Lipschitzness constraint.}
    \label{fig:sm:gradient_rank_C100}
\end{figure}

In this section, we present experiments conducted on 10\% CIFAR-100 using BigGAN ($d\!=\!256$). Figure \ref{fig:sm:similarity_C100} provides additional validation of Theorem \ref{theorem:bn_centering}, illustrating how the centering step leads to feature differences and an associated increase in gradients. Meanwhile, Figure \ref{fig:sm:gradient_rank_C100} confirms Theorem \ref{theorem:bn_scaling}, highlighting that the scaling step causes gradient explosions during GAN training and results in rank deficiency.

\subsection{The rank efficiency of \our over AGP$_\text{weight}$}
\begin{figure}[h!]
    \centering
    \includegraphics[width=0.7\linewidth]{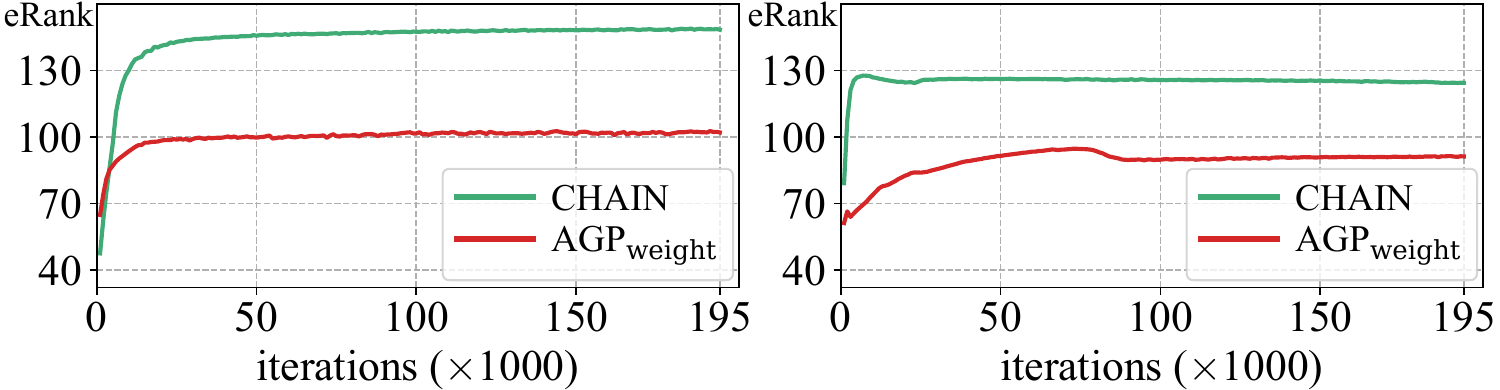}
    \begin{subfigure}{0.35\linewidth}
        \caption{10\% CIFAR-10 with OmniGAN ($d\!=\!256$).}\label{subfig:sm:rank_agp_C10}
    \end{subfigure}
    \begin{subfigure}{0.35\linewidth}
        \caption{10\% CIFAR-100 with BigGAN ($d\!=\!256$.)}\label{subfig:sm:rank_agp_C100}
    \end{subfigure}
    \vspace{-0.3cm}
    \caption{Effective Rank \cite{roy2007effective} for \our and AGP$_\text{weight}$ on (a) 10\% CIFAR-10 using OmniGAN ($d\!=\!256$) and (b) 10\% CIFAR-100 with BigGAN ($d\!=\!256$).}
    \label{fig:sm:rank_agp}
\end{figure}

Both \our and AGP$_\text{weight}$ can reduce the discriminator weight gradient to improve generalization, but \our gains a crucial advantage from normalization. The normalization step in \our balances features among channels and orthogonalizes features \cite{supp_bn_ortho, daneshmand2020batch}. Figure \ref{fig:sm:rank_agp} clearly illustrates that  \our achieves a higher effective rank compared to AGP$_\text{weight}$. Discriminators with higher rank efficiency can fully utilize their width (balanced channels) and depth, resulting in enhanced expressivity and superior representation capability.

\subsection{The stability of feature norm of \our during training}
\begin{figure}[h!]
    \centering
    \includegraphics[width=0.7\linewidth]{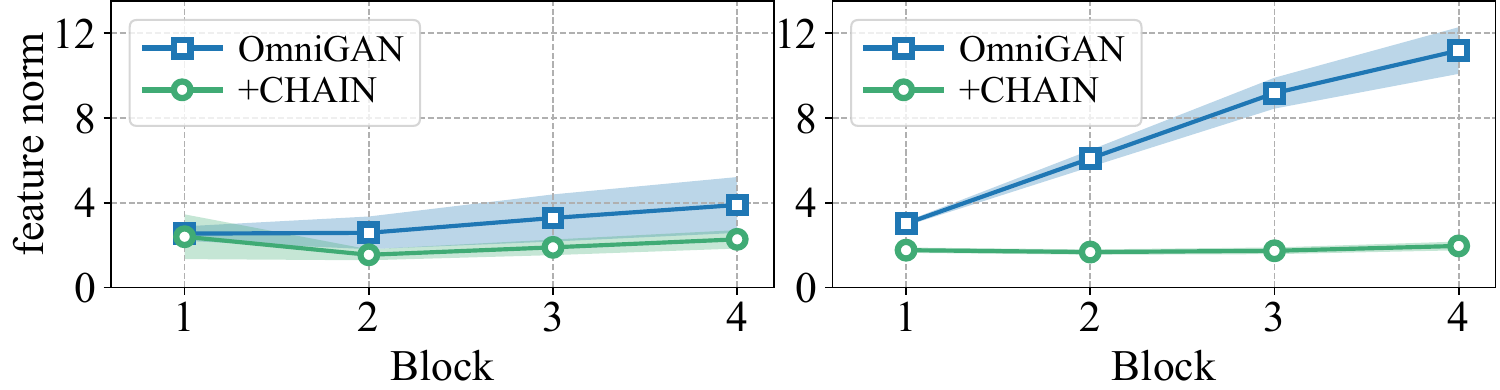}
    \begin{subfigure}{0.35\linewidth}
        \caption{$<5k$ iteration.}
    \end{subfigure}
    \begin{subfigure}{0.35\linewidth}
        \caption{$>5k$ iteration.}
    \end{subfigure}
    \vspace{-0.3cm}
    \caption{Feature norms during training w/ \vs w/o \our, are evaluated on 10\% CIFAR-10 using OmniGAN ($d\!=\!256$).}
    \label{fig:sm:feature_norm}
\end{figure}

Our work examines modern discriminators with residual blocks, where the main and skip branch features are added at the end of each block (see Figure \ref{fig:pipline}). Despite the scaling factor $\leq 1$ induced by the Lipschitz constraint (as in Eq. \ref{eq:rms}), feature norms remain stable across layers thanks to the skip connections. Figure \ref{fig:sm:feature_norm} presents feature norms at the end of each block, averaged over early ($0-5k$ iteration) and later training stages ($>5k$ iteration). Initially, both methods exhibit similar feature norms, but as training processes, baseline norms increase while CHAIN maintains stable norms across layers due to the adaptive interpolation between normalized and unnormalized features (as in Eq. \ref{eq:brmsn}).

\section{Training overhead}\label{sm:sec:overhead}
Table \ref{tab:time_cost} presents the number of parameters, multiply-accumulate (MACs) operations (for both generator and discriminator), the number of GPUs, and the cost in time (seconds per 1000 images, secs/$k$img). Notably, \our introduces only a small fraction of the time cost, ranging from 6.3\% to 9.6\% across these datasets.

\begin{table}[!ht]
\begin{center}
\footnotesize
\renewcommand{\arraystretch}{1.1}

\caption{Number of parameters, MACs and secs/$k$img for models with \vs without the \our. Experiments were performed on NVIDA A100 GPUs.}
\vspace{-0.3cm}
\begin{tabular}{!{\vrule width \boldlinewidth}ccccc!{\vrule width \boldlinewidth}ccc!{\vrule width \boldlinewidth}ccc!{\vrule width \boldlinewidth}}
\topline
\multirow{2}{*}{Dataset} & \multirow{2}{*}{Resolution} & \multirow{2}{*}{Backbone} & \multirow{2}{*}{$d$} & \multirow{2}{*}{GPUs} & \multicolumn{3}{c!{\vrule width \boldlinewidth}}{Baseline} & \multicolumn{3}{c!{\vrule width \boldlinewidth}}{+\our} \\
\cline{6-11}
& & & & & \#Par. & MACs & sec/$k$img & \#Par. & MACs & sec/$k$img \\
\hline
\multirow{2}{*}{CIFAR-10} & \multirow{2}{*}{$32\!\times\!32$} &  BigGAN & \multirow{2}{*}{256} & \multirow{2}{*}{1} & 8.512M & 2.788G & 0.79 & 8.512M & 2.791G & 0.84 \\
 & & OmniGAN & &  & 8.512M & 2.788G & 0.80 & 8.512M & 2.790G & 0.85 \\
\hline
\multirow{2}{*}{CIFAR-100} & \multirow{2}{*}{$32\!\times\!32$} & BigGAN & \multirow{2}{*}{256} & \multirow{2}{*}{1} & 8.811M & 2.788G & 0.80 & 8.811M & 2.791G & 0.85 \\
& & OmniGAN & & & 8.811M & 2.788G & 0.81 & 8.811M & 2.791G & 0.85 \\
\hline
ImageNet & $64\!\times\!64$ & BigGAN & 384 & 2 & 115.69M & 18.84G & 1.79 & 115.69M & 19.12G & 1.91\\
\hline
5 Low-shot datasets & $256\!\times\!256$ & StyleGAN2 & 512 & 2 & 48.77M & 44.146G & 5.66 & 48.77M & 44.151G & 6.06 \\
\hline
7 Few-shot datasets & $1024\!\times\!1024$ & \fastgandbig & 64 & 1 & 42.11M & 23.98G & 32.79 & 42.11M & 24.00G & 35.94 \\
\bottomline
\end{tabular}
\label{tab:time_cost}
\end{center}
\end{table}

\section{Generated Images}\label{sm:sec:generated_images}
\begin{figure*}[!ht]
\centering
    \begin{subfigure}{0.49\linewidth}
    \centering
    \includegraphics[width=\linewidth]{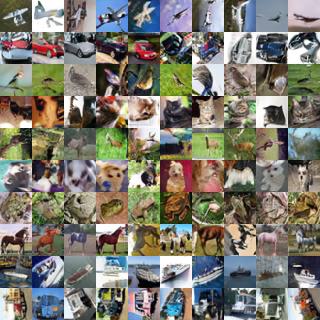}
    \caption{ADA}
    \label{subfig:C10_ada}
    \end{subfigure}
    \hfill
     \begin{subfigure}{0.49\linewidth}
    \centering
    \includegraphics[width=\linewidth]{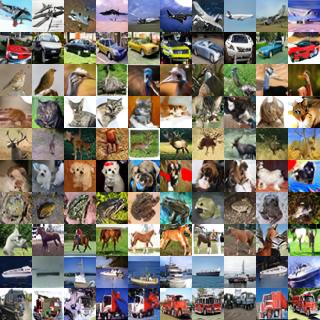}
    \caption{ADA+\our}
    \label{subfig:C10_chain}
    \end{subfigure}
    \caption{Generated images using (a) ADA and (b) ADA+\our on 10\% CIFAR-10 with OmniGAN ($d\!=\!1024$). Note that ADA leaks the rotation augmentation artifacts (row 1, 2 and 10).}
    \label{fig:C10}
\end{figure*}

\begin{figure*}[!ht]
\centering
    \begin{subfigure}{0.49\linewidth}
    \centering
    \includegraphics[width=\linewidth]{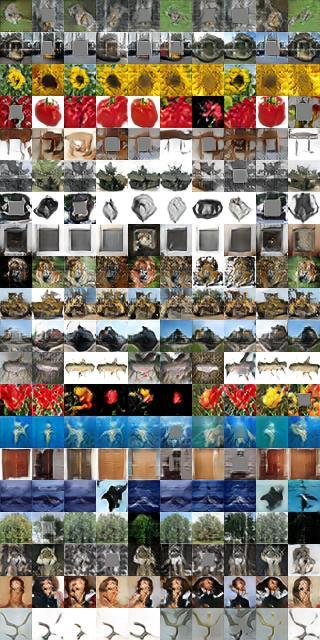}
    \caption{DA}
    \label{subfig:C100_da}
    \end{subfigure}
    \hfill
     \begin{subfigure}{0.49\linewidth}
    \centering
    \includegraphics[width=\linewidth]{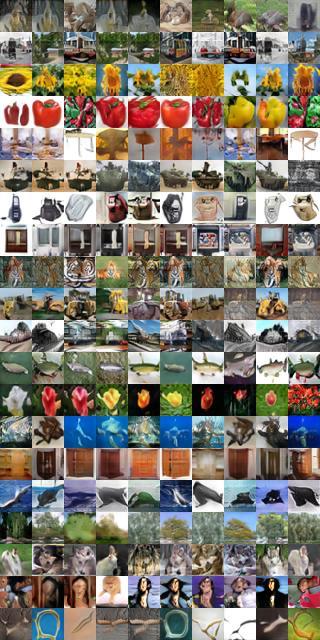}
    \caption{DA+\our}
    \label{subfig:C100_chain}
    \end{subfigure}
    \caption{Generated images using (a) DA and (b) DA+\our on 10\% CIFAR-100 with BigGAN ($d=256$). We present the last 20 of 100 classes. \our clearly enhances the diversity and quality of the generated images. Notably, DA leaks the cutout augmentation artifacts (row 1, 2, 4 and 18).}
    \label{fig:C100}
\end{figure*}

\begin{figure*}[!ht]
    \centering
    \begin{minipage}{0.03\linewidth}
        \rotatebox{90}{\small 2.5\% ImageNet ($64\times64$)}
    \end{minipage}
    \begin{minipage}{0.96\linewidth}
        \includegraphics[width=0.485\linewidth]{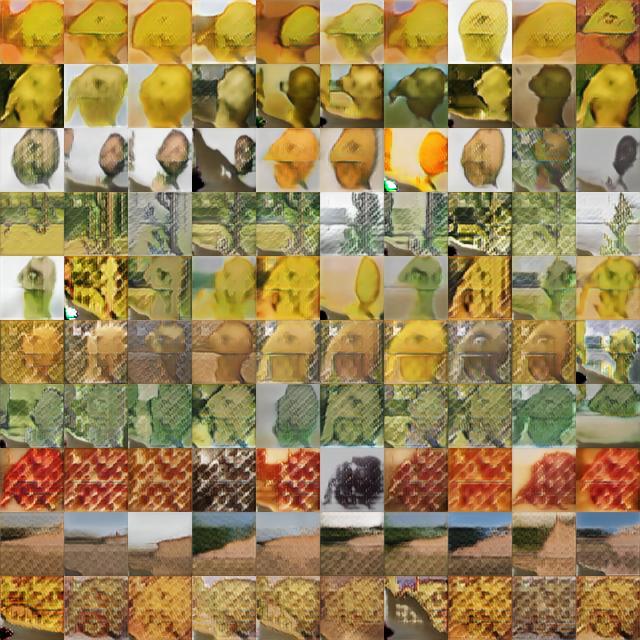}
        \hspace{0.01\linewidth}
        \includegraphics[width=0.485\linewidth]{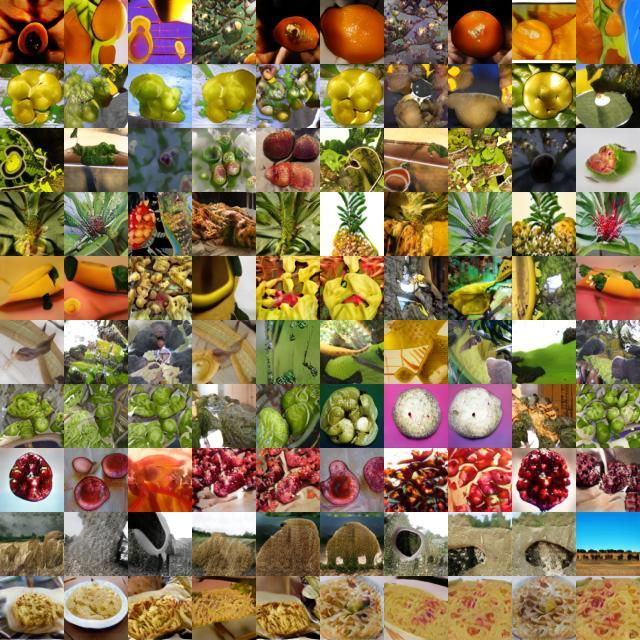}
        \vspace{0.1cm}
    \end{minipage}

    \begin{minipage}{0.03\linewidth}
        \rotatebox{90}{\small 10\% ImageNet ($64\times64$)}
    \end{minipage}
    \begin{minipage}{0.96\linewidth}
        \includegraphics[width=0.485\linewidth]{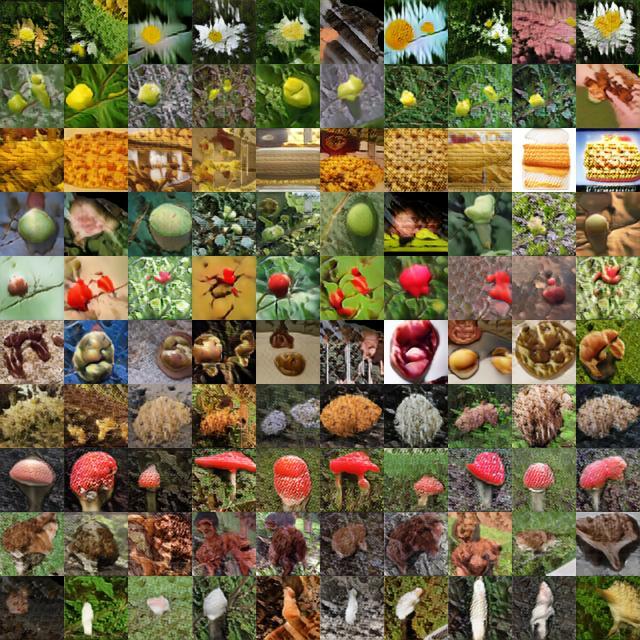}
        \hspace{0.01\linewidth}
        \includegraphics[width=0.485\linewidth]{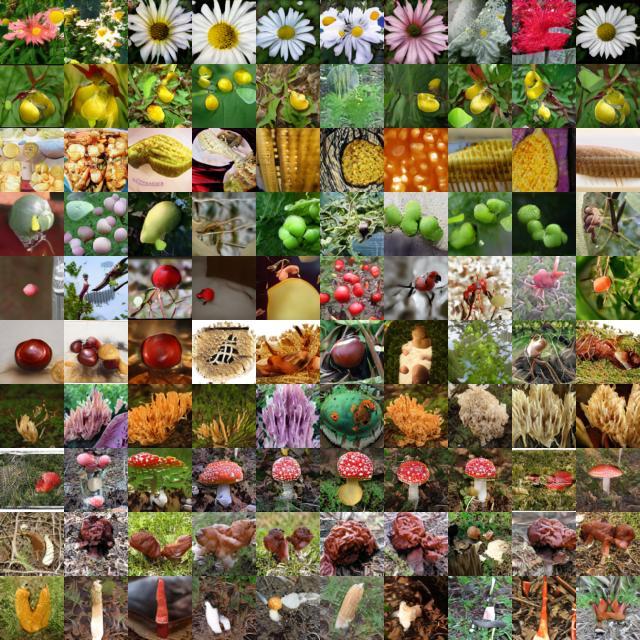}
        \vspace{0.1cm}
    \end{minipage}
    
    \begin{subfigure}{0.49\linewidth}
    \centering
    \caption{BigGAN+ADA}
    \label{subfig:I64_ada}
    \end{subfigure}
    \begin{subfigure}{0.49\linewidth}
    \centering
    \caption{BigGAN+ADA+\our}
    \label{subfig:I64_chain}
    \end{subfigure}
    \caption{Visual comparison between ADA \vs ADA+\our on 2.5\% and 10\% ImageNet($64\times64$) data. ADA struggles to capture the structure and diversity of the data, while \our clear improves the diversity and visual quality of generated images.}
    \label{fig:I64}
\end{figure*}

\begin{figure*}[!ht]
    \centering
    \begin{minipage}{0.03\linewidth}
        \rotatebox{90}{\small Obama}
    \end{minipage}
    \begin{minipage}{0.96\linewidth}
        \includegraphics[width=0.485\linewidth]{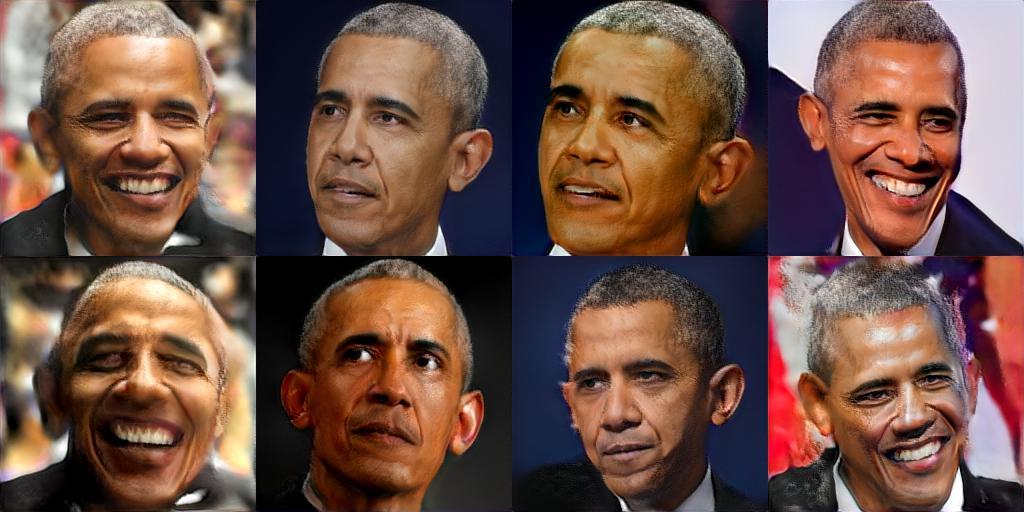}
        \hspace{0.01\linewidth}
        \includegraphics[width=0.485\linewidth]{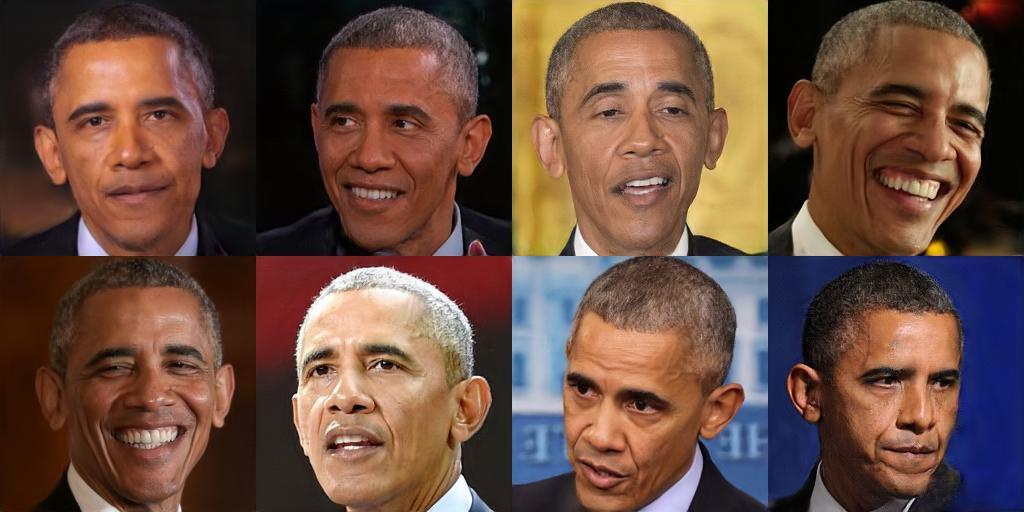}
        \vspace{0.1cm}
    \end{minipage}
    \begin{minipage}{0.03\linewidth}
        \rotatebox{90}{\small Grumpy Cat}
    \end{minipage}
    \begin{minipage}{0.96\linewidth}
        \includegraphics[width=0.485\linewidth]{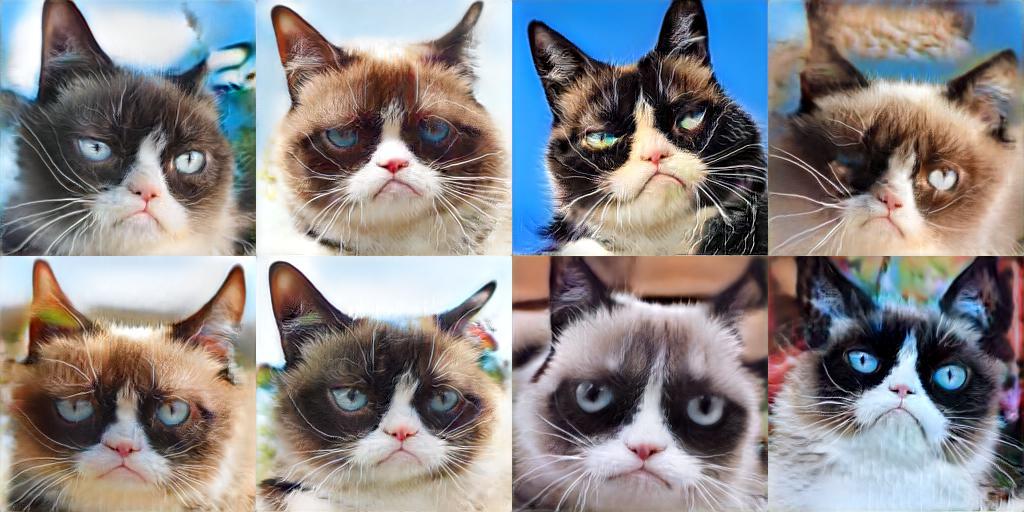}
        \hspace{0.01\linewidth}
        \includegraphics[width=0.485\linewidth]{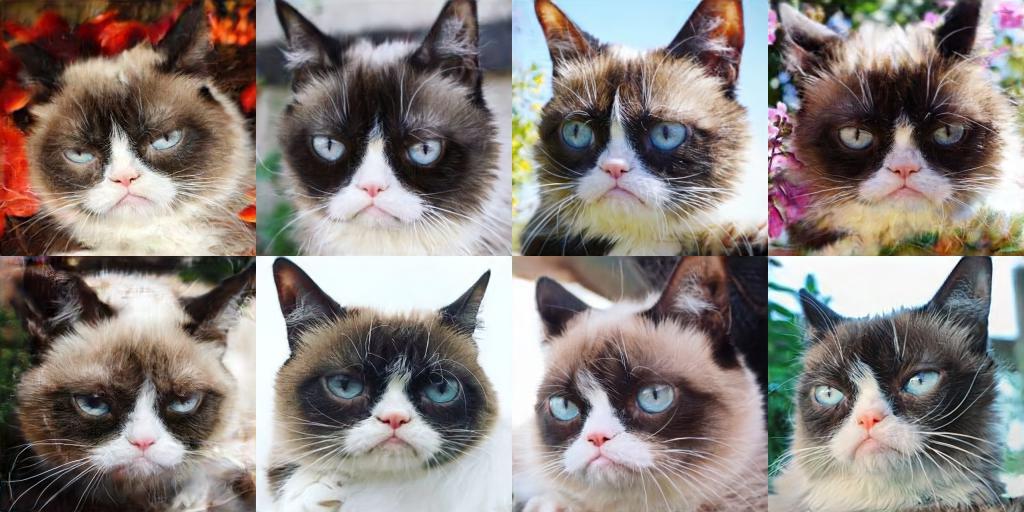}
        \vspace{0.1cm}
    \end{minipage}
    
    \begin{minipage}{0.03\linewidth}
        \rotatebox{90}{\small Panda}
    \end{minipage}
    \begin{minipage}{0.96\linewidth}
        \includegraphics[width=0.485\linewidth]{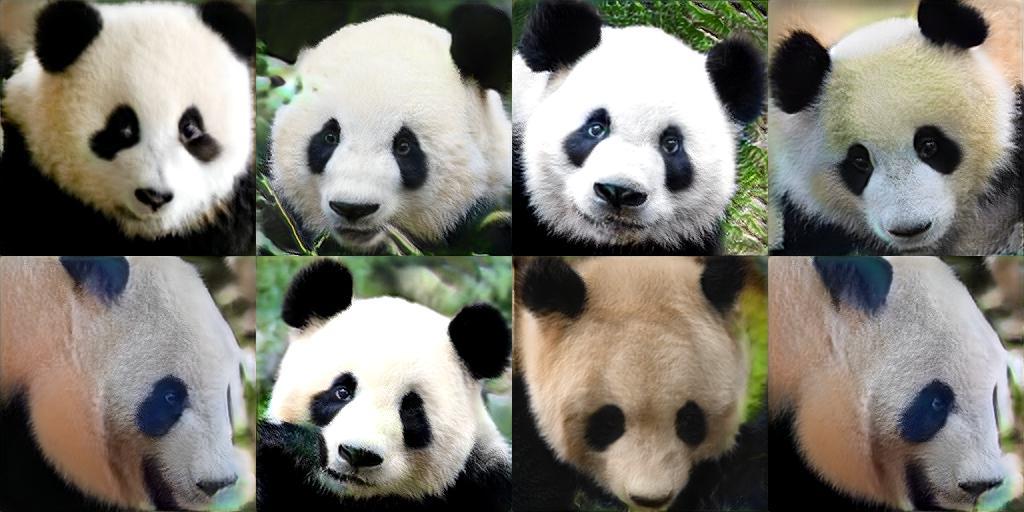}
        \hspace{0.01\linewidth}
        \includegraphics[width=0.485\linewidth]{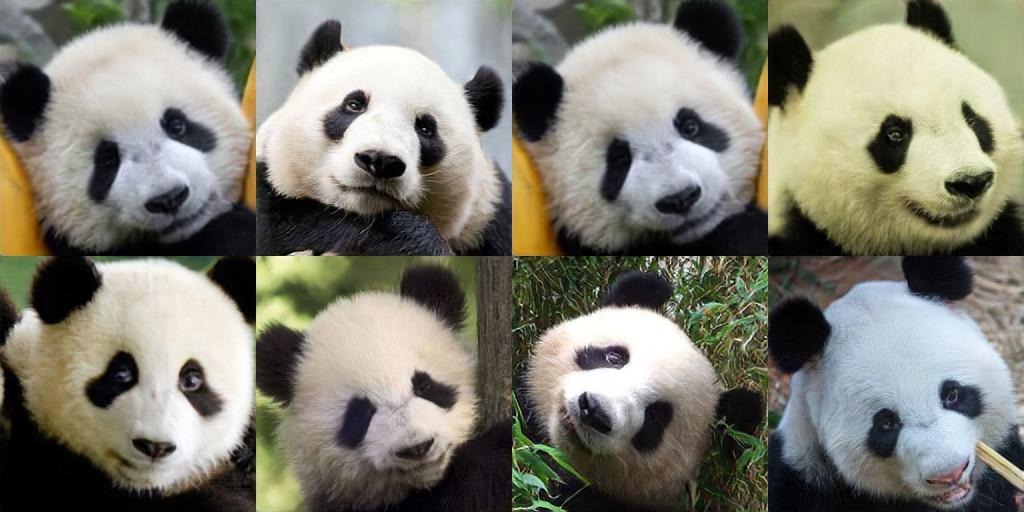}
    \vspace{0.1cm}
    \end{minipage}
    \begin{minipage}{0.03\linewidth}
        \rotatebox{90}{\small AnimalFace Cat}
    \end{minipage}
    \begin{minipage}{0.96\linewidth}
        \includegraphics[width=0.485\linewidth]{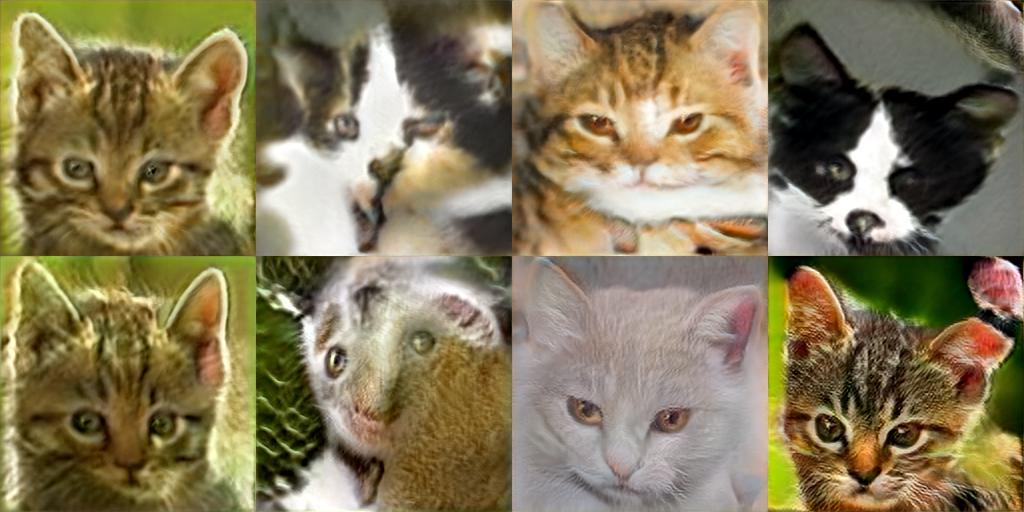}
        \hspace{0.01\linewidth}
        \includegraphics[width=0.485\linewidth]{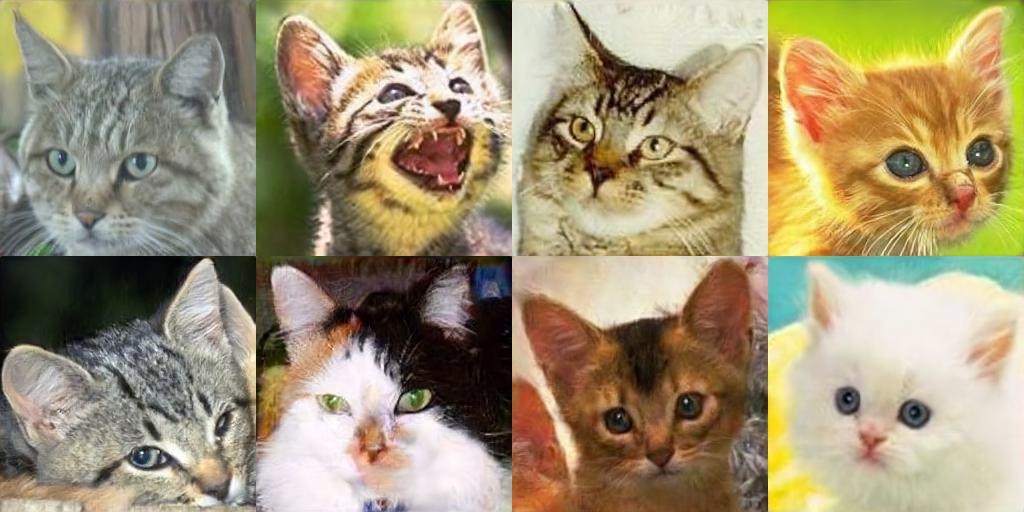}
    \vspace{0.1cm}
    \end{minipage}
    \begin{minipage}{0.03\linewidth}
        \rotatebox{90}{\small AnimalFace Dog}
    \end{minipage}
    \begin{minipage}{0.96\linewidth}
        \includegraphics[width=0.485\linewidth]{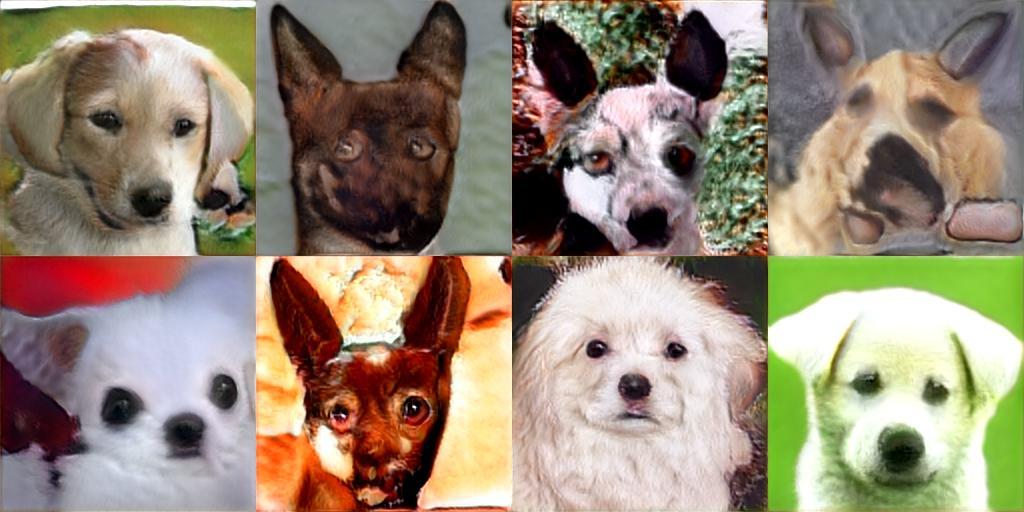}
        \hspace{0.01\linewidth}
        \includegraphics[width=0.485\linewidth]{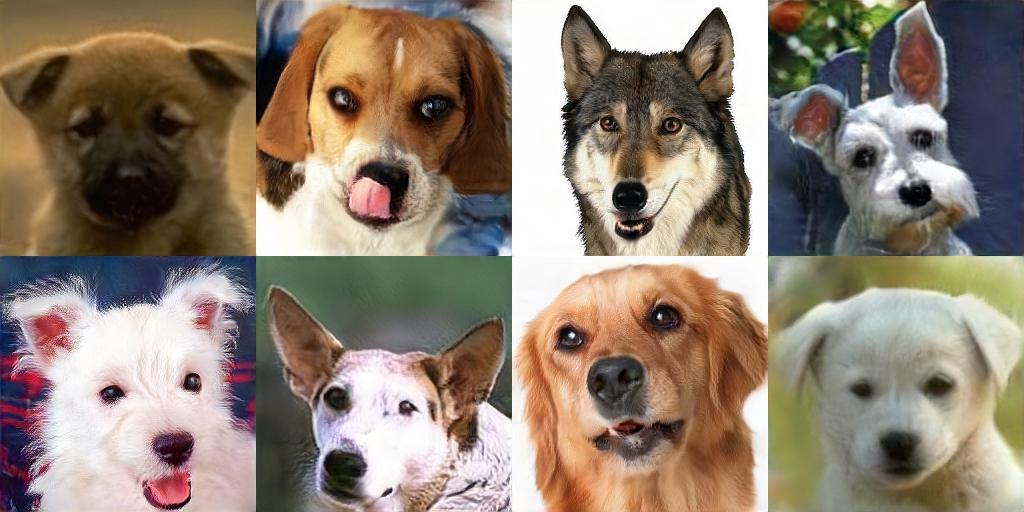}
    \end{minipage}
    \begin{subfigure}{0.49\linewidth}
    \centering
    \caption{StyleGAN2+ADA}
    \label{subfig:lowshot_ada}
    \end{subfigure}
    \begin{subfigure}{0.49\linewidth}
    \centering
    \caption{StyleGAN2+ADA+\our}
    \label{subfig:lowshot_ada_nice}
    \end{subfigure}
    \caption{Visual comparison between ADA and ADA+\our on 100-shot and AnimalFace datasets ($256\!\times\!256$). The integration of \our clearly improves the image quality.}
    \label{fig:lowshot256}
\end{figure*}

\begin{figure*}[!ht]
    \centering
    \begin{minipage}{0.02\linewidth}
        \rotatebox{90}{\small Shells}
    \end{minipage}
    \begin{minipage}{0.97\linewidth}
        \includegraphics[width=0.373\linewidth]{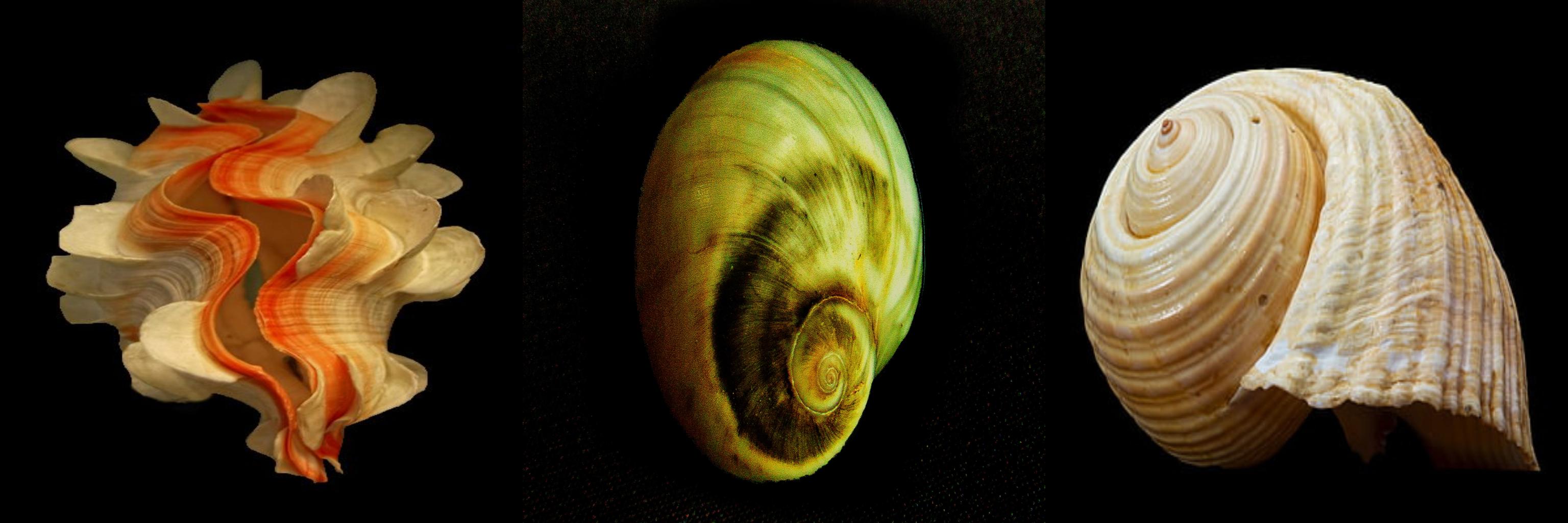}
        \includegraphics[width=0.621\linewidth]{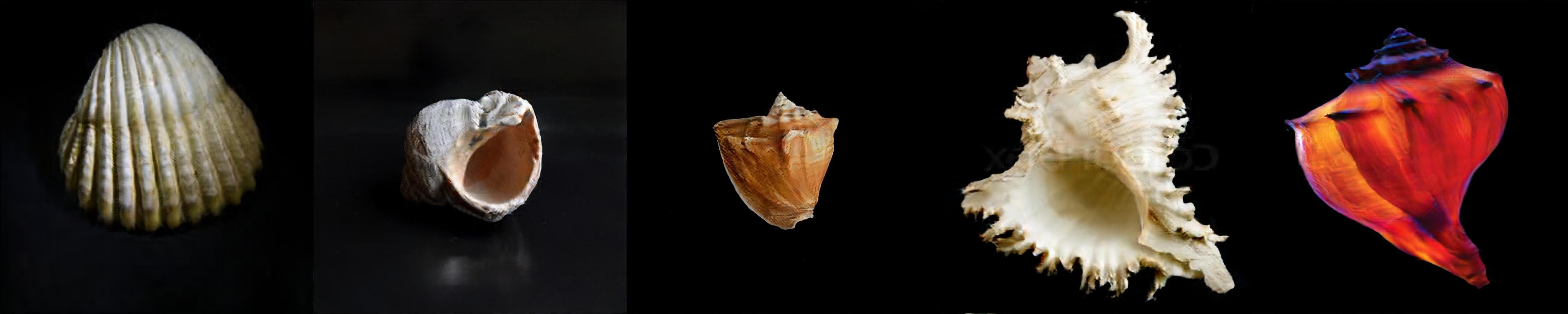}
    \end{minipage}

    \begin{minipage}{0.02\linewidth}
        \rotatebox{90}{\small Skulls}
    \end{minipage}
    \begin{minipage}{0.97\linewidth}
        \includegraphics[width=0.373\linewidth]{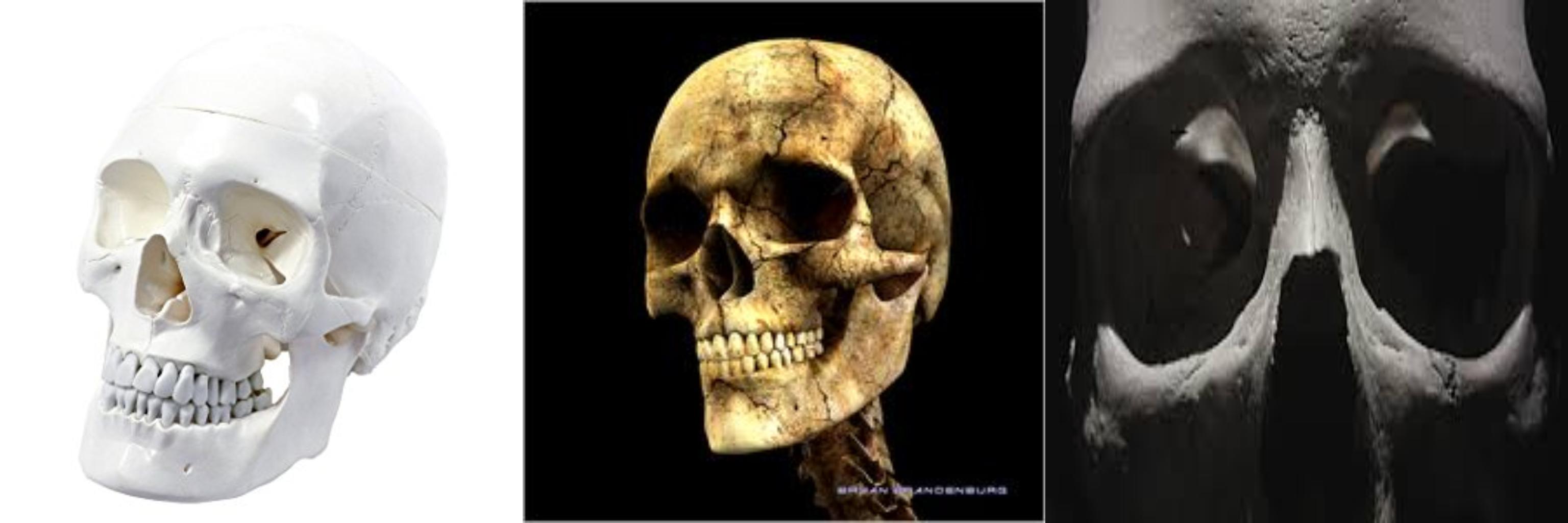}
        \includegraphics[width=0.621\linewidth]{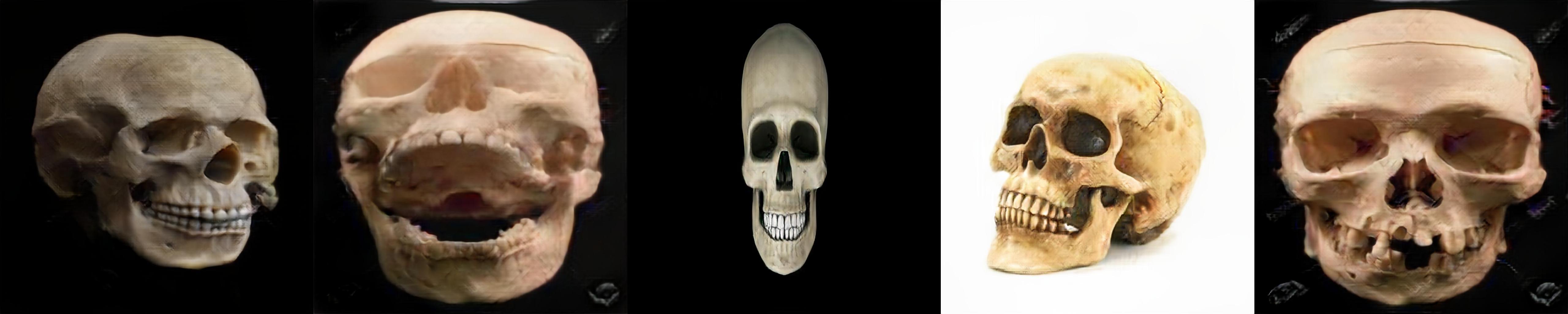}
    \end{minipage}
    
    \begin{minipage}{0.02\linewidth}
        \rotatebox{90}{\small Anime Face}
    \end{minipage}
    \begin{minipage}{0.97\linewidth}
        \includegraphics[width=0.373\linewidth]{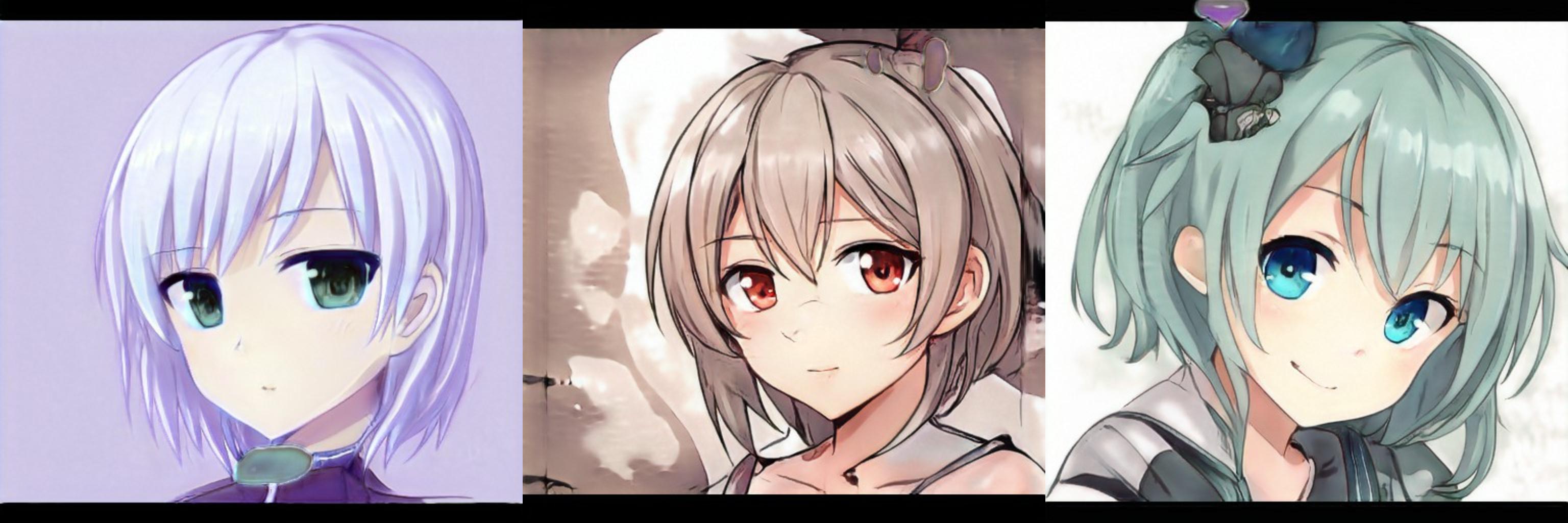}
        \includegraphics[width=0.621\linewidth]{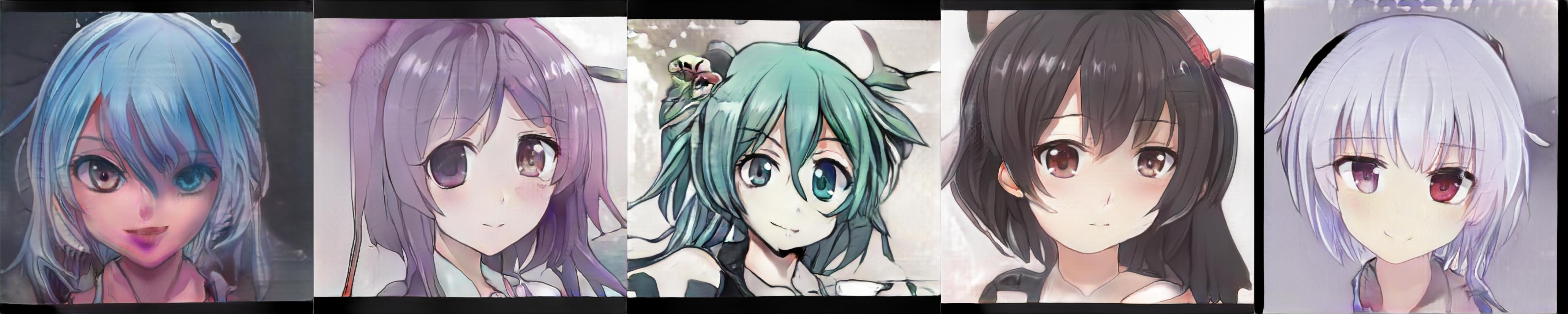}
    \end{minipage}

    \begin{minipage}{0.02\linewidth}
        \rotatebox{90}{\small BreCaHAD}
    \end{minipage}
    \begin{minipage}{0.97\linewidth}
        \includegraphics[width=0.373\linewidth]{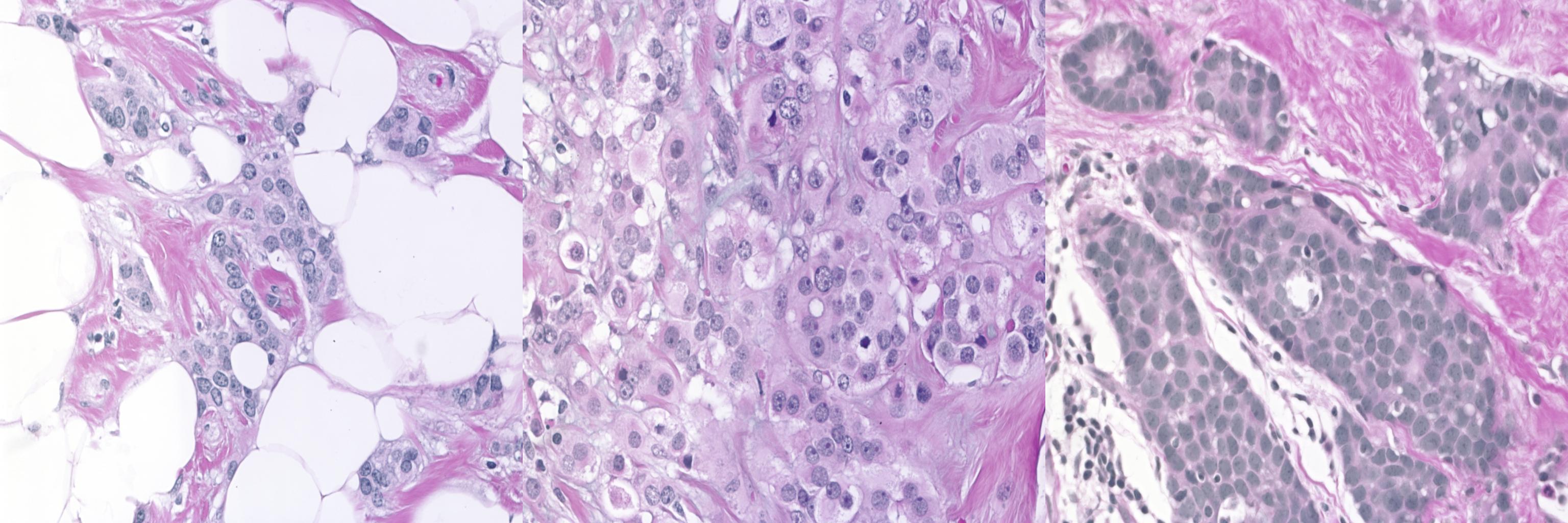}
        \includegraphics[width=0.621\linewidth]{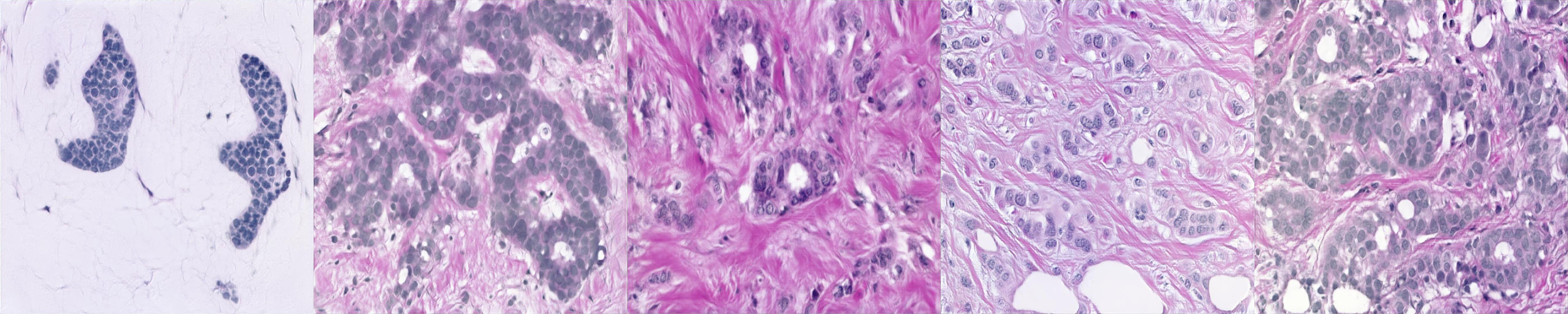}
    \end{minipage}

    \begin{minipage}{0.02\linewidth}
        \rotatebox{90}{\small MessidorSet1}
    \end{minipage}
    \begin{minipage}{0.97\linewidth}
        \includegraphics[width=0.373\linewidth]{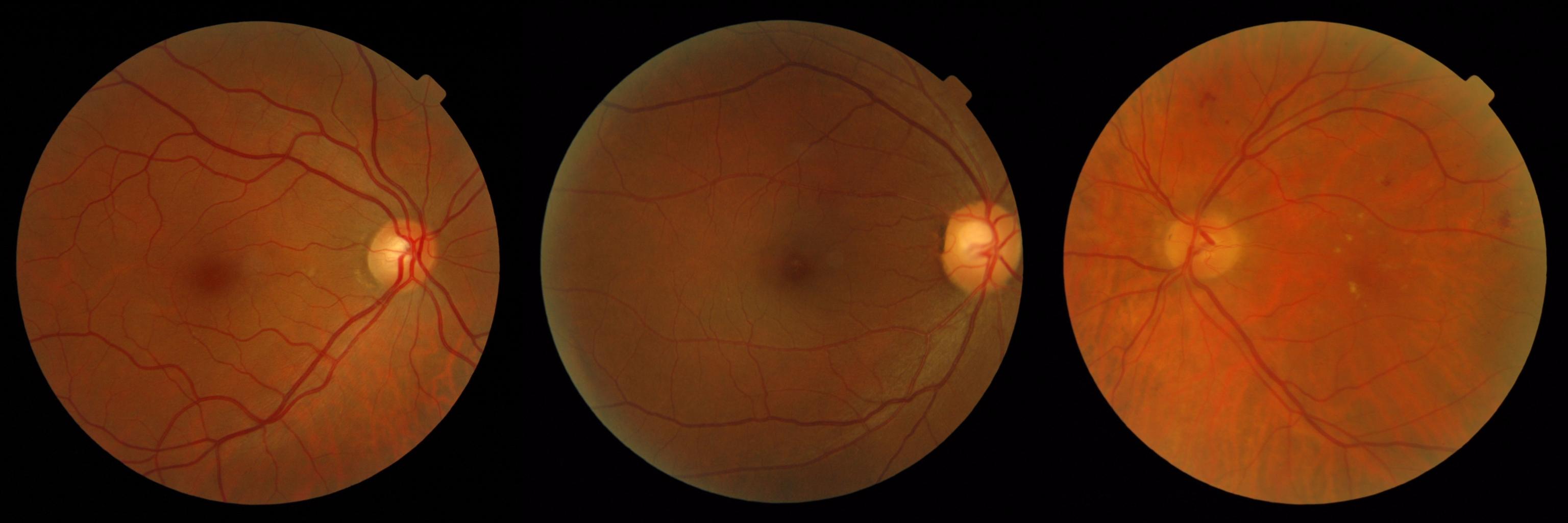}
        \includegraphics[width=0.621\linewidth]{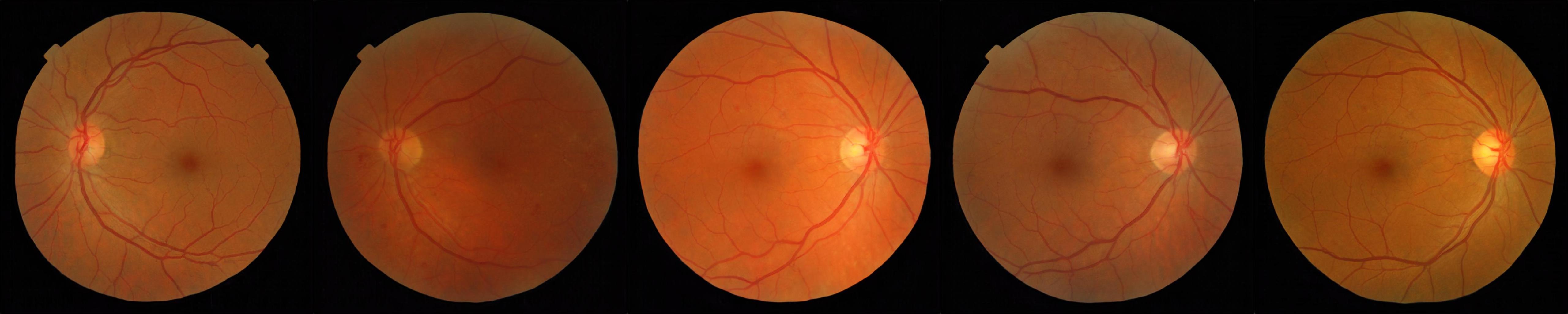}
    \end{minipage}

    \begin{minipage}{0.02\linewidth}
        \rotatebox{90}{\small Pokemon}
    \end{minipage}
    \begin{minipage}{0.97\linewidth}
        \includegraphics[width=0.373\linewidth]{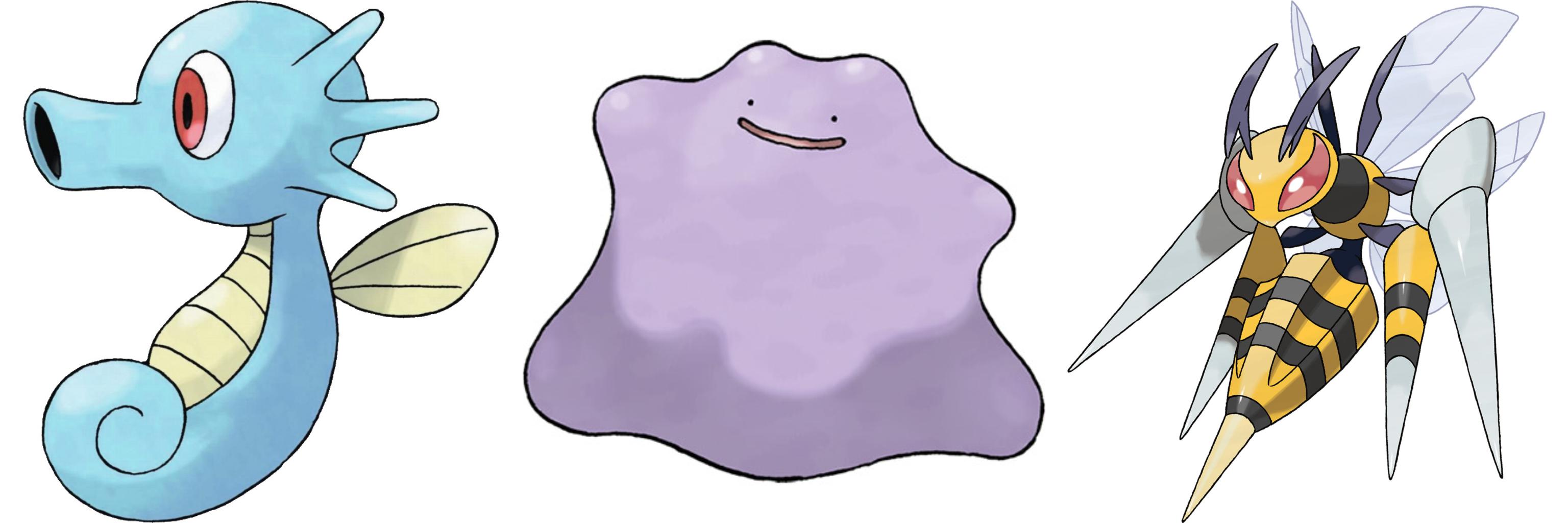}
        \includegraphics[width=0.621\linewidth]{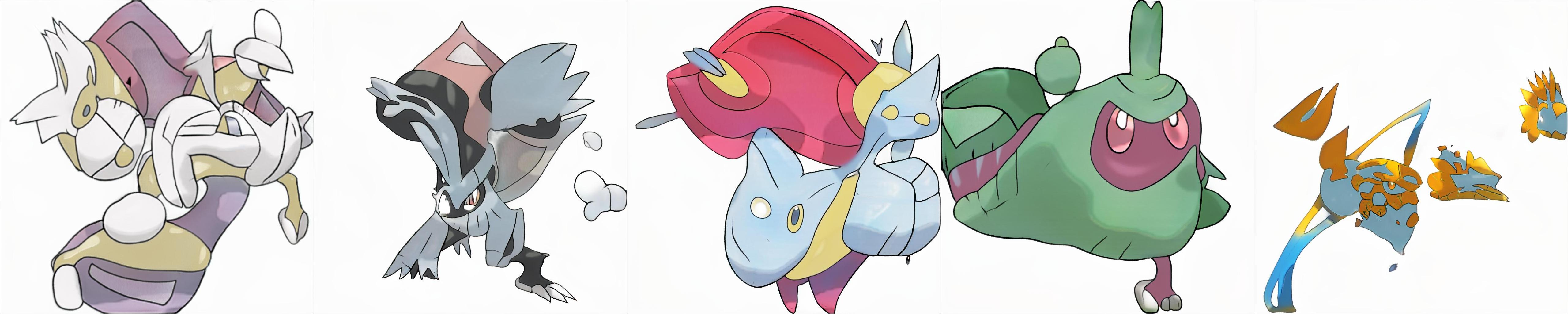}
    \end{minipage}

    \begin{minipage}{0.02\linewidth}
        \rotatebox{90}{\small ArtPainting}
    \end{minipage}
    \begin{minipage}{0.97\linewidth}
        \includegraphics[width=0.373\linewidth]{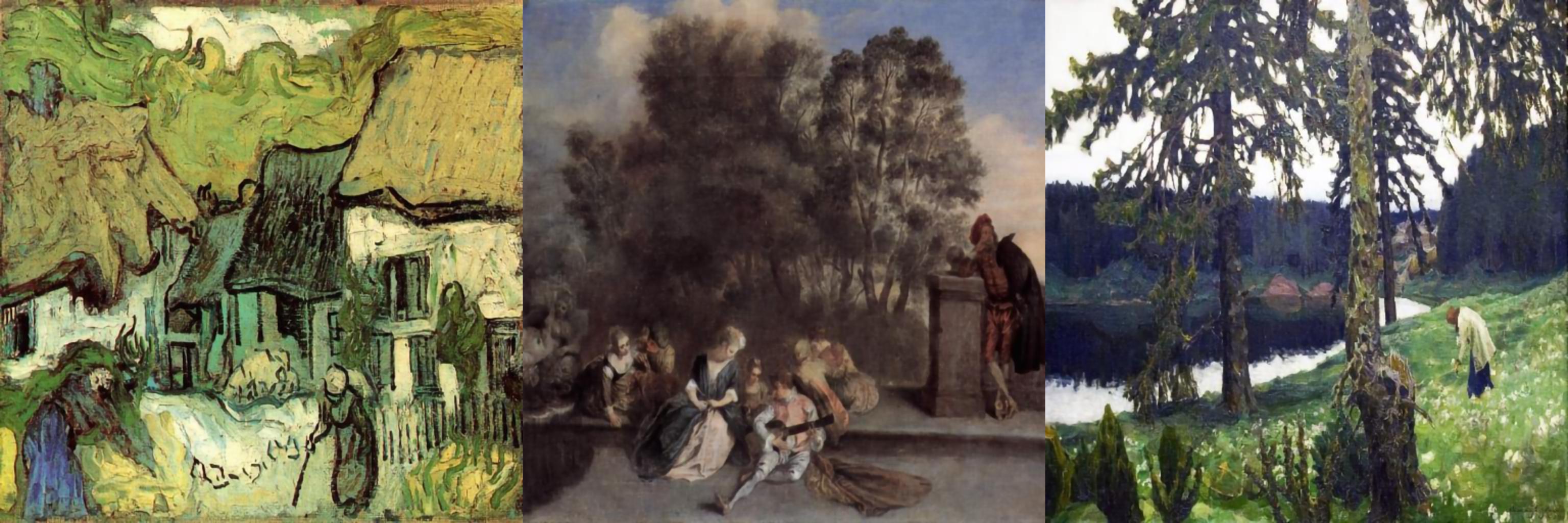}
        \includegraphics[width=0.621\linewidth]{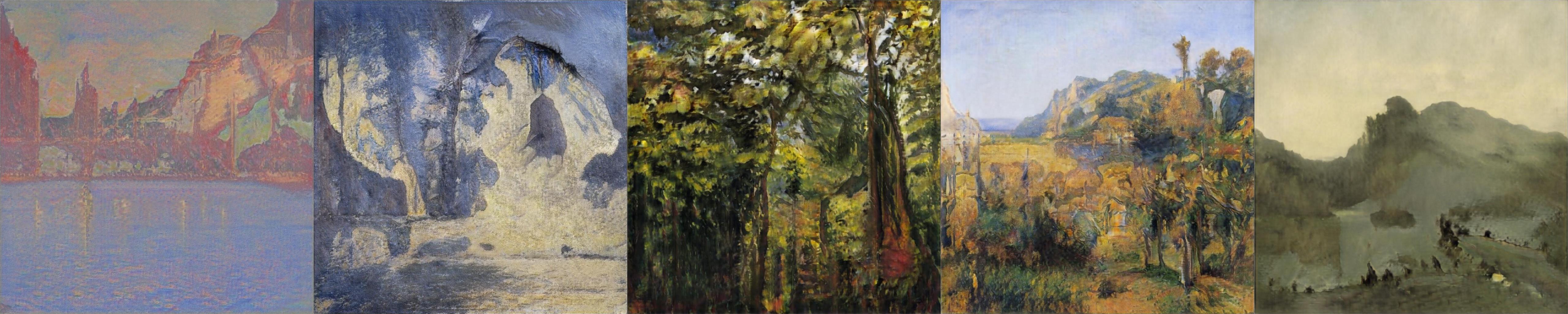}
    \end{minipage}

    \begin{subfigure}{0.372\linewidth}
    \centering
    \caption{Real images}
    \label{subfig:lowshot1024_real}
    \end{subfigure}
    \begin{subfigure}{0.622\linewidth}
    \centering
    \caption{\fastgandbig+\our.}
    \label{subfig:lowshot1024_chain}
    \end{subfigure}
    \caption{Qualitative results of \fastgandbig+\our on 7 few-shot image dataset ($1024\!\times\!1024$). (a) shows real training images and the (b) presents images generated by \fastgandbig+\our. \our is capable of generating photo-realistic images with fine details even from a limited number of training samples.}
    \label{fig:lowshot1024}
\end{figure*}

Figures \ref{fig:C10}, \ref{fig:C100}, \ref{fig:I64}, \ref{fig:lowshot256} and \ref{fig:lowshot1024} provide images generated on CIFAR-10, CIFAR-100, ImageNet, the 5 low-shot image and the 7 few-shot image datasets, with or without \our. The comparison highlights the enhancement in image quality and diversity achieved with the application of \our.

